\documentclass[pdflatex,sn-mathphys]{sn-jnl}
\usepackage[commandnameprefix=always, defaultcolor=red, final]{changes}

\usepackage{program}
\usepackage{hyperref}
\usepackage{amssymb}
\usepackage{graphicx}
\usepackage{amsmath}
\usepackage{enumitem}
\usepackage{multirow}
\usepackage{floatpag}
\usepackage{float}
\usepackage{array}
\usepackage{tabularray}
\usepackage{booktabs}
\usepackage{balance}
\usepackage{subcaption}
\usepackage{mathtools}
\usepackage{tabularx}
\usepackage[usestackEOL]{stackengine}
\newcommand*{\matminus}{%
  \leavevmode
  \hphantom{0}%
  \llap{%
    \settowidth{\dimen0 }{$0$}%
    \resizebox{1.1\dimen0 }{\height}{$-$}%
  }%
}
\newcommand\tab[1][0.25cm]{\hspace*{#1}}
\usepackage{amssymb}

\jyear{2022}%

\theoremstyle{thmstyleone}%
\theoremstyle{thmstyletwo}%

\theoremstyle{thmstylethree}%

\begin{document}

\title[ ]{A Domain-Region Based Evaluation of ML Performance Robustness to Covariate Shift}


\author[1]{\fnm{Firas} \sur{Bayram}}\email{firas.bayram@kau.se}

\author*[1,2]{\fnm{Bestoun} \sur{S. Ahmed}}\email{bestoun@kau.se}

\affil[1]{\orgdiv{Department of Mathematics and Computer Science}, \orgname{Karlstad University}, \orgaddress{ \city{Karlstad}, \postcode{65188}, \country{Sweden}}}

\affil[2]{\orgdiv{Department of Computer Science, Faculty of Electrical Engineering, }, \orgname{Czech Technical University in Prague}, \orgaddress{ \city{Prague}, \country{Czech Republic}}}

\abstract{Most machine learning methods assume that the input data distribution is the same in the training and testing phases. However, in practice, this stationarity is usually not met and the distribution of inputs differs, leading to unexpected performance of the learned model in deployment. The issue in which the training and test data inputs follow different probability distributions while the input-output relationship remains unchanged is referred to as covariate shift. In this paper, the performance of conventional machine learning models was experimentally evaluated in the presence of covariate shift. Furthermore, a region-based evaluation was performed by decomposing the domain of probability density function of the input data to assess the classifier's performance per domain region. Distributional changes were simulated in a two-dimensional classification problem. Subsequently, a higher four-dimensional experiments were conducted. Based on the experimental analysis, the Random Forests algorithm is the most robust classifier in the two-dimensional case, showing the lowest degradation rate for accuracy and F1-score metrics, with a range between 0.1\% and 2.08\%. Moreover, the results reveal that in higher-dimensional experiments, the performance of the models is predominantly influenced by the complexity of the classification function, leading to degradation rates exceeding 25\% in most cases}. It is also concluded that the models exhibit high bias towards the region with high density in the input space domain of the training samples.

\keywords{Covariate shift, concept drift, robust machine learning, classifier evaluation, model degradation}



\maketitle

\section{Introduction}
\label{sec:introduction}
Robustness to changes is one of the most exemplary properties of a high-quality machine learning (ML) model in this ever-shifting world. Maintaining this property will allow users to work with trustworthy ML systems that perform consistently and efficiently when exposed to the deployment environment. The issue arises from the assumption of most ML models that the data points are independently and identically distributed (i.i.d), which presume that the data are sampled from the same probability distribution. However, this assumption is often not satisfied in real-world applications \cite{HEUREUX12017Big}. This violation would cause the performance of the predictive model to degrade substantially in practice. The main reason for this \textit{model degradation} is that probability theory forms a central foundation for many data mining and machine learning techniques \cite{WITTEN2017Probabilistic}, as it provides a framework for quantifying uncertainty \cite{Brownlee2019Probability}. 

Changes in the underlying probability distributions have been extensively studied in the literature \cite{Gama2014Adaptation, Lu2016LearningUnder}. The situation in which there is an inconsistency between the joint probability distributions $P(x,y)$ of the training and test datasets is called \textit{dataset shift} \cite{Moreno2012DShift}.
Researchers have classified the general dataset shift phenomenon into several types based on the probabilistic source of change. These types are \textit{covariate shift}, \textit{concept drift} and \textit{prior probability shift} \cite{Raza2015EWMA}. Detailed taxonomy and definitions of change types can be found in our recent overview \cite{Bayram2022Degradation}. Covariate shift takes the form of a change between the marginal distributions of the training and test inputs $P_{tr}(x)$ and $P_{te}(x)$, while the conditional distribution $P(y \mid x)$ remains unchanged \cite{Shimodaira2000Covariate, Sugiyama2012IntroductiontoCS}. Whereas the concurrent occurrence of changes in both marginal and conditional distributions is called concept drift \cite{Tsymbal2004DefandRel}. Another type of dataset shift occurs when the distribution of classes $P(y)$ changes, known as prior probability shift \cite{Tasche2017Fisher}. The focus of this paper is on analyzing the behavior of ML models in covariate shift situations as it is one of the most common cases in practice \cite{Li2020Robust}.

Covariate shift could appear in many real-world domains, such as healthcare systems \cite{Subbaswamy2020Development}, computer vision \cite{Schneider2020Compvision}, NLP \cite{Duchi2019NLP}, and social media \cite{Fei2015Social}. In practice, many sources cause covariate shift situations. The main reason is that the training input is not a representative sample of the entire population \cite{He2019Unlearn}. In this case, the training samples do not represent the overall problem due to the biased data collection process towards a specific subpopulation, known as \textit{sample selection bias} \cite{Wiemann2022Bias}, for example, bias towards specific demographic groups. Therefore, the model will be implemented on an unseen distribution in deployment, which, as a consequence, would affect the performance of the model.

Analyzing the out-of-sample performance of ML models before deployment is not trivial because of the ubiquitousness of the system's evolution in real-life applications. The classical way is to perform out-of-sample validation using held-out datasets through cross-validation or bootstrap techniques. These techniques assume stationarity in the data distribution of training and test data. Therefore, they cannot fully estimate the model's success rate in the deployment environment where a distributional shift is likely to occur \cite{Tsuboi2009Direct}. Another approach to validate the ML models' performance is analyzing the accuracy and bias variability across data sub-populations. Exploring such variability would help diagnose the performance and potentially develop methods to mitigate its effects.

In this paper, the robustness of several common machine learning algorithms on synthetic data is evaluated in the presence of covariate shift in binary classification problems. An overall performance assessment and a region-based evaluation of the model's robustness are performed by decomposing the probability density function (pdf) of the input space domain of the test samples according to the input density distribution ratio between the test and training data samples. To this end, the contributions of this paper are summarized as follows:

\begin{enumerate}
    \item An evaluation framework is presented to assess the robustness of common ML classifiers in several drift scenarios.
    \item Comprehensive comparisons and experiments are conducted to measure model degradation rates under covariate shift in different classification problem settings.
    \item Decomposition of the input space based on the ratio of test-to-training input densities is proposed to evaluate the models' performance per domain-region.
\end{enumerate}

For the sake of visualization, exhaustive comparison experiments are conducted on two-dimensional artificial classification problems with covariate shifts formulated by simulating a broad spectrum of distribution drifts before evaluating the model robustness using additional higher-dimensional datasets. The experiments in this paper are designed to address the following primary research questions:

\tab \textbf{RQ1:} What are the main factors that affect the robustness of performance in covariate shift problems and lead to model degradation?

\tab \textbf{RQ2:} Which ML classification algorithms tend to be more robust (vulnerable) to specific problem settings?

\tab \textbf{RQ3:} How does the performance of the classifiers vary in the input space domain when decomposed by regions based on the test-to-training density ratio?

The remainder of this paper is organized as follows. Section \ref{sec:background} gives a general background on the covariate shift problem and provides an overview of the related work. Next, Section \ref{sec:settings} contains a detailed explanation of the settings specified for the experiments. The methodology for the performance evaluation of the ML models is detailed in Section \ref{sec:methodology}. Section \ref{sec:analysis} presents the results of the different experiments. Threats to validity are discussed in Section \ref{sec:threats}. Finally, Section \ref{sec:conclusion} concludes the work and sets out future directions for further extended use of this paper.

\section{Background and Related Work}
\label{sec:background}
This section provides an overview of the covariate shift problem and reviews the related work to evaluate the robustness of the ML model’s performance under covariate shift.

\subsection{Problem Formulation and Notation}
In binary classification problems, the goal is to obtain a prediction function $f: X \longmapsto Y$ that has been trained on training dataset $\mathcal{D}_{tr}=\left\{\left(x_{i}, y_{i}\right)\right\}_{i=1}^{N}$ of size $N$ drawn i.i.d from a joint distribution $p_{tr}(x,y)$, where $x_i \in \mathbb{R}^d$ is a $d$-dimensional data instance, or covariates vector $\mathbf{x}$, and $y_i \in \{-1,+1\}$ is the class label. The classifier's performance is evaluated on a test dataset $\mathcal{D}_{te}$ drawn from a joint distribution $p_{te}(x,y)$.
The classical method to learn the classifier is through solving the following empirical risk minimization (ERM) problem:
\begin{equation}
\label{minimization}
\small
\begin{split}
\min_{\theta \in \Theta} R(f) = \min_{\theta \in \Theta} \mathbb{E}_{p_{te}}[\ell (f_{\theta}(x),y)] \approx \\ \min _{\theta \in \Theta} \frac{1}{N} \sum_{i=1}^{N} \ell\left(f_{\theta}\left(x_{i}\right), y_{i}\right),
\end{split}
\end{equation}
where $\theta$ denotes the parameter vector, $\mathbb{E}_{p_{te}(x,y)}$ is the expectation over the test distribution $p_{te}(x,y)$, and $\ell$ is a selected loss function. In stationary distributions, i.e., when $p_{te}(x) = p_{tr}(x)$, ERM provides a \textit{consistent} estimator \cite{Sugiyama2007ICV}. While adaptation techniques are required for non-stationary distributions, such as covariate shift situations.


\subsection{Covariate Shift Adaptation}

\begin{figure}
\centering
\includegraphics[width=0.5\linewidth]{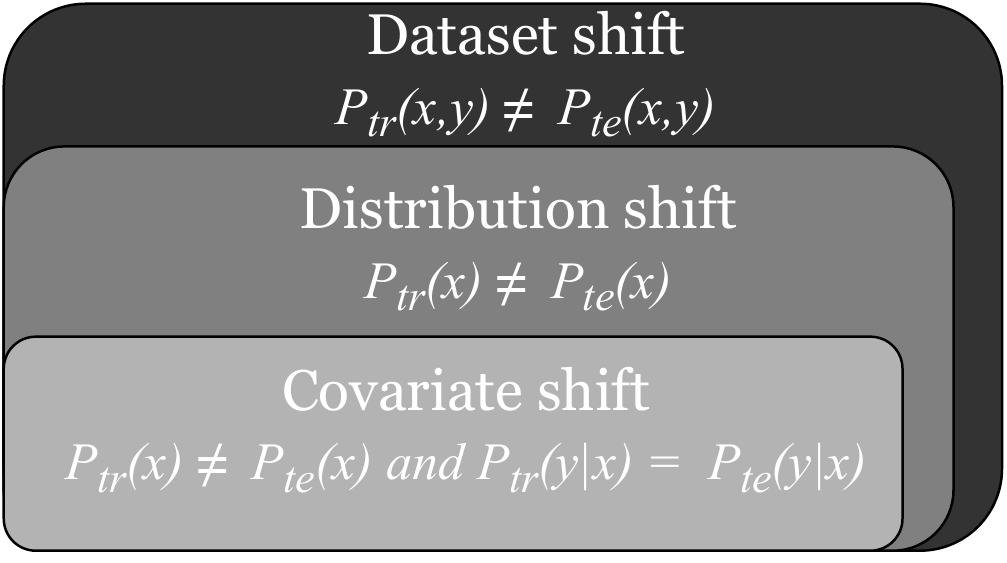}
\caption{Visual representation of the relationship between the types of distributional change.}
\label{classification}
\end{figure}

The minimization problem in Eq. \ref{minimization} works well under the assumption that $p_{te}(x,y)$ is the same as $p_{tr}(x,y)$, which usually does not hold in practice. Thus, an adjustment to ERM introduced in Eq.\ref{minimization} should be performed for dataset shift situations, that is, when $p_{te}(x,y) \neq p_{tr}(x,y)$. A probabilistic source for the dataset shift is a change in the input data distributions, i.e., when $p_{te}(x) \neq p_{tr}(x)$, which is referred to as \textit{distribution shift} \cite{Subbaswamy2021Evaluating}, also known as \textit{data drift} \cite{Liu2020Well}. As depicted in Fig. \ref{classification}, covariate shift is a subset of the generic distribution shift situation when the conditional probability distribution that represents the input-output rule is the same between the training and the test data \cite{Sakai2019PU}:
\begin{equation}
\label{covshift}
\small
p_{tr}(y \mid x) = p_{te}(y \mid x),\; p_{tr}(x) \neq p_{te}(x).
\end{equation}
One of the most common techniques to address the covariate shift problem is to employ an importance weighting function defined as:
$w(x) = \frac{p_{te}(x)}{p_{tr}(x)}$ which estimates the test-to-training density ratio \cite{Candela2009DatasetS}. Subsequently, the risk minimization problem in Eq.\ref{minimization} is adjusted to the importance-weighted risk:
\begin{equation}
\label{weighted_minimization}
\small
\hspace*{-0.3cm}
\min_{\theta \in \Theta} R(f) =   \min _{\theta \in \Theta} \frac{1}{N} \sum_{i=1}^{N} w(x_i)\ell\left(f_{\theta}\left(x_{i}\right), y_{i}\right).
\end{equation}

There are many popular algorithms to compute the importance weights $w(x_i)$, such as Kernel Mean Matching (KMM) \cite{Huang2006KMM}, Kullback-Leibler Importance Estimation Procedure (KLIEP) \cite{Sugiyama2007KLIEP} and least squares importance fitting (LSIF) \cite{Kanamori2009LSIF}. It was shown that the importance weighting procedure can provide consistent learning under covariate shift, and thus increase the robustness of the performance in changing distributions \cite{Sugiyama2008Direct}. However, the main drawback of the procedure is the high computational cost \cite{Chapaneri2019Covariate}.



\subsection{Measuring Robustness to Distribution Shift}
Much research is devoted to evaluating ML models' performance under distribution shifts. Such an evaluation would be useful for determining the model's performance after deployment, where the model might perform unexpectedly on unseen data points. A study has investigated the relationship between robustness and complexity of classifiers and concluded that complex classifiers remain more robust to changes than simple classifiers \cite{Alaiz2008Assessing}. Similarly, in \cite{Abbasian2010Robustness}, the authors have tested the robustness of common classifiers in distributional shift situations. The drift was simulated by changing a particular attribute and assessing the influence on the information gain. The authors have concluded that ML models that use more attributes, such as K-Nearest-Neighbours (KNN), in making predictions tend to be more robust than those which rely on fewer attributes, such as Naive Bayes and Logistic Regression.

Several recent studies have investigated the variability in the model's performance across data regions and sub-populations. MANDOLINE framework \cite{Chen2021Mandoline} was proposed to estimate the model's performance under distribution shift. The framework uses a \textit{labeled} validation set from the source distribution and an \textit{unlabeled} set from the target distribution. Users can use their prior knowledge to \textit{group} the data along the possible axes of distribution shift. Then, the reweighted performance estimates are computed. Another framework was proposed to proactively evaluate the model's performance on the \textit{worst-case} distribution \cite{Subbaswamy2021Evaluating}. Users choose two sets of variables, \textit{immutable} variables whose distribution should remain unchanged and \textit{mutable} variables whose distribution can be changed. The method identifies the sub-populations with the worst-case risk. Similarly, Sagawa \textit{et al.} \cite{Sagawa2019Distributionally} developed a training procedure that uses prior knowledge to form groups in training data and minimizes worst-case loss over these data groups.

In contrast to previous studies, another line of research follows a different approach by predicting the model's performance in distributional shift situations using the regression function. The Average Thresholded Confidence (ATC) method \cite{Garg2022Leveraging} was proposed to obtain a threshold on a model confidence score that enables the prediction of out-of-distribution model's accuracy. In another recent study, Guillory \textit{et al.} \cite{Guillory2021Predicting} proposed the so-called difference of confidences (DoC) method that predicts the model's performance under distribution shift. DoC is used to directly estimate the classifier's accuracy gap between the training and the target distributions. When lacking true labels for test sets, the authors \cite{Deng2021Labels} have derived distribution statistics that can benefit from the Automatic model Evaluation (AutoEval) problem and estimate the classifier's accuracy.

Evaluating the model's robustness has also been investigated in domain-specific problems. In image processing, the model's performance has been evaluated for many prominent image classification benchmark datasets, such as CIFAR and ImageNet \cite{Recht2019Imagenet, Taori2020Measuring, Miller2021Accuracy}, MNIST \cite{Yadav2019Cold}. The problems are constructed by inducing a wide range of distribution shifts. In natural language processing (NLP), Miller \textit{et al.} \cite{Miller2020Effect} evaluated the model's robustness to distribution shift using the popular Stanford Question Answering Dataset (SQuAD) \cite{Rajpurkar2016squad}. The authors noted that training models on more out-of-distribution data did not lead to improved robustness for ML models. Despite the rich literature present in the area, the evaluation of the performance of the ML model per region decomposed by test-to-training density ratio has not been yet investigated and is provided by this paper.

\section{Experimental Settings}
\label{sec:settings}
In this section, the specifications of the experiments performed using synthetic binary classification problems are illustrated. The experiments are simulated in a two-dimensional input space and a higher input space of four-dimensions. The experimental settings were made more diverse by selecting two- and four-dimensional space settings and omitting the three-dimensional case. And hence to gain a better insight into the performance of the different ML models in varied scenarios. 

For the experiment design part, the training dataset is considered to be sampled from a standard Gaussian distribution $X \sim \mathcal{N}_d(0, 1)$ throughout all experimental setups. Since any normally distributed variable $X$, with a particular population mean $\mu$ and standard deviation $\sigma$, can easily be transformed into a standard Gaussian distribution by applying the equation $z=\frac{X-\mu}{\sigma}$. Therefore, the training input density function is as follows:
\begin{equation}
\label{normaldist}
p_{tr}(x)=\frac{1}{(2 \pi)^{d / 2}} \exp \left(-\frac{1}{2} X^{T} X\right),
\end{equation}
where $d$ is the dimension of the input space. 

To generate the test data, several affine transformations are applied to the density of the training data. This method has been widely adopted in the literature to simulate the covariate shift problem in the dataset \cite{Sugiyama2012IntroductiontoCS, Sugiyama2007ICV}. The data points are considered to be sampled from a Gaussian distribution $X \sim \mathcal{N}_d(\mu, \Sigma)$ for the test data. Therefore, the test input density function is given by:
\begin{equation}
\label{drifteddist}
\small
p_{te}(x)= \frac{1}{(2 \pi)^{d / 2}\mid\Sigma\mid ^{1 / 2}} \exp \left(-\frac{1}{2}(X-\mu)^{T} \Sigma^{-1}(X-\mu)\right),
\end{equation}
where $\mu \in \mathbb{R} ^d$ is the mean and $\Sigma \in \mathcal{S}_{++}^{d}$ is the $d \times d$ symmetric positive-definite covariance matrix whose $(i,j)$th entry is $Cov[X_i, X_j]$. The statistics $\mu$ and $\Sigma$ represent the parameters of the affine transformations that simulate the drift in the experiments. Specifically, the mean $\mu$ simulates a drift induced by translation, and $\Sigma$ simulates a drift induced by scaling through the variance elements $var (x_i) = \sigma_{ii}^2$, or rotation through the correlation coefficient $\rho$ in the covariance elements $cov (x_i, x_j) = \sigma_{ij}^2 = \rho\sigma_i\sigma_j$. For each drift type, the experiments are run in a two-dimensional input space and a higher four-dimensional input space. 

Regarding the definition of class posterior probability functions, the formulations that are commonly used in covariate shift research have been followed \cite{sugiyama2007covariate, hachiya2012importance}. In particular, for the two-dimensional datasets settings, two different class posterior probability functions have been defined for classifying the points. The first function is defined as follows:
\begin{equation}
\label{f1}
\small
\begin{split}
F_{1}: p\Big(y=+1 \mid X=\left(x_{1}, x_{2}\right)\Big)= \\ \frac{1}{2}\left(1+\tanh \left(\min \left(0, x_{1}\right)+4 x_{2}\right)\right).
\end{split}
\end{equation}
The second class posterior probability function that was designed is more complex to learn and is defined as follows:
\begin{equation}
\label{f2}
\small
\begin{split}
F_{2}: p\Big(y=+1 \mid X=\left(x_{1}, x_{2}\right)\Big)=\\ \frac{1}{2}\left(1+\sin \left(\min \left(0, x_{1}\right)+2 x_{2}\right)\right).
\end{split}
\end{equation}

Similarly, for the higher-dimensional datasets settings, Two different class posterior probability functions have been defined as follows:
\begin{equation}
\label{f3}
\begin{split}
F_{3}: p\Big(y=+1 \mid X=\left(x_{1}, x_{2}, x_{3}, x_{4}\right)\Big)=\\
\frac{1}{2}\left(1+\tanh \left(\min \left(0, x_{1}\right)- x_{2}+2x_3+2x_4\right)\right),
\end{split}
\end{equation}
\begin{equation}
\label{f4}
\begin{split}
F_{4}: p\Big(y=+1 \mid X=\left(x_{1}, x_{2}, x_{3}, x_{4}\right)\Big)=\\
\frac{1}{2}\left(1+\sin \left(\min \left(0, x_{1}\right)+4 x_{2}-3x_3+2x_4\right)\right),
\end{split}
\end{equation}
where $p(y=-1\mid X)= 1- p(y=+1 \mid X)$ and the optimal decision boundary is the set of points that satisfy $p(y=-1\mid X)= p(y=+1 \mid X) = \frac{1}{2}$. For all experiments, training data points of size $N_{tr} = 20000$ are sampled from the probability density function (pdf) defined in Eq.\ref{normaldist}, test data points sampled from the same distribution of size $N_{ts} = 20000$, and another test data points of size $N_{td} = 20000$ sampled from the drifted distribution defined in Eq.\ref{drifteddist}. Parameter settings and specifications of the two-dimensional experiments are summarized in Table \ref{tab:2dsummary} in Appendix \ref{sec:AppA}, and the four-dimensional experiments in Table \ref{tab:4dsummary} in Appendix \ref{sec:AppA}. A more detailed description of each experiment is provided in the following subsections.

\begin{figure}
    \centering
    \includegraphics[width=0.46\textwidth ]{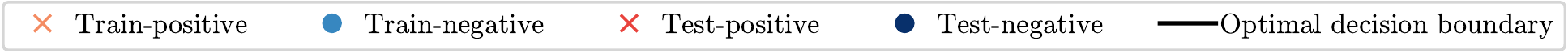}

    \centering
    \includegraphics[width=0.46\textwidth ]{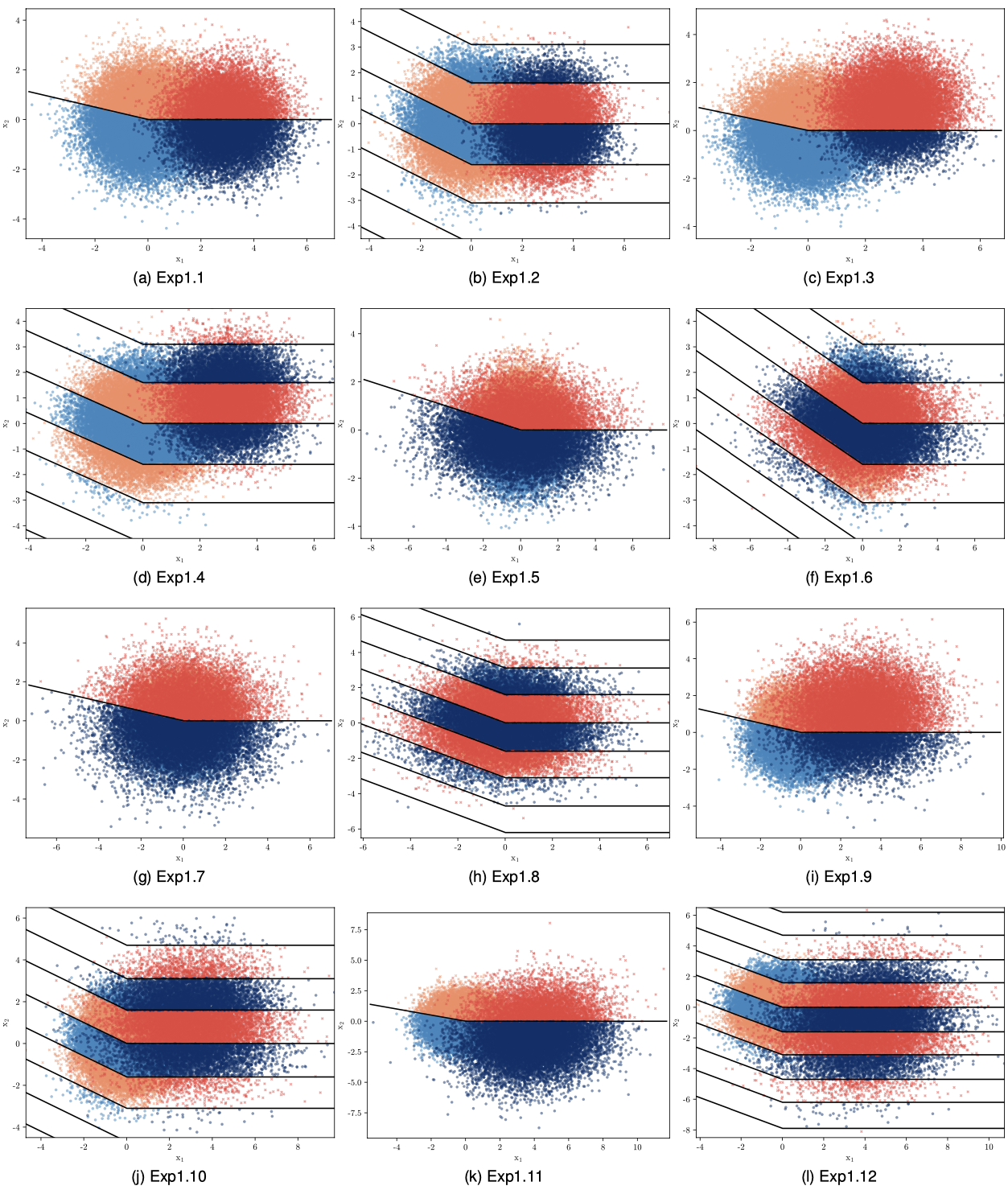}
    \caption{Training and test data points and the optimal decision boundary of the experiments}
    \label{fig:all_plots}
\end{figure}


\subsection{Drift Simulated by Translation}
\label{drift_translation}
The first set of experiments was created by shifting the mean of the data $\mu$ and fixing the covariance matrix $\Sigma$. For this type of drift, two settings were created. One setting is characterized by the one-axis translation simulating a \textit{local concept drift} case \cite{Almeida2018Adapting}, and the other by the two-axis translation simulating a \textit{global concept drift} case \cite{Khamassi2018DiscussionAR}. The details of the experiments are given as follows:

\begin{enumerate}[label=\textbf{\alph*)}]
\item \textbf{One-axis Translation:} For the two-dimensional input space, the translation vector $\begin{bmatrix} 3 & 0\end{bmatrix}^T$ is used to shift the original mean $\begin{bmatrix} 0 & 0\end{bmatrix}^T$. Using the aforementioned drift parameters, two experiments have been created, one experiment whose class posterior probability function defined in Eq.\ref{f1}, it is referred to as \textbf{Exp1.1}, and visualized in Fig. \ref{fig:all_plots}(a), and the other one using the function defined in Eq.\ref{f2}, it is referred to as \textbf{Exp1.2}, and visualized in Fig. \ref{fig:all_plots}(b).
\item \textbf{Two-axis Translation:} For the two-dimensional space, the original mean is shifted by the translation vector $\begin{bmatrix} 3 & 1\end{bmatrix}^T$. \textbf{Exp1.3} refers to the experiment whose class posterior probability function defined in Eq.\ref{f1}, and \textbf{Exp1.4} using the function defined in Eq.\ref{f2}. \textbf{Exp1.3} and \textbf{Exp1.4} are visualized in Fig.\ref{fig:all_plots}(c) and Fig.\ref{fig:all_plots}(d), respectively.
\\Similarly for the four-dimensional data, the original mean is shifted by the translation vector $\begin{bmatrix} 0 &-2&-1& 1\end{bmatrix}^T$. The experiment whose class posterior probability function is defined in Eq.\ref{f3} is denoted as \textbf{Exp2.1}, while \textbf{Exp2.2} denotes the experiment whose class posterior probability function is defined in Eq.\ref{f4}.
\end{enumerate}

\subsection{Drift Simulated by Scaling}
On the contrary of the drift explained in the previous section \ref{drift_translation}, the data have been transformed by scaling the covariance matrix $\Sigma$ and fixing the mean $\mu$ of the data. In this set of experiments, two settings have been created, one setting for one-axis scaling simulating a local concept drift situation and another one for two-axis scaling simulating a global concept drift situation. To scale the covariance matrix $\Sigma$ using the scalars $s_i > 0$ for $i =1,\dots ,d$ that represent the scaling factors for each axis direction, the scaling matrix can be written in matrix form as:

\begin{equation}
\label{scaling_matrix}
     S_d = \left[\begin{array}{cccc}s_{1} & 0 & \cdots & 0 \\ 0 & s_{2} & \cdots & 0 \\ \vdots & \vdots & \ddots & \vdots \\ 0 & 0 & \cdots & s_{d}\end{array}\right], \quad s_i>0.
\end{equation}

Therefore, the scaled covariance matrix $\Sigma^{'}$ can be found by computing the product $\Sigma^{'} = SS\Sigma$, the scaled covariance matrix is cataloged as:

\begin{equation}
\label{scaled_cov}
\begin{split}
     \Sigma^{'} = \left[\begin{array}{cccc}s_{1}^2 & 0 & \cdots & 0 \\ 0 & s_{2}^2 & \cdots & 0 \\ \vdots & \vdots & \ddots & \vdots \\ 0 & 0 & \cdots & s_{d}^2\end{array}\right] \left[\begin{array}{cccc}\sigma_{1}^2 & \sigma_{12}^2 & \cdots & \sigma_{1d}^2 \\ \sigma_{21}^2 & \sigma_{2}^2 & \cdots & \sigma_{2d}^2 \\ \vdots & \vdots & \ddots & \vdots \\ \sigma_{d1}^2 & \sigma_{d2}^2 & \cdots & \sigma_{d}^2\end{array}\right]\\
     = \left[\begin{array}{cccc}(s_{1}\sigma_{1})^2 & (s_{1} \sigma_{12})^2 & \cdots & (s_{1} \sigma_{1d})^2 \\ (s_{2} \sigma_{21})^2 & (s_{2} \sigma_{2})^2 & \cdots & (s_{2} \sigma_{2d})^2 \\ \vdots & \vdots & \ddots & \vdots \\ (s_{d} \sigma_{d1})^2 & (s_{d}\sigma_{d2})^2 & \cdots & (s_{d}\sigma_{d})^2\end{array}\right],
\end{split}
\end{equation}
where $s_i$ is the scaling factor of dimension $i$ and $\sigma_{ij}$ is the covariance between dimensions $i$ and $j$. The details of the experiments are given as follows:
\begin{enumerate}[label=\textbf{\alph*)}]

\item \textbf{One-axis Scaling:} 
For the two-dimensional input space, the scaling matrix $S=\left[\begin{array}{cc}2 & 0 \\ 0 & 1\end{array}\right]$ is used to transform the original covariance matrix $\Sigma=\left[\begin{array}{cc}1 & 0 \\ 0 & 1\end{array}\right]$. Applying Eq.\ref{scaled_cov}, the scaled covariance matrix is $\Sigma^{'}=\left[\begin{array}{cc}4 & 0 \\ 0 & 1\end{array}\right]$. Using the scaled covariance matrix $\Sigma^{'}$, two experiments have been created: \textbf{Exp1.5}, visualized in Fig.\ref{fig:all_plots}(e), and \textbf{Exp1.6}, visualized in Fig.\ref{fig:all_plots}(f), whose class posterior probability function defined in Eq.\ref{f1} and Eq.\ref{f2}, respectively.

\item \textbf{Two-axis Scaling:} For the two-dimensional space, the original covariance matrix is transformed by the scaling matrix $S=\left[\begin{array}{cc}\sqrt{3} & 0 \\ 0 & \sqrt{2}\end{array}\right]$, resulting in a scaled covariance matrix $\Sigma^{'}=\left[\begin{array}{cc}3& 0 \\ 0 & 2\end{array}\right]$. \\\textbf{Exp1.7}, visualized in Fig.\ref{fig:all_plots}(g), and \textbf{Exp1.8}, visualized in Fig.\ref{fig:all_plots}(h), refer to the experiments whose class posterior probability function defined in Eq.\ref{f1} and Eq.\ref{f2}, respectively.\\
Similarly for the four-dimensional data, the original covariance matrix is transformed by the scaling matrix:\\ $S=\left[\begin{array}{cccc}\sqrt{3} & 0 &0 & 0\\ 0 & \sqrt{2}&0&0\\ 0 & 0  &\sqrt{2} & 0\\0 & 0 & 0& \sqrt{3} \end{array}\right]$. \\By applying Eq.\ref{scaling_matrix}, the scaled covariance matrix is found:\\ $\Sigma^{'}=\left[\begin{array}{cccc}3 & 0 &0 & 0\\ 0 & 2&0&0\\ 0 & 0  &2 & 0\\0 & 0 & 0& 3 \end{array}\right]$.\\ \textbf{Exp2.3} and \textbf{Exp2.4} denote the experiments whose class posterior probability function defined in Eq.\ref{f3} and Eq.\ref{f4}, respectively.
\end{enumerate}


\subsection{Drift Simulated by Translation and Scaling}
In this set of experiments, a combination of linear transformations of the data points is used by translating and scaling the dataset to simulate the drift. For two-dimensional data, translation is formed using the translation vector $\begin{bmatrix} 3 & 1\end{bmatrix}^T$, while scaling is formed using the scaling matrix $S=\left[\begin{array}{cc}\sqrt{3} & 0 \\ 0 & \sqrt{2}\end{array}\right]$, resulting in a scaled covariance matrix $\Sigma^{'}=\left[\begin{array}{cc}3& 0 \\ 0 & 2\end{array}\right]$. \\\textbf{Exp1.9}, visualized in Fig.\ref{fig:all_plots}(i), and \textbf{Exp1.10}, visualized in Fig.\ref{fig:all_plots}(j), refer to the experiments whose class posterior probability function defined in Eq.\ref{f1} and Eq.\ref{f2}, respectively.

\subsection{Drift Simulated by Translation, Scaling and Rotation}
In this set of experiments, the drift is simulated by applying three different affine transformations to the dataset by translating, scaling, and rotating the data points. The covariance matrix after scaling and rotating can be calculated by means of the following matrix multiplication: 

\begin{equation}
\label{scalerotate}
    \Sigma^{''} = R \Sigma^{'} R^T, \quad \Sigma^{'} = SS\Sigma, 
\end{equation}
where $R$ is the rotation matrix. Note that the order of the transformation methods affects the end results. In these experiments, the assumption is made that the data is being scaled and rotated.

For the rotation of the data points, a rotation matrix $R$ is defined to rotate the data points through a desired angle $\theta$ about their origin in space. In a two-dimensional space, the general definition of the rotation matrix $R$ is given by the following equation:
\begin{equation}
\label{2drotate}
R (\theta)= 
\left[\begin{array}{cc}
\cos \theta & -\sin \theta \\
\sin \theta & \cos \theta
\end{array}\right].
\end{equation}
For the four-dimensional space, a basic rotation matrix $R$ is given by the following equation:
\begin{equation}
\label{4drotate}
R (\theta)=
\left[\begin{array}{cccc}
\cos \theta & -\sin \theta & 0 & 0 \\
\sin \theta & \cos \theta & 0 & 0 \\
0 & 0 & 1 & 0 \\
0 & 0 & 0 & 1
\end{array}\right],
\end{equation}
where $\theta$ is the rotation angle.

In our experimental settings, for the two-dimensional case, a translation vector $\begin{bmatrix} 4 & -1\end{bmatrix}^T$ is used to shift the original mean $\mu$. For scaling and rotation, the scaling matrix $S=\left[\begin{array}{cc}2 & 0 \\ 0 & \sqrt{3}\end{array}\right]$ is used to scale and rotate the data by $45^{\circ}$. This parameter setting would result in the following scaled and rotated covariance matrix by applying Eq.\ref{scalerotate} and the rotation matrix defined in Eq.\ref{2drotate}: 
$\Sigma^{''}=\left[\begin{array}{cc}3.5& 0.5 \\ 0.5 & 3.5\end{array}\right]$.\\ \textbf{Exp1.11}, visualized in Fig.\ref{fig:all_plots}(k), and \textbf{Exp1.12}, visualized in Fig.\ref{fig:all_plots}(l), refer to the experiments whose class posterior probability function defined in Eq.\ref{f1} and Eq.\ref{f2}, respectively.

For four-dimensional data, the original mean $\mu$ is shifted through the translation vector $\begin{bmatrix} 0 &-2&-1& 1\end{bmatrix}^T$ and scale the data points through the scaling matrix $S=\left[\begin{array}{cccc}\sqrt{3} & 0  &0 & 0\\ 0 & \sqrt{2}&0&0\\ 0 & 0 &\sqrt{2} & 0\\0 & 0 & 0& \sqrt{3} \end{array}\right]$, and then rotate the scaled data points by $45^{\circ}$.
This parameter setting would result in the following scaled and rotated covariance matrix by applying Eq.\ref{scalerotate} and the rotation matrix defined in Eq.\ref{4drotate}: 

 $\Sigma^{''}=\left[\begin{array}{cccc}2.5 & 0.5 &0 & 0\\ 0.5 & 2.5& 0 &0\\ 0 & 0 &2 & 0\\0 & 0 & 0& 3 \end{array}\right]$. \textbf{Exp2.5} and \textbf{Exp2.6} denote the experiments whose class posterior probability function is defined in Eq.\ref{f3} and Eq.\ref{f4}, respectively.



\section{Evaluation Methodology}
In this section, the methodology to address the research questions and quantify the robustness of ML models' performance to covariate shift situations is explained. The ML models used for evaluation are first iterated, followed by a description of the evaluation metrics used to assess performance. Lastly, the decomposition of the pdfs by the density ratio regions is described.

\label{sec:methodology}
\subsection{ML Models}
In this paper, the performance of several popular conventional ML algorithms is compared under covariate shift to address \textbf{RQ1} and \textbf{RQ2}. To draw conclusions and gain better insight into the robustness to distributional shifts, the chosen ML algorithms have been diversified by selecting classification algorithms from different families.

The selected ML algorithms for our evaluation are the following:

\begin{itemize}
    \item \textbf{Support Vector Machines (SVM)}: SVM \cite{Cortes1995SVM} is one of the most popular ML classification algorithms. SVM classifies the points by finding the optimal separating hyperplane between the classes.
    \item \textbf{Logistic regression (LR)}: LR models are used to predict the likelihood of the target class and are usually solved by maximum likelihood methods \cite{Hastie2009Elements}. For binary classification, the logistic regression model predicts the probability as follows:
    \begin{equation}
    p(y=\pm 1 \mid \mathbf{x}, \mathbf{w})= \frac{1}{1+\exp \left(-y \mathbf{w}^{\mathrm{T}} \mathbf{x}\right)}.
    \end{equation}
    \item \textbf{Random forests (RF)}: Random forests are an ensemble of decision tree models and an extension of the bagging method \cite{Breiman2001RF}. It creates a collection of de-correlated decision trees and then aggregates the predictions of each individual tree.
    \item \textbf{Gaussian Naive Bayes (GNB)}: GNB is a simple probabilistic classification algorithm that implements Bayes' Theorem \cite{Bishop2006Pattern}. It involves calculating the posterior probability by applying the Bayes rule:
    \begin{equation}
    \small
    p(y=\pm 1 \mid  \mathbf{x}) = p(y=\pm 1) \frac{\prod_{i=1}^{d} p(x_i \mid y=\pm 1)} {p( \mathbf{x})}.
    \end{equation}
   GNB algorithm assumes that the class-conditional probability distribution, i.e. $p(\mathbf{x} \mid y=\pm 1)$, follows the Gaussian distribution and estimates the mean and standard deviation parameters from the training data.
    \item \textbf{K-Nearest-Neighbours (KNN)}: KNN is a popular algorithm that uses the neighborhood of the data point to make predictions \cite{Hand2007Principles}. A voting mechanism is used to determine the class of the new data point. The votes are retrieved from the \textit{k} data points that are the closest to the new data point.
\end{itemize}


\subsection{Evaluation Metrics}
Observing the loss rate in the performance of ML models is a commonly adopted approach to measure the robustness of the learning algorithms to distributional changes \cite{Ovadia2019Uncertainty, Amodei2016Concrete}. The performance is usually measured on two data samples, the training samples and out-of-distribution samples. This paper evaluates the robustness of the ML models used in the experiments using the degradation rate of different performance metrics, including accuracy, F-score, and Matthews coefficient of correlation (MCC). The percentage of data points that are correctly classified represents the accuracy of the model. With \textbf{TP} representing the count of true positives, \textbf{TN} represents the count of true negatives, \textbf{FP} represents the count of false positives, and \textbf{FN} the count of false negatives, the accuracy can be expressed as:
\begin{equation}
\operatorname { Accuracy }=\frac{TP+TN}{TP+TN+FP+FN}.
\end{equation}

On the other hand, the F-score, also called the F1 score, is the harmonic mean between the precision and recall metrics \cite{Lesch2005metric}. Recall measures the ratio of positive data points that are correctly classified, whereas precision measures the ratio of data points that are classified as positive that are truly positive:
\begin{equation}
\operatorname{F-score} = 2 \cdot \frac{\text {Precision} \cdot \text {Recall}}{\text {Precision}+\text {Recall}},
\end{equation} 
where
\begin{equation}
\operatorname{Recall}=\frac{TP}{TP+FN},
\end{equation}

\begin{equation}
\operatorname{Precision}=\frac{TP}{TP+FP}.
\end{equation}

Similarly, MCC uses the confusion matrix to score the quality of the classification in the interval [-1, +1], with -1 denoting perfect misclassification and +1 denoting perfect classification, where 0 means prediction more of a random prediction:

\begin{equation}
\footnotesize
\operatorname{MCC}=\frac{TP \cdot TN-FP \cdot FN}{\sqrt{(TP+FP) \cdot(TP+FN) \cdot(TN+FP) \cdot(TN+FN)}}.
\end{equation}


\subsection{PDF Domain Region Decomposition}
To address \textbf{RQ3} and evaluate the performance on subpopulations of the dataset according to the density ratio between the test and training datasets, the input space domain of the test dataset is decomposed into two regions. The first region, denoted \textbf{R1}, is the region with a high density of the training dataset; i.e. where the density ratio is $r = \frac{p_{te}(x)}{p_{tr}(x)} \leq 1$. The other region denoted \textbf{R2}, is the region with high density of the test dataset, that is, where the density ratio $r = \frac{p_{te}(x)}{p_{tr}(x)} >1$. To decompose the pdfs by density regions, the following equation must be solved:
\begin{equation}
\label{intersection}
    {p_{te}(x)} = {p_{tr}(x)}.
\end{equation}
Since the distribution is multivariate in our experiments, the pdfs are hypersurfaces of dimension $d$ embedded in $(d+1)$-dim space, and they intersect in a set of $d$-dim points that lie on a hypersurface $\mathfrak{H}$, also embedded in ($d+1$)-dim space; where $d$ is the dimension of the data points. The solution of Eq.\ref{intersection} would provide the hypersurface $\mathfrak{H}$ equation.

Solving Eq.\ref{intersection} for the $d$-dimensional case of our experiments; $X \sim \mathcal{N}_d(\mu, \Sigma)$:\\
\begin{equation}
\begin{split}
    \frac{1}{(2 \pi)^{d / 2}\mid\Sigma\mid^{1 / 2}} \exp \left(-\frac{1}{2}(X-\mu)^{T} \Sigma^{-1}(X-\mu)\right) \\=
    \frac{1}{(2 \pi)^{d / 2}} \exp \left(-\frac{1}{2}X^{T} X\right),
    \end{split}
\end{equation}
results in the hypersurface $\mathfrak{H}$ defined by the following equation:
\begin{equation}
\label{ndimsolution}
      \mathfrak{H}(X) :(X-\mu)^{T} \Sigma^{-1}(X-\mu) + \log(\mid\Sigma\mid)- X^{T} X =0.
\end{equation}

The data points that lie in the region \textbf{R1} are those points which satisfy the inequality:
\begin{equation}
\label{ndimineqsolution}
      (X-\mu)^{T} \Sigma^{-1}(X-\mu) + \log(\mid\Sigma\mid)- X^{T} X \leq 0,
\end{equation}
otherwise, the points lie in the region \textbf{R2}.

Specifically, solving Eq.\ref{intersection} for the two-dimensional datasets of our experiments; i.e. where the test input density is $\left(\begin{array}{l}x \\ y\end{array}\right) \sim N\left[\left(\begin{array}{l}\mu_{1} \\ \mu_{2}\end{array}\right),\left(\begin{array}{cc}\sigma_{1}^{2} & \rho \sigma_{1} \sigma_{2} \\ \rho \sigma_{1} \sigma_{2} & \sigma_{2}^{2}\end{array}\right)\right]$:\\

\begin{equation}
\label{ndsurface}
\begin{split}
\frac{1}{2 \pi\left(1-\rho^{2}\right)^{1 / 2} \sigma_{1} \sigma_{2}} \exp \left\{-\frac{1}{2\left(1-\rho^{2}\right)}\left[\left(\frac{x-\mu_{1}}{\sigma_{1}}\right)^{2}\right.\right. \\
\left.\left.-2 \rho\left(\frac{x-\mu_{1}}{\sigma_{1}}\right)\left(\frac{y-\mu_{2}}{\sigma_{2}}\right)+\left(\frac{y-\mu_{2}}{\sigma_{2}}\right)^{2}\right]\right\} \\
=\frac{1}{2 \pi} \exp \left[-\frac{1}{2}\left(x^{2}+y^{2}\right)\right],
\end{split}
\end{equation}
results in a 2D curve that lies on a surface $\mathbf{S}$. We obtain the following equation defining the surface $\mathbf{S}$ in the 3D space:
\begin{equation}
\label{2dsurface}
\begin{split}
\mathbf{S}(x,y) :\left[\frac{-1}{2}+\frac{b}{\sigma_{1}^{2}}\right] x^{2}+\left[\frac{-1}{2}+\frac{b}{\sigma_{2}^{2}}\right] y^{2} -\\
\left[\frac{2 b \mu_{1}}{\sigma_{1}^{2}}-\frac{2 b \rho \mu_{2}}{\sigma_{1} \sigma_{2}}\right] x
-\left[\frac{2 b \mu_{2}}{\sigma_{2}^{2}}-\frac{2 b \rho \mu_{1}}{\sigma_{1} \sigma_{2}}\right] y- \\ \frac{2 b \rho}{\sigma_{1} \sigma_{2}} x y=r,
\end{split}
\end{equation}
where:
\begin{equation}
r=c+\frac{2 b \rho}{\sigma_{1} \sigma_{2}} \mu_{1} \mu_{2},
\end{equation}
\begin{equation}
c=a-\frac{b \mu_{1}^{2}}{\sigma_{1}^{2}}-\frac{b \mu_{2}^{2}}{\sigma_{2}^{2}},
\end{equation}
\begin{equation}
 b=\frac{1}{2\left(1-\rho^{2}\right)},
\end{equation}
\begin{equation}
a=\log \frac{1}{\sigma_{1} \sigma_{2} \sqrt{1-\rho^{2}}}. 
\end{equation}

The data points that lie in the region \textbf{R1} are those points which satisfy the inequality:
\begin{equation}
\begin{split}
\left[\frac{-1}{2}+\frac{b}{\sigma_{1}^{2}}\right] x^{2}+\left[\frac{-1}{2}+\frac{b}{\sigma_{2}^{2}}\right] y^{2} -
\left[\frac{2 b \mu_{1}}{\sigma_{1}^{2}}-\frac{2 b \rho \mu_{2}}{\sigma_{1} \sigma_{2}}\right]\\ x
-\left[\frac{2 b \mu_{2}}{\sigma_{2}^{2}}-\frac{2 b \rho \mu_{1}}{\sigma_{1} \sigma_{2}}\right] y-\frac{2 b \rho}{\sigma_{1} \sigma_{2}} x y \geq r,
\end{split}
\end{equation}
otherwise, the points lie in the region \textbf{R2}.


The surface $\mathbf{S}$ has a particular shape based on the type of drift simulated in the pdf of the dataset. Specifically, in the case of drifts simulated by two-axis translation, the surface $\mathbf{S}$ is a vertical plane in the 3D space defined by the following equation:
\begin{equation}
2 \mu_{1} x - \mu_{1}^2 + 2 \mu_{2} y - \mu_{2}^2 = 0.
\end{equation}
For a translation on one axis, the surface $\mathbf{S}$ is a vertical plane that is parallel to the xz-plane or the yz-plane. In particular, if the translation is along the x-axis, the surface $\mathbf{S}$ is a vertical plane parallel to the yz-plane and is given by the equation:
\begin{equation}
    x = \frac{\mu_1}{2}.
\end{equation}
On the contrary, if the translation is along the y-axis, the surface $\mathbf{S}$ is a vertical plane parallel to the xz-plane and is given by the equation:
\begin{equation}
    y = \frac{\mu_2}{2}.
\end{equation}

In the case of two-axis scaling; $\sigma_1 > 1$ and $\sigma_2 > 1$; the surface $\mathbf{S}$ is an elliptic cylinder defined by the equation:
\begin{equation}
\frac{x^{2}}{\alpha^{2}}+\frac{y^{2}}{\beta^{2}}=1,
\end{equation}
where:
\begin{equation}
\alpha^2=\frac{2\sigma_{1}^2 \log (\sigma_{1} \sigma_{2})}{\sigma_{1}^{2}-1},
\end{equation}
and:
\begin{equation}
\beta^2=\frac{2\sigma_{2}^2 \log (\sigma_{1} \sigma_{2})}{\sigma_{2}^{2}-1}.
\end{equation}
In the special case of scaling at the same magnitude in both dimensions; i.e., $\sigma_1 =\sigma_2 \implies \alpha=\beta$; the surface $\mathbf{S}$ is a right-circular cylinder.

In the case of a one-axis scaling, the two pdfs intersect in curves that lie on two parallel planes. If $\sigma_1 =1$; these parallel planes are defined by the following equation:
\begin{equation}
    y^2 = \frac{2 \sigma_2^2 \log\sigma_2}{\sigma_2^2-1}.
\end{equation}
Whereas if $\sigma_2 =1$; the parallel planes are defined by the following equation:
\begin{equation}
    x^2 = \frac{2 \sigma_1^2 \log\sigma_1}{\sigma_1^2-1}.
\end{equation}

\begin{figure*}[!t]
     \begin{flushright}
    \includegraphics[width=0.16\textwidth]{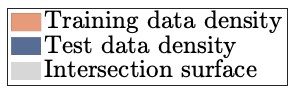}
    \end{flushright}

\centering
\subfloat[Exp1.1 and Exp1.2 densities; Intersection surface equation\\  $\mathbf{S}: x-\frac{3}{2}=0$]{\includegraphics[width=0.32\textwidth]{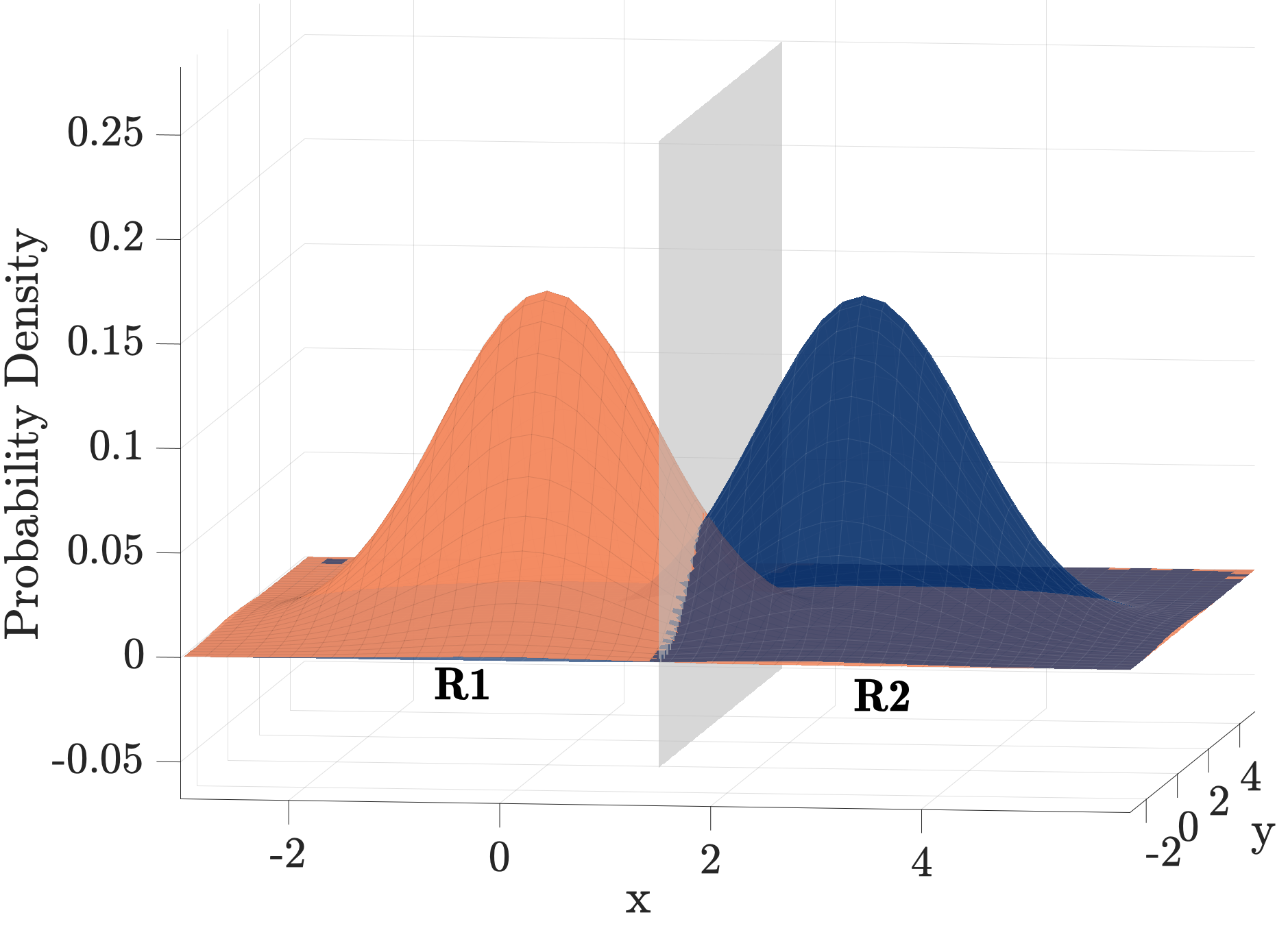}%
\label{exp1pdf}}
\hfil
\subfloat[Exp1.3 and Exp1.4 densities; Intersection surface equation\\ $\mathbf{S}: -3 x-y+5=0$]{\includegraphics[width=0.32\textwidth]{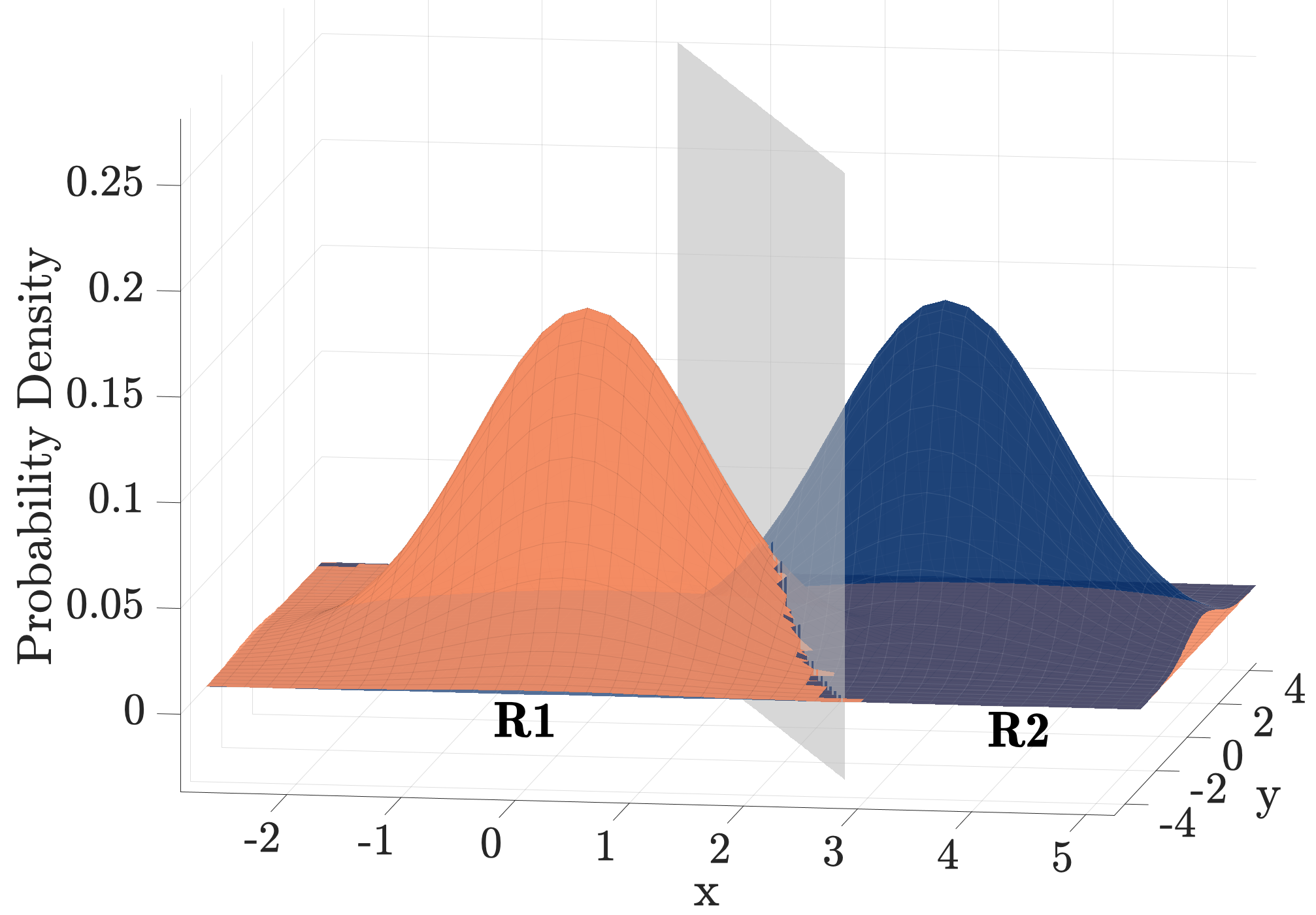}%
\label{exp3pdf}}
\hfil
\subfloat[Exp1.5 and Exp1.6 densities; Intersection surface equation\\ $\mathbf{S}: -3 \frac{x^{2}}{8}+\log(2)=0$]{\includegraphics[width=0.32\textwidth]{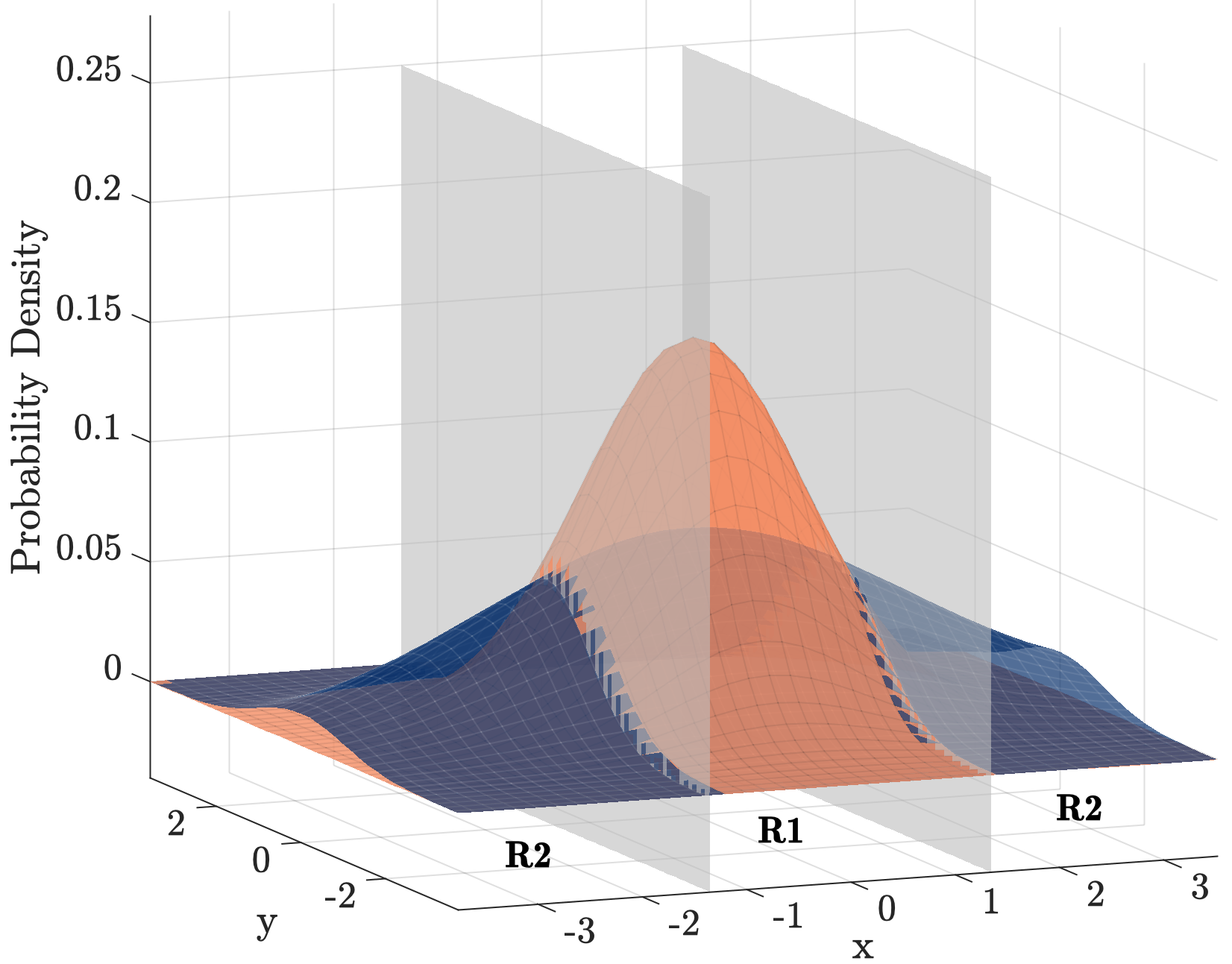}%
\label{exp5pdf}}
\hfil
\subfloat[Exp1.7 and Exp1.8 densities; Intersection surface equation\\ $\mathbf{S}: \frac{-x^{2}}{3}-\frac{y^{2}}{4}+\frac{\log (6)}{2}=0$]{\includegraphics[width=0.32\textwidth]{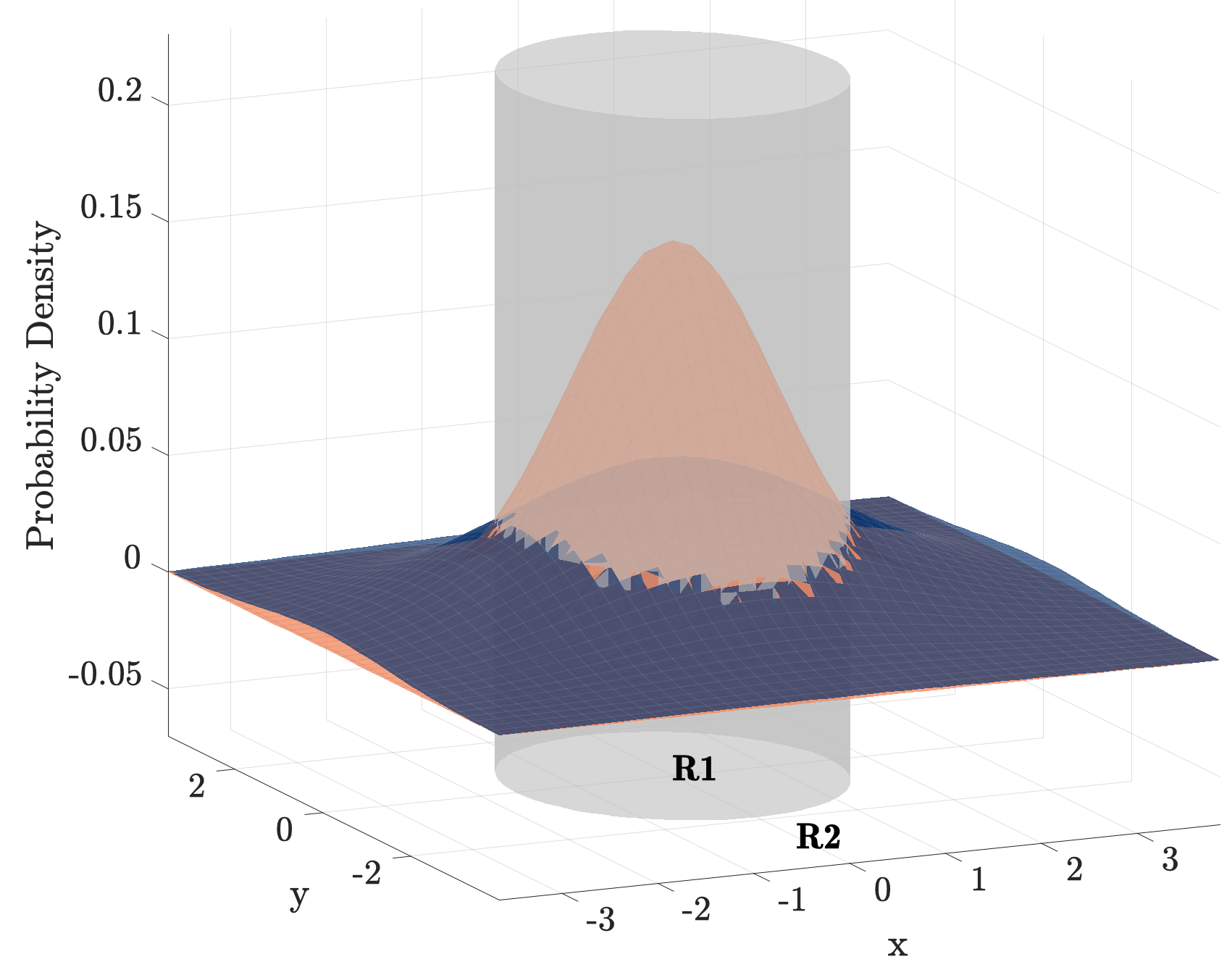}%
\label{exp7pdf}}
\hfil
\subfloat[Exp1.9 and Exp1.10 densities; Intersection surface equation\\ $\mathbf{S}: \frac{-x^{2}}{3}-x-\frac{y^{2}}{4}-\frac{y}{2}+\frac{\log(6)}{2}+\frac{7}{4}=0$]{\includegraphics[width=0.32\textwidth]{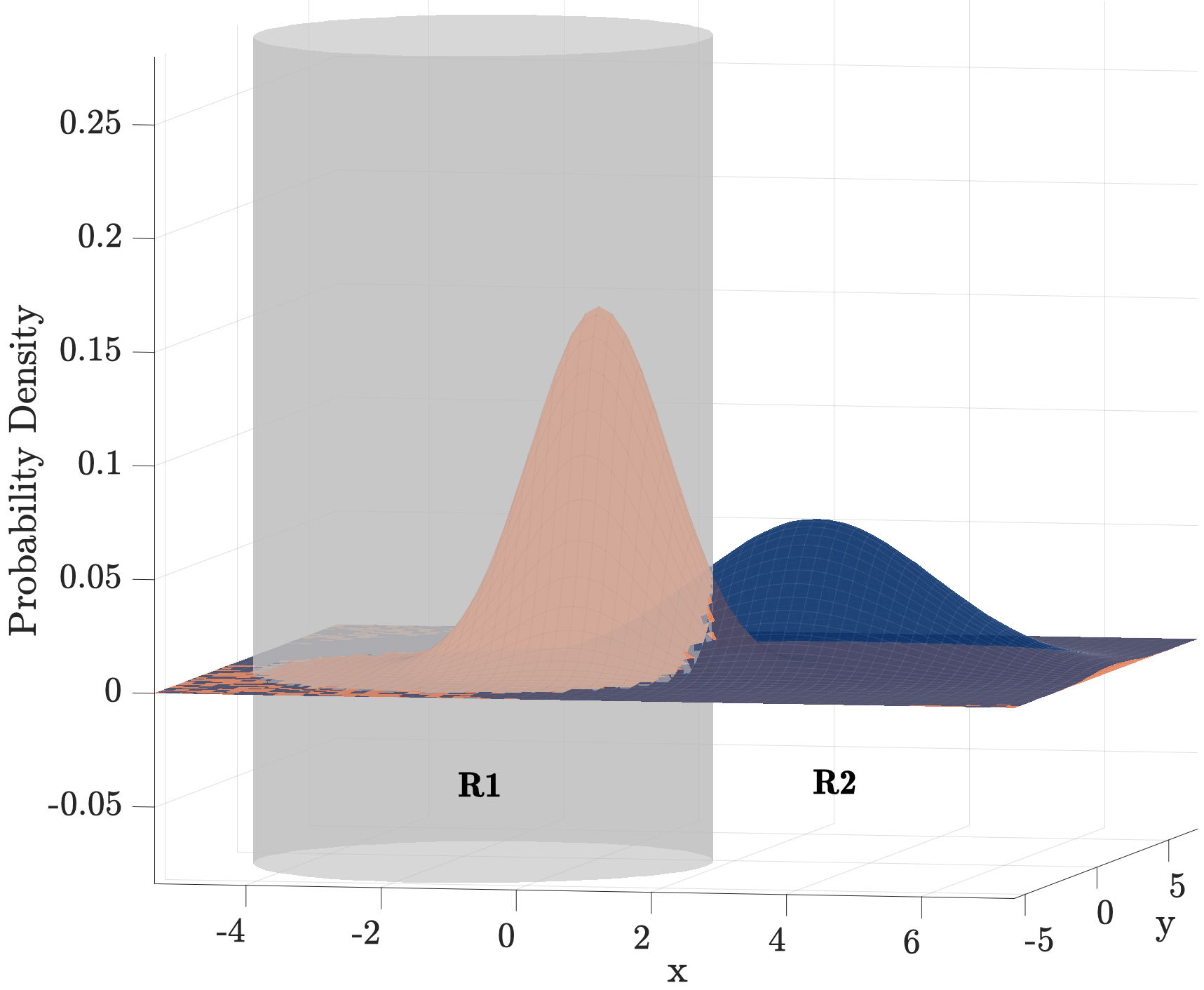}%
\label{exp9pdf}}
\hfil
\subfloat[Exp1.11 and Exp1.12 densities; Intersection surface equation\\ $\mathbf{S}:  \frac{-17x^{2}}{48}- \frac{xy}{24}- \frac{29x}{24}-  \frac{17y^{2}}{48}+  \frac{11y}{24}+\frac{\log(12)}{2}+\frac{127}{48}=0$]{\includegraphics[width=0.32\textwidth]{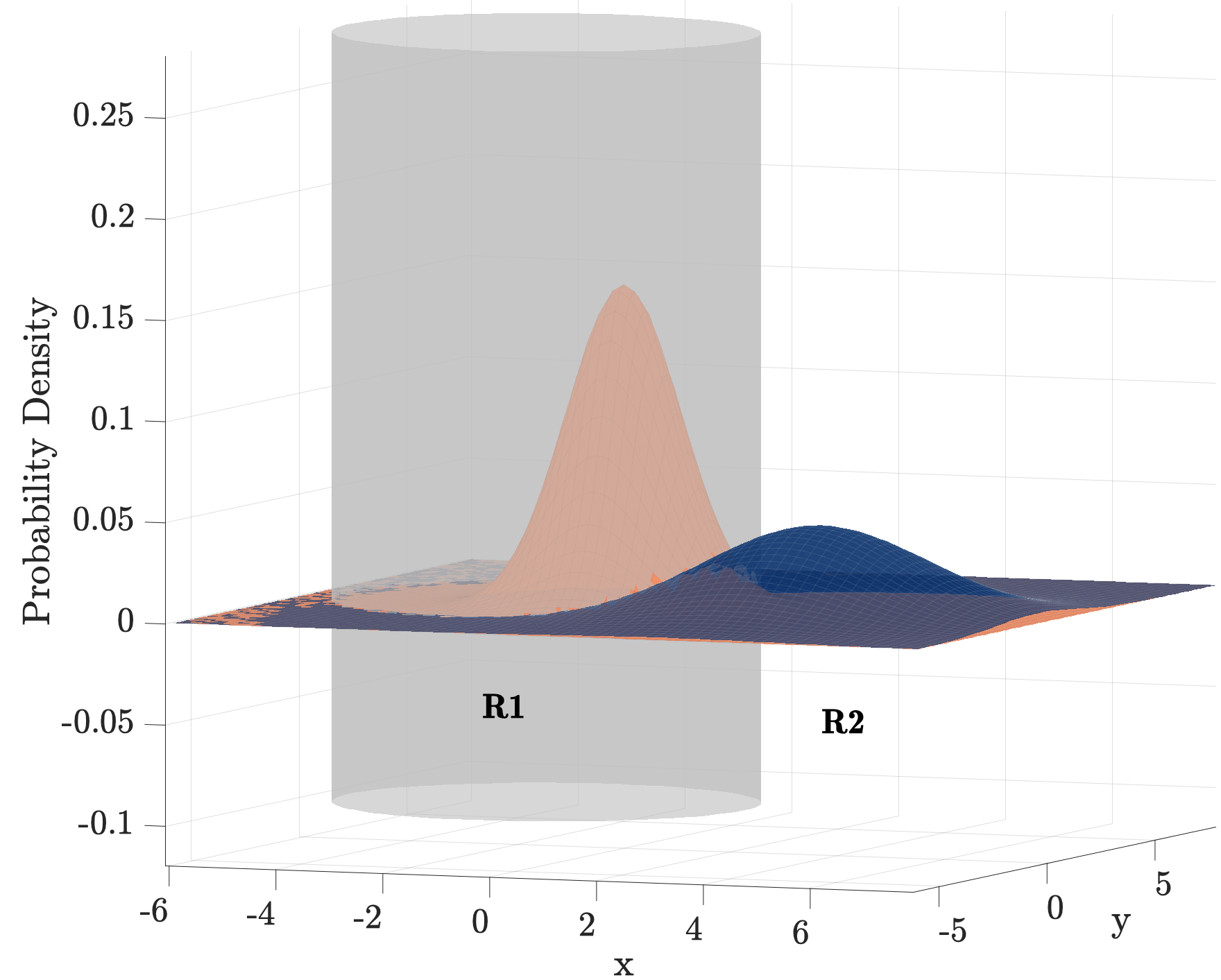}%
\label{exp11pdf}}
\caption{Training and test input densities decomposed by the \textbf{R1} and \textbf{R2} regions and the intersection surface.}
\label{exp_pdf}
\end{figure*}

Similarly, in the case of translation and scaling at the same time, the surface where the two pdfs intersect is a cylinder that is centered at $\left(\frac{\mu_{1}}{1-\sigma_{1}^{2}}, \frac{\mu_{2}}{1-\sigma_{2}^{2}}\right)$. Fig. \ref{exp_pdf} shows the densities of the training and test data along with the corresponding intersection surface between the two pdfs of the two-dimensional experiments.

\section{Empirical Results and Analysis}
\label{sec:analysis}
The results of the experiments detailed in Section \ref{sec:settings} will be scrutinized in this section to answer the research questions. Specifically, the overall results of applying the ML models to different datasets characterized by various types of drift will be presented. Additionally, a model-wise performance analysis will follow. To execute the experiments, the scikit-learn library\footnote{https://scikit-learn.org/}was utilized, a widely used open-source Python ML package that provides the implementation of several classification algorithms.

\subsection{Degradation Rate in Evaluation Metrics}
Changes in data distribution usually lead to a degradation of performance metrics \cite {Rabanser2019Failing}. To measure the performance loss after the drift and address \textbf{RQ1}, the degradation rate is calculated, this rate represents the percentage of the performance drop of the ML model on the drifted dataset. Table \ref{tab:2dresults} recapitulates the results of the two-dimensional experiments across the different performance metrics of the ML models. 

The first observation from the results is that the Random Forests algorithm demonstrated a high robustness level in most experiments across all performance metrics. This is an advantage of ensemble-based methods that can alleviate the biases of the datasets by combining individual classifiers' votes \cite{Clark2019Ensemble}. However, this is not the case in the four-dimensional experiment results summarized in Table \ref{tab:4dresults}. RF has shown much poorer robustness in higher-dimensional data. This could be attributable to the fact that learners which use more covariates to make predictions tend to show greater robustness than those that rely on a subset of covariates \cite{Abbasian2010Robustness}. The default value provided by the scikit-learn library was used to select the parameter that controls the subset of covariates used to find the best tree split in the Random Forests algorithm. The library assigns the square root value of the total covariates to the parameter. Furthermore, LR showed the highest degradation rate in most of the two-dimensional experiments, which is a recognized drawback of the algorithm in the case of out-of-sample predictions, where the maximum likelihood estimator tends to display poor performance due to the overfitting effect \cite{Shafieezadeh2015RobustLR}.

\begin{table*}
\fontsize{6}{6}\selectfont
\centering
\caption{Two-dimensional robustness evaluation results}
\label{tab:2dresults}
\setlength\tabcolsep{2pt}
\begin{tabular}{c|l|lll|lll|lll} 
\hline
\multicolumn{1}{l|}{\multirow{2}{*}{\textbf{Experiment}}} & \multirow{2}{*}{\begin{tabular}[c]{@{}l@{}}\textbf{ML }\\\textbf{model}\end{tabular}} & \multicolumn{3}{c|}{\textbf{Same distribution}} & \multicolumn{3}{c|}{\textbf{Drifted distribution}} & \multicolumn{3}{c}{\textbf{Degradation rate}}  \\ 
\multicolumn{1}{l|}{}                                     & \multicolumn{1}{c|}{}                                   & \textbf{Acc} & \textbf{F1} & \textbf{MCC}                & \textbf{Acc} & \textbf{F1} & \textbf{MCC}                   & \textbf{Acc} & \textbf{F1} & \textbf{MCC }   \\ 
\toprule
\multirow{5}{*}{\textbf{Exp1.1}}                           & \textbf{ SVM }                                          & 0.998         & 0.9978       & 0.9959                       & 0.9846        & 0.9847       & 0.9695                          & 1.34~\%       & 1.31~\%      & 2.65~\%          \\ 

                                                          & \textbf{ LR }                                           & 0.9768        & 0.975        & 0.9535                       & 0.8916        & 0.9014       & 0.8022                          & 8.72~\%       & 7.55~\%      & 15.87~\%         \\ 

                                                          & \textbf{ RF }                                                                  & 0.9998        & 0.9998       & 0.9997  & 0.9987        & 0.9986       & 0.9974                        &\textbf{0.11~\%*}       & \textbf{0.12~\%*}     & \textbf{0.23~\%*}          \\ 

                                                          & \textbf{ GNB }                                          & 0.971         & 0.9686       & 0.9417                       & 0.9523        & 0.9495       & 0.9087                          & 1.93~\%       & 1.97~\%      & 3.50~\%          \\ 

                                                          & \textbf{ KNN }                                          & 0.9978        & 0.9977       & 0.9957                       & 0.9634        & 0.9638       & 0.9274                          & 3.45~\%       & 3.40~\%      & 6.86~\%          \\
\midrule
\multirow{5}{*}{\textbf{Exp1.2}} & \textbf{ SVM } & 0.9976 & 0.9973 & 0.9951 & 0.9612 & 0.9765 & 0.8742 & 3.65~\% & 2.09~\% & 12.15~\%  \\
                                 & \textbf{ LR }  & 0.9786 & 0.9768 & 0.9569 & 0.9457 & 0.9688 & 0.7825 & 3.36~\% & 0.82~\% & 18.23~\%  \\
                                 & \textbf{ RF }  & 1.0    & 1.0    & 1.0    &0.9988 & 0.9988 & 0.9977 & \textbf{0.12~\%*} & \textbf{0.12~\%*} & \textbf{0.23~\%*}   \\
                                 & \textbf{ GNB } & 0.9711 & 0.9687 & 0.9419 & 0.9663 & 0.9796 & 0.8886 & 0.49~\% & 1.13~\% & 5.66~\%   \\
                                 & \textbf{ KNN } & 0.9973 & 0.9971 & 0.9946 & 0.9823 & 0.9896 & 0.9318 & 1.50~\% & 0.75~\% & 6.31~\%   \\
                                 
\midrule
\multirow{5}{*}{\textbf{Exp1.3}} & \textbf{ SVM } & 0.997  & 0.9967 & 0.9939 & 0.986  & 0.9832 & 0.9712 & 1.10~\% & 1.35~\% & 2.28~\%  \\
                                 & \textbf{ LR }  & 0.9768 & 0.9747 & 0.9533 & 0.955  & 0.9487 & 0.9107 & 2.23~\% & 2.67~\% & 4.47~\%  \\
                                 & \textbf{ RF }  & 0.9983 & 0.9981 & 0.9966 & 0.9931 & 0.9919 & 0.986  & \textbf{0.52~\%*} & \textbf{0.62~\%*} & \textbf{1.06~\%*}  \\
                                 & \textbf{ GNB } & 0.9655 & 0.9622 & 0.9305 & 0.9502 & 0.9415 & 0.8982 & 1.58~\% & 2.15~\% & 3.47~\%  \\
                                 & \textbf{ KNN } & 0.9962 & 0.9958 & 0.9922 & 0.9896 & 0.9878 & 0.9788 & 0.66~\% & 0.80~\% & 1.35~\%  \\
                                 
\midrule
\multirow{5}{*}{\textbf{Exp1.4}} & \textbf{ SVM } & 0.9966 & 0.9963 & 0.9932 & 0.9936 & 0.993  & 0.9872 & 0.30~\% & 0.33~\% & 0.60~\%  \\
                                 & \textbf{ LR }  & 0.9766 & 0.9747 & 0.953  & 0.9699 & 0.9675 & 0.9402 & 0.69~\% & 0.74~\% & 1.34~\%  \\
                                 & \textbf{ RF }  & 0.9983 & 0.9982 & 0.9966 & 0.997  & 0.9968 & 0.9941 & 0.13~\% & \textbf{0.14~\%*} & \textbf{0.25~\%*}  \\
                                 & \textbf{ GNB } & 0.97   & 0.9674 & 0.9397 & 0.9689 & 0.9657 & 0.9373 & \textbf{0.11~\%*} & 0.18~\% & 0.26~\%  \\
                                 & \textbf{ KNN } & 0.9965 & 0.9962 & 0.993  & 0.9922 & 0.9914 & 0.9844 & 0.43~\% & 0.48~\% & 0.87~\%  \\
\midrule
\multirow{5}{*}{\textbf{Exp1.5}} & \textbf{ SVM } & 0.9982 & 0.998  & 0.9964 & 0.872  & 0.9077 & 0.7375 & 12.64~\% & 9.05~\% & 25.98~\%  \\
                                 & \textbf{ LR }  & 0.9776 & 0.9756 & 0.9548 & 0.944  & 0.9643 & 0.8463 & 3.44~\%  & 1.16~\% & 11.36~\%  \\
                                 & \textbf{ RF }  & 0.9998 & 0.9999 & 0.9996 & 0.9988 & 0.9988 & 0.9977 &  \textbf{0.10~\%*}  & \textbf{0.11~\%*} & \textbf{0.19~\%*}   \\
                                 & \textbf{ GNB } & 0.9684 & 0.9656 & 0.9365 & 0.9539 & 0.9686 & 0.888  & 1.50~\%  & 0.31~\% & 5.18~\%   \\
                                 & \textbf{ KNN } & 0.9969 & 0.9966 & 0.9938 & 0.9906 & 0.9938 & 0.9744 & 0.63~\%  & 0.28~\% & 1.95~\%   \\
\midrule
\multirow{5}{*}{\textbf{Exp1.6}} & \textbf{ SVM } & 0.9978 & 0.9976 & 0.9956 & 0.8842 & 0.7538 & 0.7177 & 11.39~\% & 24.44~\% & 27.91~\%  \\
                                 & \textbf{ LR }  & 0.9752 & 0.9729 & 0.9501 & 0.9232 & 0.8834 & 0.8398 & 5.33~\%  & 9.20~\%  & 11.61~\%  \\
                                 & \textbf{ RF }  &  1.0  & 0.9999 & 0.9999 & 0.9984 & 0.9982 & 0.9967 & \textbf{0.16~\%*}  & \textbf{0.17~\%*}  & \textbf{0.32~\%*}   \\
                                 & \textbf{ GNB } & 0.9696 & 0.9667 & 0.9389 & 0.9532 & 0.9127 & 0.8874 & 1.69~\%  & 5.59~\%  & 5.49~\%   \\
                                 & \textbf{ KNN } & 0.9958 & 0.9954 & 0.9914 & 0.9868 & 0.9773 & 0.9681 & 0.90~\%  & 1.82~\%  & 2.35~\%   \\
\midrule
\multirow{5}{*}{\textbf{Exp1.7}} & \textbf{ SVM } & 0.9907 & 0.9902 & 0.9814 & 0.9478 & 0.9481 & 0.8956 & 4.33~\% & 4.25~\% & 8.74~\%   \\
                                 & \textbf{ LR }  & 0.8198 & 0.8036 & 0.6391 & 0.7532 & 0.7816 & 0.5241 & 8.12~\% & 2.74~\% & 17.99~\%  \\
                                 & \textbf{ RF }  & 0.9998 & 0.9998 & 0.9996 & 0.9956 & 0.9954 & 0.9913 &  \textbf{0.42~\%*} & \textbf{0.44~\%*} & \textbf{0.84~\%*}   \\
                                 & \textbf{ GNB } & 0.7645 & 0.7137 & 0.5408 & 0.7558 & 0.7197 & 0.5301 & 1.14~\% & 0.84~\% & 1.98~\%   \\
                                 & \textbf{ KNN } & 0.9906 & 0.9901 & 0.9812 & 0.9242 & 0.9254 & 0.8487 & 6.70~\% & 6.53~\% & 13.50~\%  \\
                                 
\midrule
\multirow{5}{*}{\textbf{Exp1.8}} & \textbf{ SVM } & 0.9903 & 0.9899 & 0.9806 & 0.9325 & 0.94   & 0.8636 & 5.84~\%  & 5.04~\% & 11.93~\%  \\
                                 & \textbf{ LR }  & 0.8231 & 0.8046 & 0.6474 & 0.6383 & 0.7581 & 0.2909 & 22.45~\% & 5.78~\% & 55.07~\%  \\
                                 & \textbf{ RF }  & 0.9952 & 0.995  & 0.9905 & 0.9868 & 0.9884 & 0.9735 & \textbf{0.84~\%*}  & \textbf{0.66~\%*} & \textbf{1.72~\%*}   \\
                                 & \textbf{ GNB } & 0.769  & 0.7192 & 0.5524 & 0.7049 & 0.7885 & 0.418  & 8.34~\%  & 9.64~\% & 24.33~\%  \\
                                 & \textbf{ KNN } & 0.9918 & 0.9914 & 0.9835 & 0.8882 & 0.9072 & 0.7734 & 10.45~\% & 8.49~\% & 21.36~\%  \\
\midrule
\multirow{5}{*}{\textbf{Exp1.9}} & \textbf{ SVM } & 0.99   & 0.9894 & 0.9798 & 0.9698 & 0.9683 & 0.9394 & 2.04~\% & 2.13~\%  & 4.12~\%   \\
                                 & \textbf{ LR }  & 0.8188 & 0.7997 & 0.6375 & 0.7405 & 0.714  & 0.4803 & 9.56~\% & 10.72~\% & 24.66~\%  \\
                                 & \textbf{ RF }  & 0.9942 & 0.9939 & 0.9884 & 0.9805 & 0.9796 & 0.961  & \textbf{1.38~\%*} & \textbf{1.44~\%*}  & \textbf{2.77~\%*}   \\
                                 & \textbf{ GNB } & 0.7664 & 0.7138 & 0.5449 & 0.7264 & 0.66   & 0.4683 & 5.22~\% & 7.54~\%  & 14.06~\%  \\
                                 & \textbf{ KNN } & 0.9916 & 0.9912 & 0.9833 & 0.9625 & 0.9607 & 0.9249 & 2.93~\% & 3.08~\%  & 5.94~\%   \\
\midrule
\multirow{5}{*}{\textbf{Exp1.10}} & \textbf{ SVM } & 0.9906 & 0.99   & 0.9811 & 0.9674 & 0.9671 & 0.935  & 2.34~\%  & 2.31~\%  & 4.70~\%   \\
                                 & \textbf{ LR }  & 0.8198 & 0.7985 & 0.6392 & 0.6716 & 0.6548 & 0.3445 & 18.08~\% & 18.00~\% & 46.10~\%  \\
                                 & \textbf{ RF }  & 0.9954 & 0.9952 & 0.9909 & 0.9792 & 0.9791 & 0.9584 & \textbf{1.63~\%*}  & \textbf{1.62~\%*}  & \textbf{3.28~\%*}   \\
                                 & \textbf{ GNB } & 0.7677 & 0.7119 & 0.547  & 0.6486 & 0.6106 & 0.3023 & 15.51~\% & 14.23~\% & 44.73~\%  \\
                                 & \textbf{ KNN } & 0.9918 & 0.9912 & 0.9834 & 0.9618 & 0.9618 & 0.9236 & 3.02~\%  & 2.97~\%  & 6.08~\%   \\
\midrule
\multirow{5}{*}{\textbf{Exp1.11}} & \textbf{ SVM } & 0.9902 & 0.9898 & 0.9805 & 0.849  & 0.8452 & 0.7006 & 14.26~\% & 14.61~\% & 28.55~\%  \\
                                 & \textbf{ LR }  & 0.822  & 0.8062 & 0.6434 & 0.6114 & 0.7094 & 0.2809 & 25.62~\% & 12.01~\% & 56.34~\%  \\
                                 & \textbf{ RF }  & 0.9952 & 0.995  & 0.9905 & 0.993  & 0.9931 & 0.986  & \textbf{0.22~\%*}  & \textbf{0.19~\%*}  & \textbf{0.45~\%*}   \\
                                 & \textbf{ GNB } & 0.7721 & 0.7315 & 0.5559 & 0.6644 & 0.7242 & 0.366  & 13.95~\% & 1.01~\%  & 34.16~\%  \\
                                 & \textbf{ KNN } & 0.9916 & 0.9912 & 0.9832 & 0.8474 & 0.853  & 0.6951 & 14.54~\% & 13.94~\% & 29.30~\%  \\
\midrule
\multirow{5}{*}{\textbf{Exp1.12}} & \textbf{ SVM } & 0.9909 & 0.9904 & 0.9817 & 0.7302 & 0.6574 & 0.5161  & 26.31~\% & 33.62~\% & 47.43~\%   \\
                                 & \textbf{ LR }  & 0.8208 & 0.7998 & 0.6421 & 0.5327 & 0.4914 & 0.0681  & 35.10~\% & 38.56~\% & 89.39~\%   \\
                                 & \textbf{ RF }  & 0.9958 & 0.9955 & 0.9915 & 0.9752 & 0.9748 & 0.9512  & \textbf{2.07~\%*}  & \textbf{2.08~\%*}  & \textbf{4.06~\%*}    \\
                                 & \textbf{ GNB } & 0.7709 & 0.7183 & 0.5548 & 0.4958 & 0.3764 & 0.0051 & 35.69~\% & 47.60~\% & 100.92~\%  \\
                                 & \textbf{ KNN } & 0.9916 & 0.9912 & 0.9833 & 0.728  & 0.7185 & 0.458   & 26.58~\% & 27.51~\% & 53.42~\%   \\
\bottomrule
\end{tabular}
\end{table*}

\begin{table*}
\fontsize{6}{6}\selectfont

\centering
\caption{Four-dimensional robustness evaluation results}
\label{tab:4dresults}
\setlength\tabcolsep{2pt}
\begin{tabular}{c|l|lll|lll|lll} 
\hline
\multicolumn{1}{l|}{\multirow{2}{*}{\textbf{Experiment}}} & \multirow{2}{*}{\begin{tabular}[c]{@{}l@{}}\textbf{ML }\\\textbf{model}\end{tabular}} & \multicolumn{3}{c|}{\textbf{Same distribution}} & \multicolumn{3}{c|}{\textbf{Drifted distribution}} & \multicolumn{3}{c}{\textbf{Degradation rate}}  \\ 
\multicolumn{1}{l|}{}                                     & \multicolumn{1}{c|}{}                                   & \textbf{Acc} & \textbf{F1} & \textbf{MCC}                & \textbf{Acc} & \textbf{F1} & \textbf{MCC}                   & \textbf{Acc} & \textbf{F1} & \textbf{MCC }   \\
\toprule
\multirow{5}{*}{\textbf{Exp2.1}} & \textbf{ SVM } & 0.993  & 0.9923 & 0.9859 & 0.9739 & 0.9812 & 0.9385 & 1.92~\% & 1.12~\% & 4.81~\%   \\
                                 & \textbf{ LR }  & 0.9691 & 0.9714 & 0.9795 & 0.9662 & 0.9377 & 0.9327 & \textbf{0.24~\%*} & 1.38~\% & \textbf{0.53~\%* }  \\
                                 & \textbf{ RF }  & 0.9768 & 0.9746 & 0.9533 & 0.9337 & 0.9517 & 0.8471 & 4.41~\% & 2.35~\% & 11.14~\%  \\
                                 & \textbf{ GNB } & 0.9641 & 0.9599 & 0.9281 & 0.92   & 0.9391 & 0.8368 & 4.57~\% & 2.17~\% & 9.84~\%   \\
                                 & \textbf{ KNN } & 0.974  & 0.9716 & 0.9477 & 0.9592 & 0.9706 & 0.9038 & 1.52~\% & \textbf{0.10~\%*} & 4.63~\%   \\
\midrule
\multirow{5}{*}{\textbf{Exp2.2}} & \textbf{ SVM } & 0.7518 & 0.7449 & 0.5038  & 0.6056 & 0.6132 & 0.2113  & 19.45~\% & 17.68~\% & 58.06~\%    \\
                                 & \textbf{ LR }  & 0.5002 & 0.6346 & 0.0053  & 0.4982 & 0.4689 & 0.0004 & \textbf{0.40~\%*}  & 35.34~\% & 92.45~\%  \\
                                 & \textbf{ RF }  & 0.9166 & 0.9158 & 0.8333  & 0.7844 & 0.7811 & 0.5692  & 14.42~\% & 14.71~\% & \textbf{31.69~\%*}    \\
                                 & \textbf{ GNB } & 0.501  & 0.5155 & 0.0091  & 0.4952 & 0.5124 & 0.002   & 1.17~\% & \textbf{ 0.60~\%*}  & 78.02~\%   \\
                                 & \textbf{ KNN } & 0.904  & 0.9029 & 0.808   & 0.7756 & 0.77   & 0.5518  & 14.20~\% & 14.72~\% & 31.71~\%    \\
                                 
\midrule
\multirow{5}{*}{\textbf{Exp2.3}} & \textbf{ SVM } & 0.9938 & 0.993  & 0.9874 & 0.9818 & 0.9793 & 0.9632 & 1.21~\% & 1.38~\% & 2.45~\%  \\
                                 & \textbf{ LR }  & 0.9704 & 0.9668 & 0.9401 & 0.9627 & 0.9591 & 0.926  & 0.79~\% & 0.80~\% & 1.50~\%  \\
                                 & \textbf{ RF }  & 0.9796 & 0.9772 & 0.9587 & 0.9652 & 0.961  & 0.9296 & 1.47~\% & 1.66~\% & 3.04~\%  \\
                                 & \textbf{ GNB } & 0.9708 & 0.9666 & 0.9411 & 0.9674 & 0.9626 & 0.9341 & \textbf{0.35~\%*} & \textbf{0.41~\%*} & \textbf{0.74~\%*}  \\
                                 & \textbf{ KNN } & 0.9768 & 0.974  & 0.953  & 0.969  & 0.9653 & 0.9374 & 0.80~\% & 0.89~\% & 1.64~\%  \\
                                 
\midrule
\multirow{5}{*}{\textbf{Exp2.4}} & \textbf{ SVM } & 0.7512 & 0.7468 & 0.5026  & 0.5994 & 0.5875 & 0.1987 & 20.21~\% & 21.33~\% & 60.47~\%   \\
                                 & \textbf{ LR }  & 0.5011 & 0.4817 & 0.0038  & 0.4983 & 0.4734 & 0.0017 & 0.56~\%  & \textbf{1.75~\%*}  & 55.26~\%  \\
                                 & \textbf{ RF }  & 0.9216 & 0.9211 & 0.8433  & 0.7838 & 0.7821 & 0.5676 & 14.95~\% & 15.09~\% & 32.69~\%   \\
                                 & \textbf{ GNB } & 0.5042 & 0.5182 & 0.0088  & 0.5017 & 0.4014 & 0.0013 & \textbf{0.50~\%*}  & 22.54~\% & 85.23~\%   \\
                                 & \textbf{ KNN } & 0.9008 & 0.9001 & 0.8015  & 0.7741 & 0.7731 & 0.5482 & 14.07~\% & 14.11~\% & \textbf{31.60~\%*}   \\
\midrule
\multirow{5}{*}{\textbf{Exp2.5}} & \textbf{ SVM } & 0.9935 & 0.9927 & 0.9869 & 0.9174 & 0.9285 & 0.8411 & 7.66~\% & 6.47~\% & 14.77~\%  \\
                                 & \textbf{ LR }  & 0.9698 & 0.9737 & 0.9389 & 0.9673 & 0.9661 & 0.9312 & \textbf{0.26~\%*} & 0.79~\% & \textbf{0.82~\%*}   \\
                                 & \textbf{ RF }  & 0.976  & 0.9731 & 0.9514 & 0.9301 & 0.9413 & 0.8575 & 4.70~\% & 3.27~\% & 9.87~\%   \\
                                 & \textbf{ GNB } & 0.9686 & 0.9642 & 0.9369 & 0.9445 & 0.9527 & 0.8914 & 2.49~\% & 1.19~\% & 4.86~\%   \\
                                 & \textbf{ KNN } & 0.974  & 0.9709 & 0.9473 & 0.959  & 0.9669 & 0.9133 & 1.54~\% & \textbf{0.41~\%*} & 3.59~\%   \\
\midrule
\multirow{5}{*}{\textbf{Exp2.6}} & \textbf{ SVM } & 0.758  & 0.7558 & 0.5161  & 0.5383 & 0.5558 & 0.077  & 28.98~\% & 26.46~\% & 85.08~\%   \\
                                 & \textbf{ LR }  & 0.5025 & 0.5234 & 0.0048  & 0.498 & 0.2368 & 0.0037 & 0.90~\%  & 54.76~\% & \textbf{22.92~\%*}  \\
                                 & \textbf{ RF }  & 0.92   & 0.92   & 0.8401  & 0.6544 & 0.6663 & 0.3099 & 28.87~\% & 27.58~\% & 63.11~\%   \\
                                 & \textbf{ GNB } & 0.5012 & 0.51   & 0.006   &  0.497  & 0.4913  & 0.0026 & \textbf{0.85~\%*}  & \textbf{3.81~\%*}  & 56.67~\%  \\
                                 & \textbf{ KNN } & 0.9014 & 0.9015 & 0.8029  & 0.6662 & 0.6837 & 0.3348 & 26.09~\% & 24.16~\% & 58.30~\%   \\

\bottomrule
\end{tabular}
\end{table*}

For the rest of the models, there are no precise general conclusions that can be deduced, since each model's degradation rates were affected by the settings of the experiments. It can be seen from Table \ref{tab:2dresults} that SVM is the model with the second lowest degradation rate after RF in most cases where multiple drifts were simulated, that is, in experiments \textbf{Exp1.9} to \textbf{Exp1.12}. On the other hand, GNB and KNN are highly affected by multiple drifts since $p(x)$ has been altered to a great extent. Each model's performance will be further investigated in the following subsection.

In terms of the effects of the complexity of the decision surface, it can be seen that the performance on the test data is worse and the degradation rates are higher in the four-dimensional experiments than in the two-dimensional cases when using the more complex decision function $F_2$, except for the LR and GNB algorithms, as they display low degradation rates due to poor performance on the original dataset. This can be associated with the positive correlation between the complexity of the decision boundary and the dimensional space of the problem \cite{Atashpaz2013complexity}. Furthermore, the effect of the complexity of the decision boundary is more discernible in experiments with mixed drifts, i.e., a higher magnitude of drift. 

\begin{figure}
    \centering
    \includegraphics[width=0.3\textwidth]{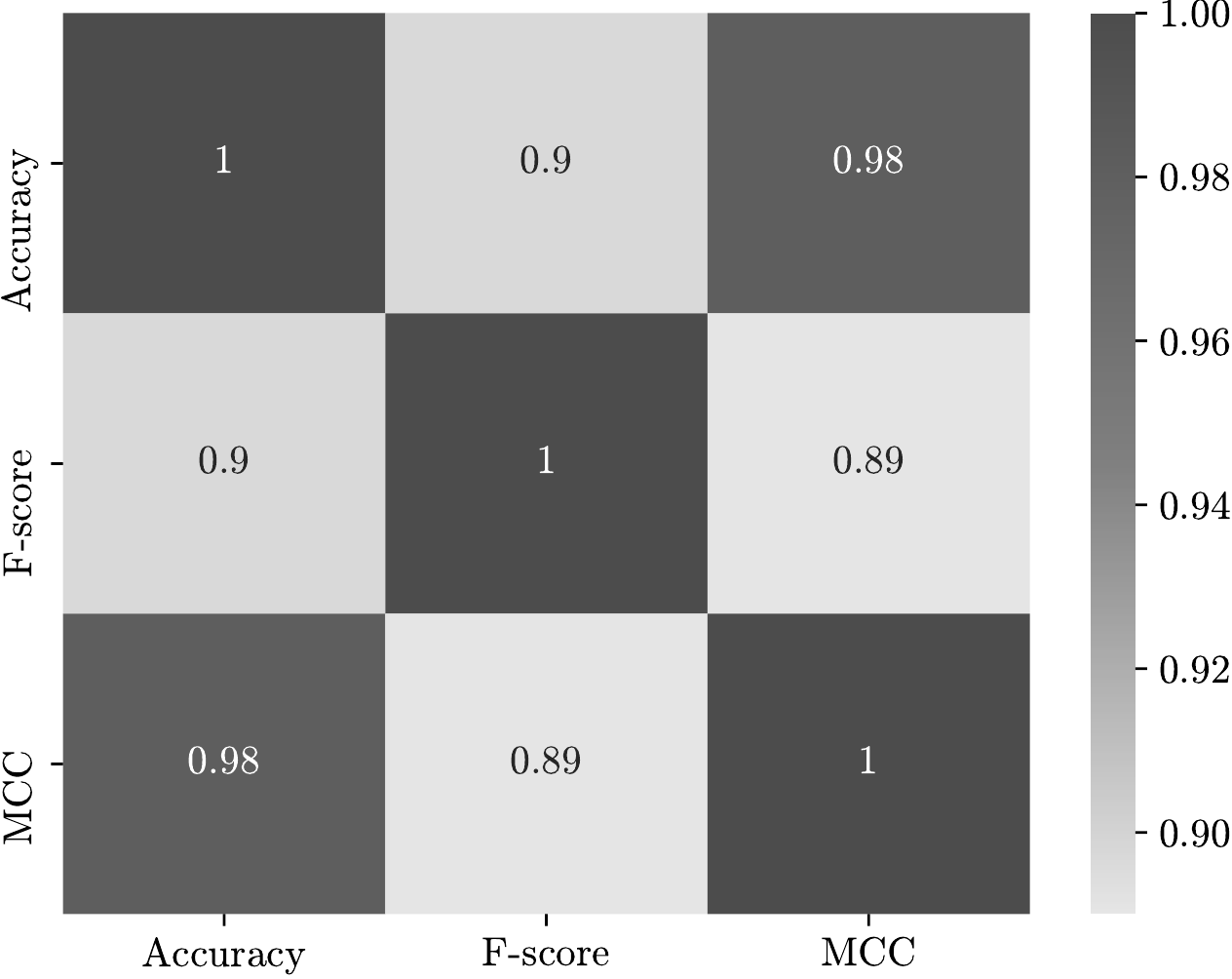}
    \caption{Correlation between degradation rates of performance metrics in the two-dimensional experiments}
    \label{fig:2dcorr}
\end{figure}

\begin{figure}
    \centering
    \includegraphics[width=0.3\textwidth]{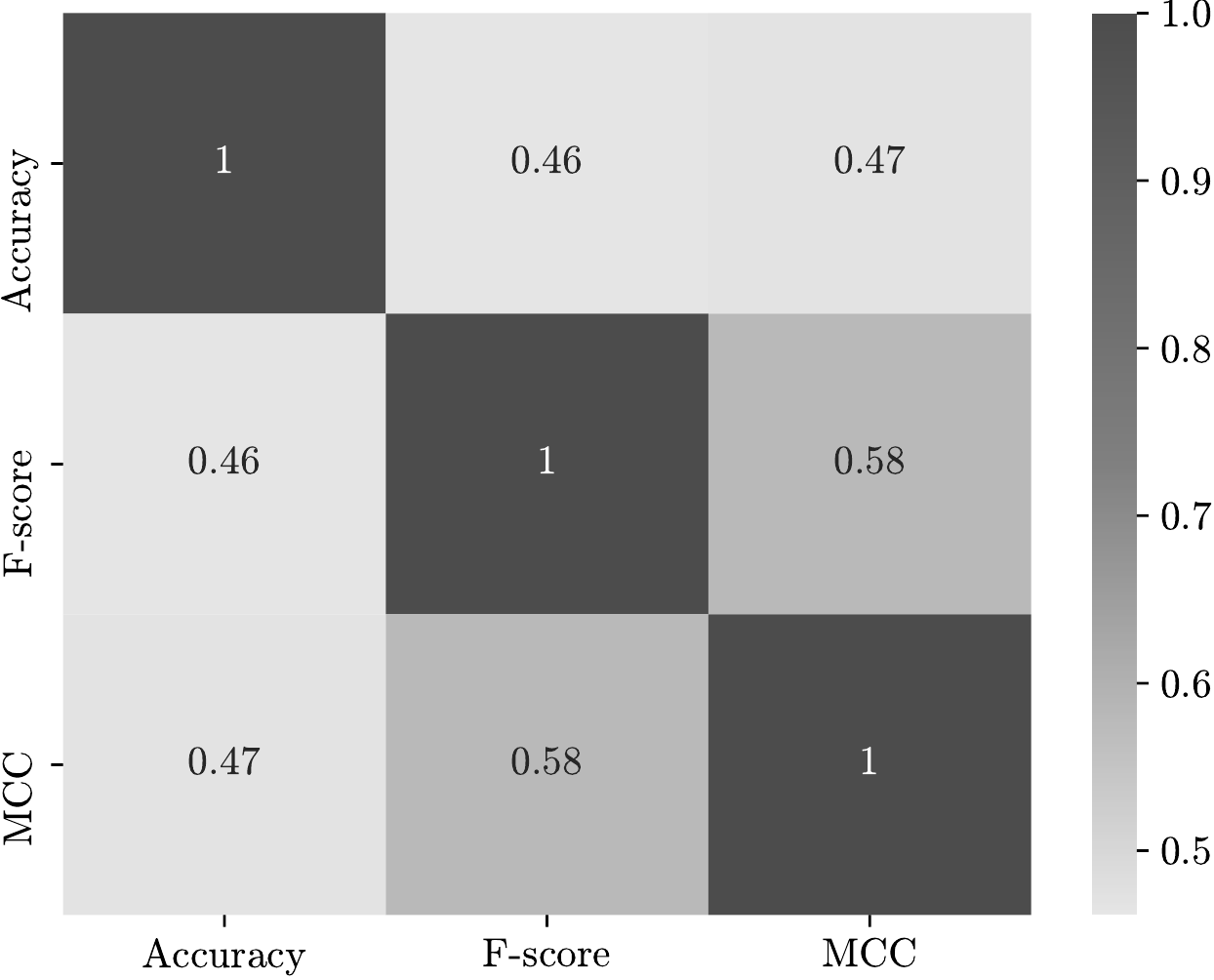}
    \caption{Correlation between degradation rates of performance metrics in the four-dimensional experiments}
    \label{fig:4dcorr}
\end{figure}

To see how the performance metrics are acting in the experiments, the correlation matrices of the two- and four-dimensional experiments between the degradation rates of different performance metrics used to evaluate the models have been plotted. The correlation matrices are shown in Fig. \ref{fig:2dcorr} and Fig. \ref{fig:4dcorr}. From the matrices it can be seen that a strong correlation between the degradation rates of the performance metrics in the two-dimensional experiments (see Fig. \ref{fig:2dcorr}). This indicates that the performance has been stirred in the same direction across the different metrics. However, this correlation between the degradation rates is lower in the four-dimensional experiments (see Fig. \ref{fig:4dcorr}). Furthermore, by comparing the degradation rates in Table \ref{tab:2dresults} and Table \ref{tab:4dresults}, it can be seen that the MCC metric is the most affected by changes, showing higher rates than accuracy and F-score. However, accuracy and F-score have shown similar degradation rates in most experiments.

\subsection{Model-Wise Analysis}
\label{sec:modelwise}
\begin{figure}
\centering
\subfloat[SVM]{\includegraphics[width=0.42\textwidth]{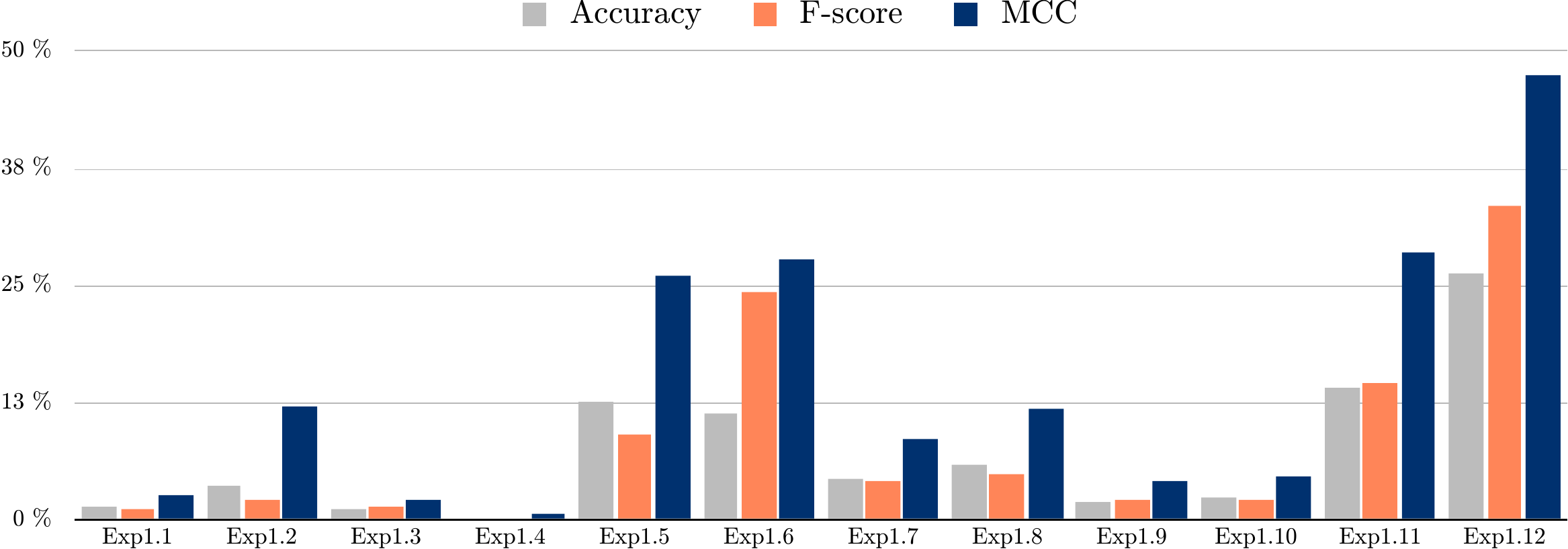}%
\label{svm_2d}}
\hfil
\subfloat[LR]{\includegraphics[width=0.42\textwidth]{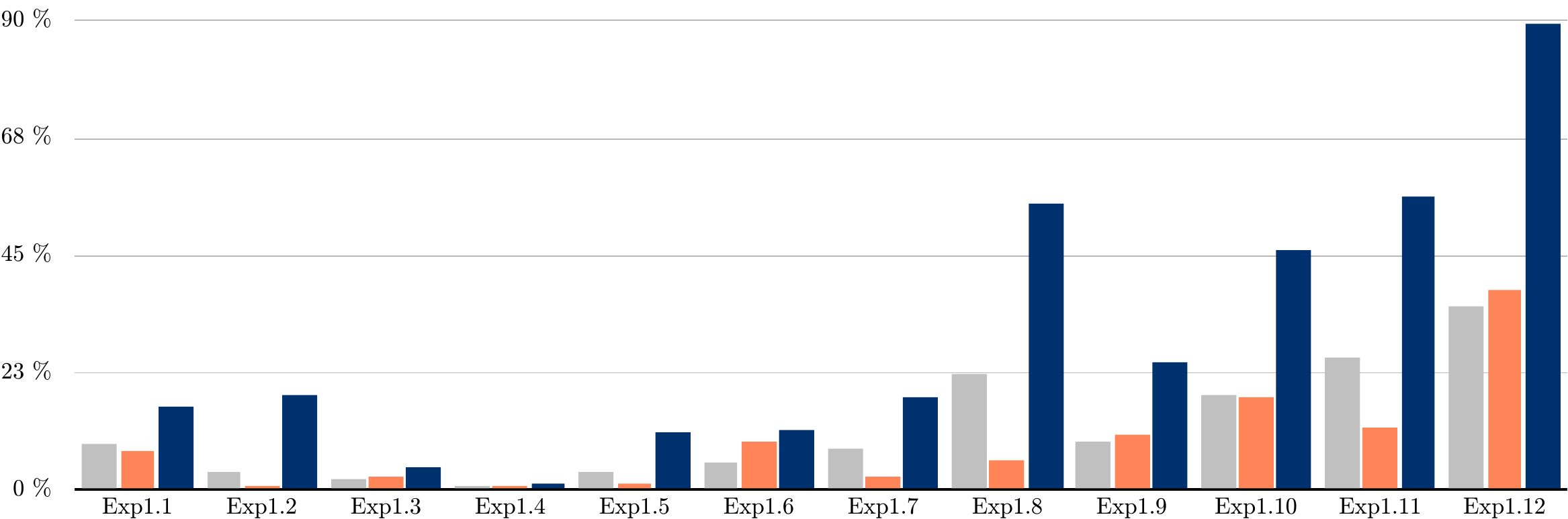}%
\label{lr_2d}}
\hfil
\subfloat[RF]{\includegraphics[width=0.42\textwidth]{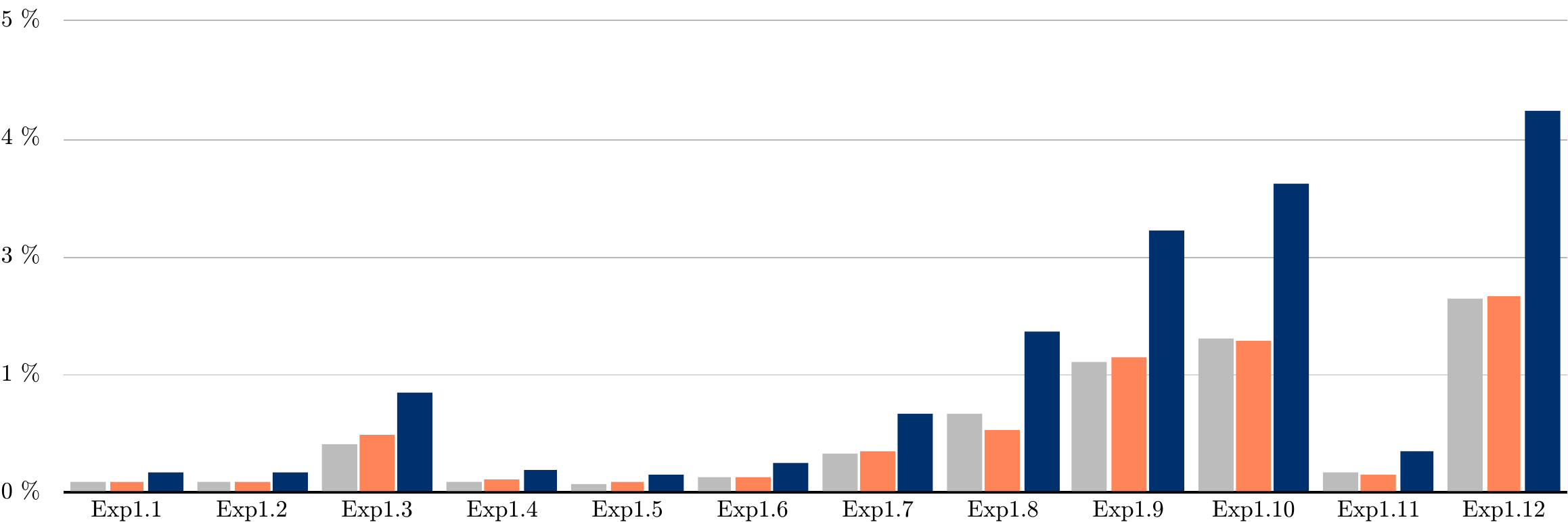}%
\label{rf_2d}}
\hfil
\subfloat[GNB]{\includegraphics[width=0.42\textwidth]{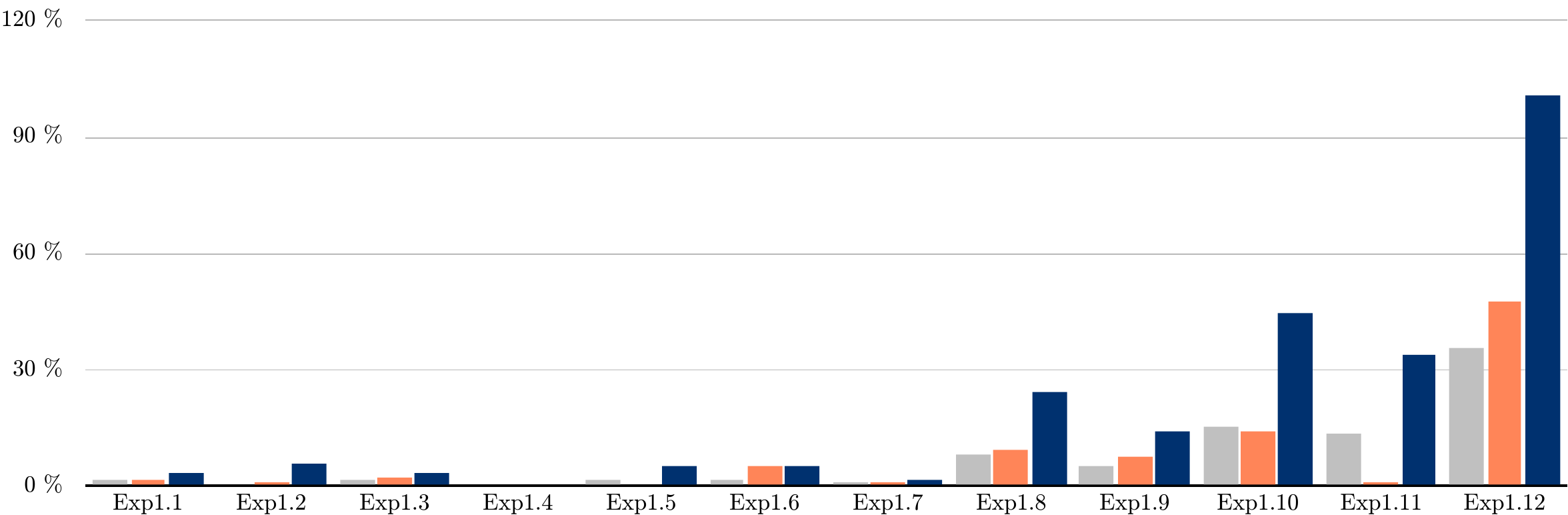}%
\label{gnb_2d}}
\hfil
\subfloat[KNN]{\includegraphics[width=0.42\textwidth]{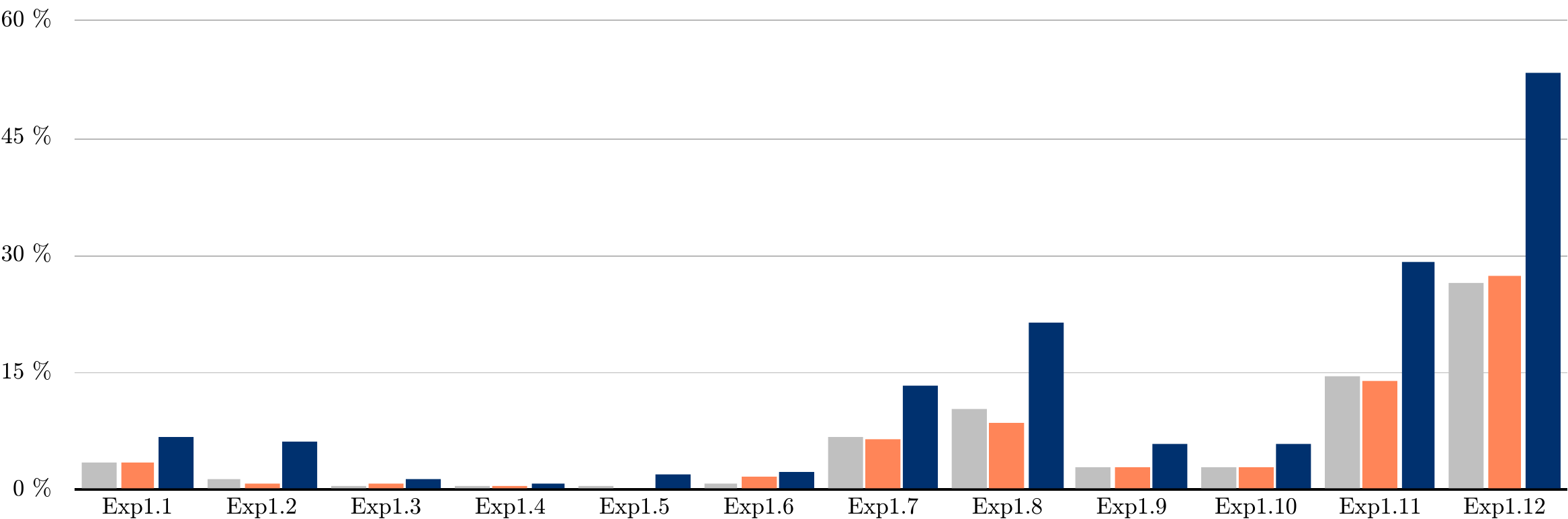}%
\label{knn_2d}}
\caption{Model-wise performance evaluation in two-dimensional experiments.}
\label{models_2d}
\end{figure}

\begin{figure}
\centering
\subfloat[SVM]{\includegraphics[width=0.23\textwidth]{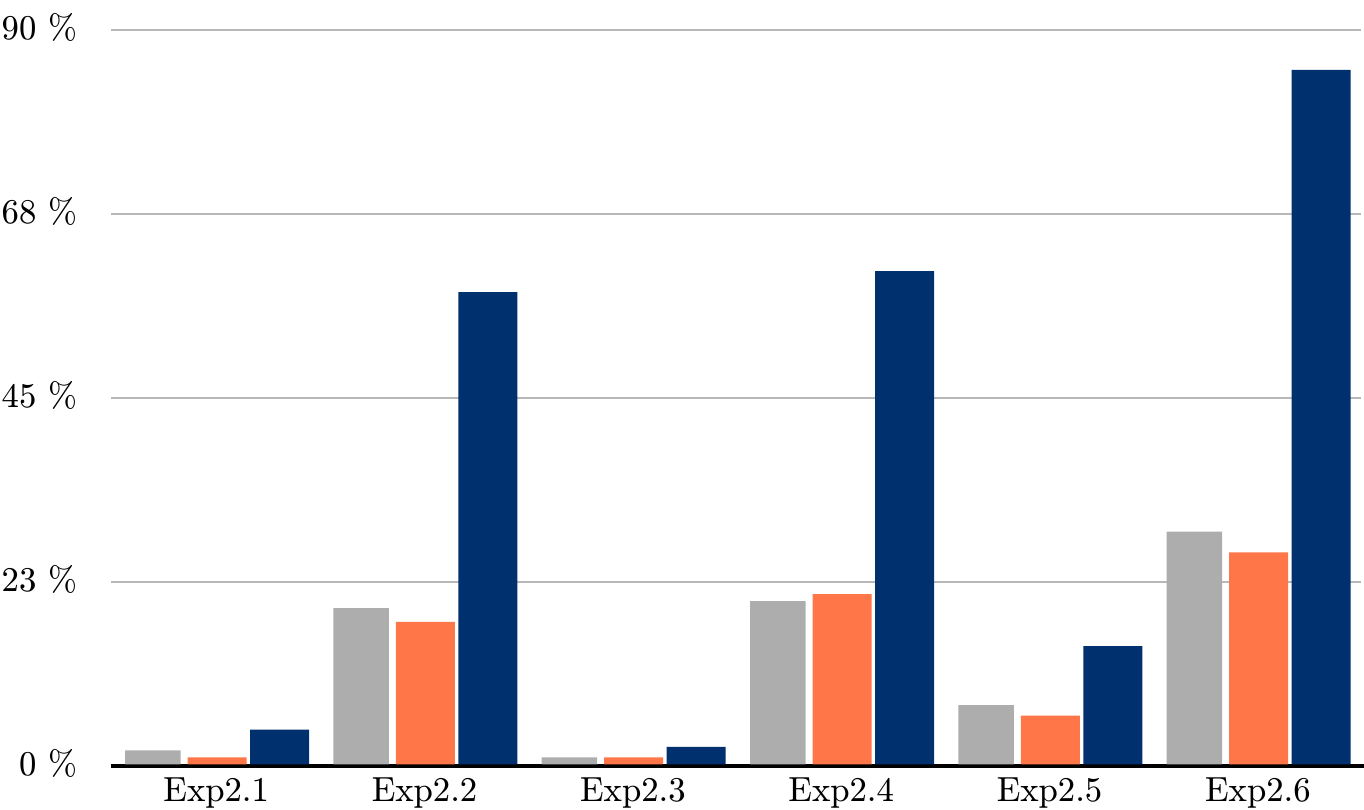}%
\label{svm_4d}}
\hfil
\subfloat[LR]{\includegraphics[width=0.23\textwidth]{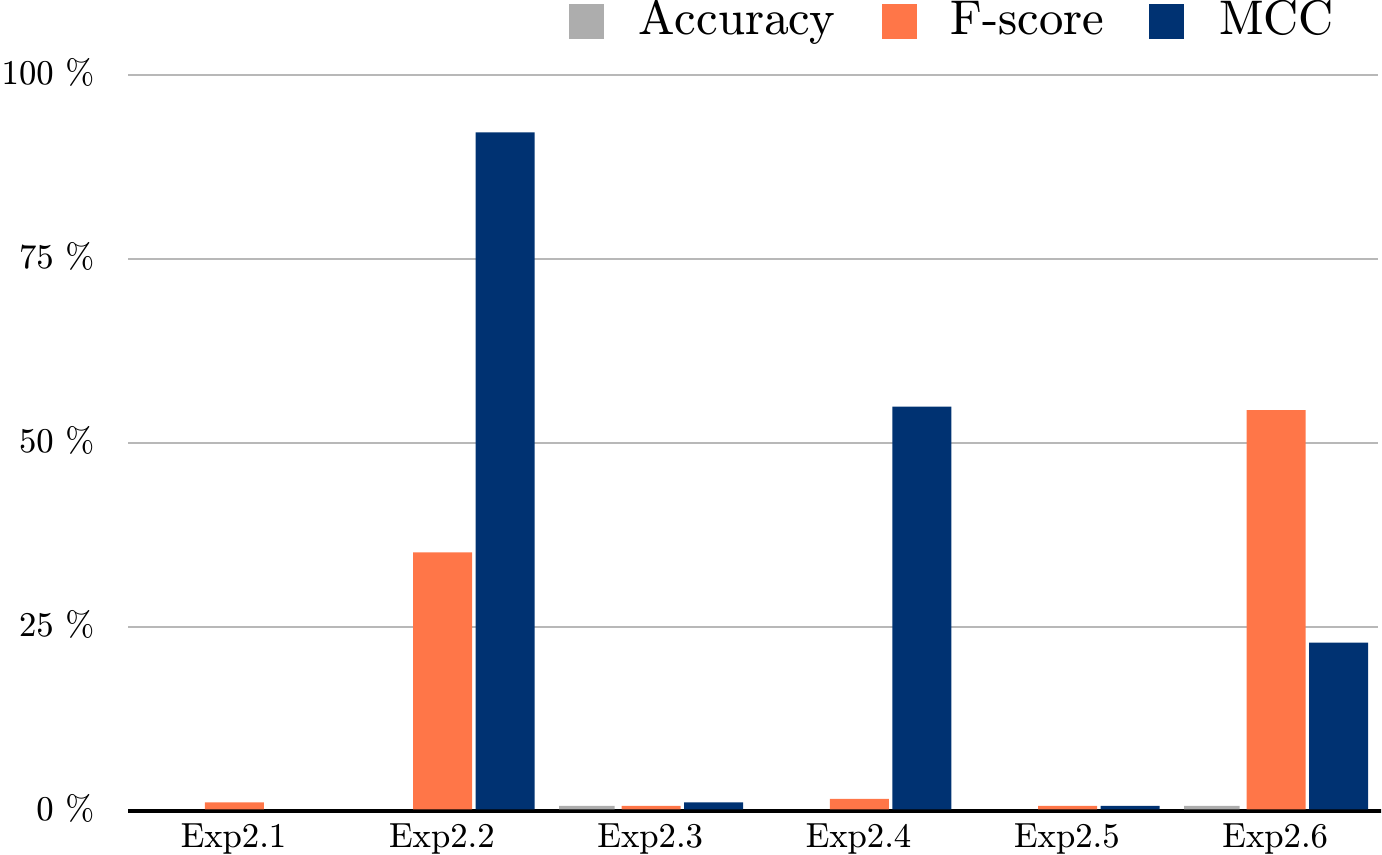}%
\label{lr_4d}}
\hfil
\subfloat[RF]{\includegraphics[width=0.23\textwidth]{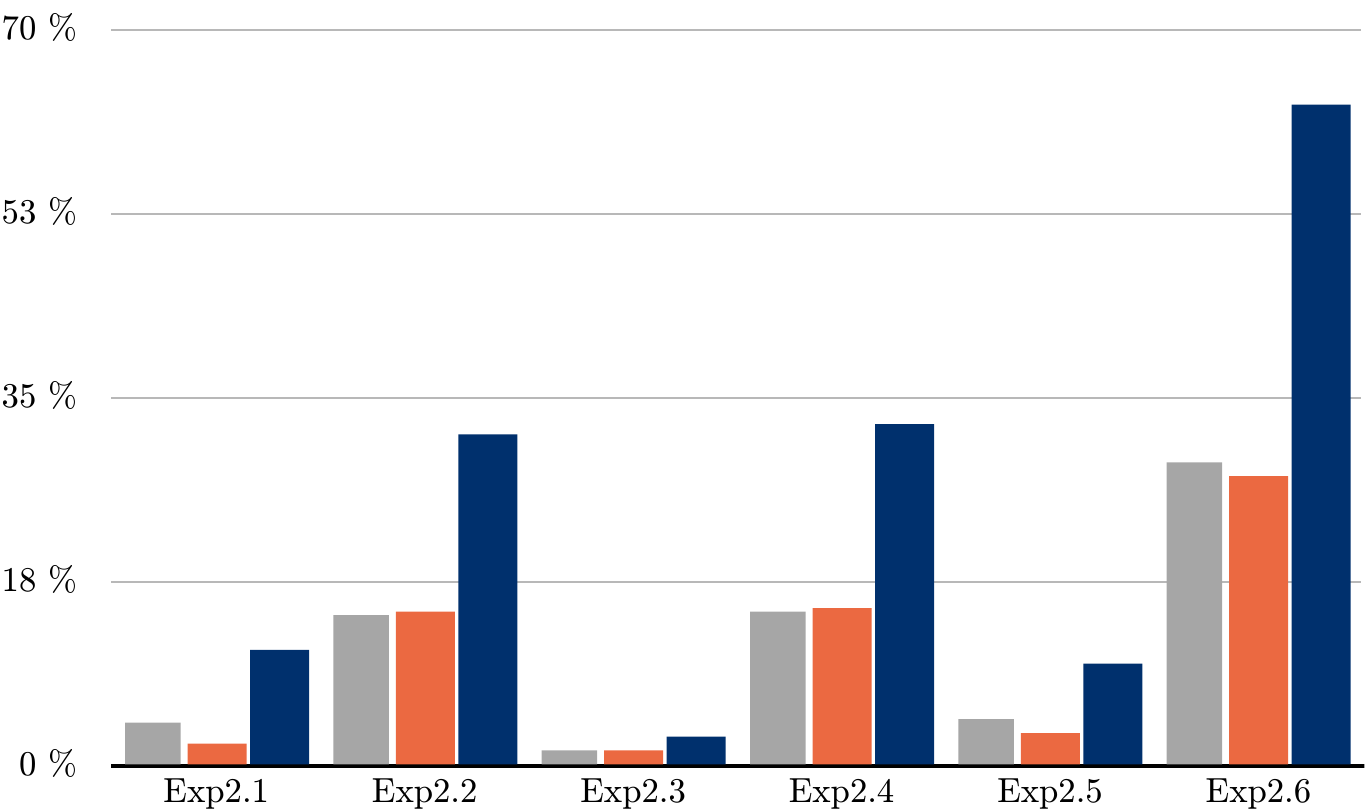}%
\label{rf_4d}}
\hfil
\subfloat[GNB]{\includegraphics[width=0.23\textwidth]{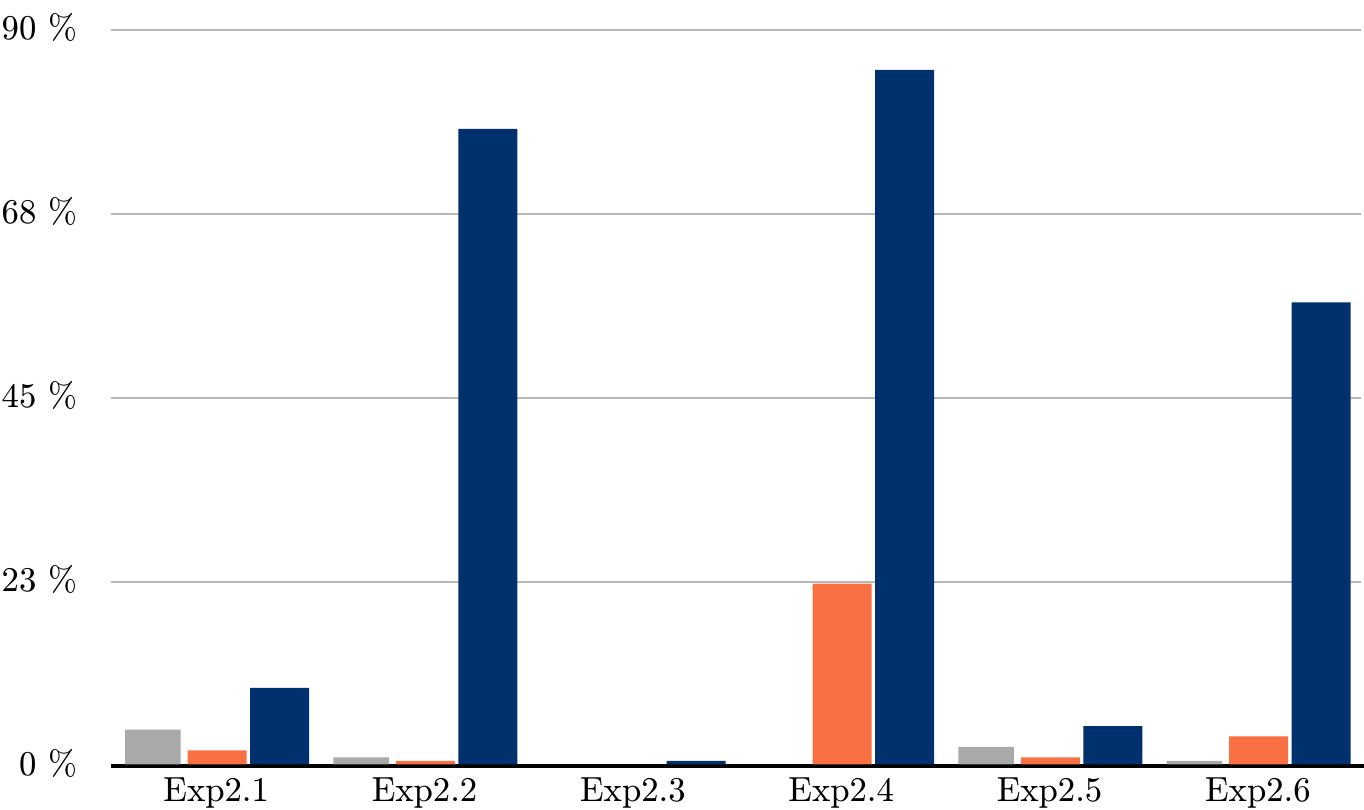}%
\label{gnb_4d}}
\hfil
\subfloat[KNN]{\includegraphics[width=0.2\textwidth]{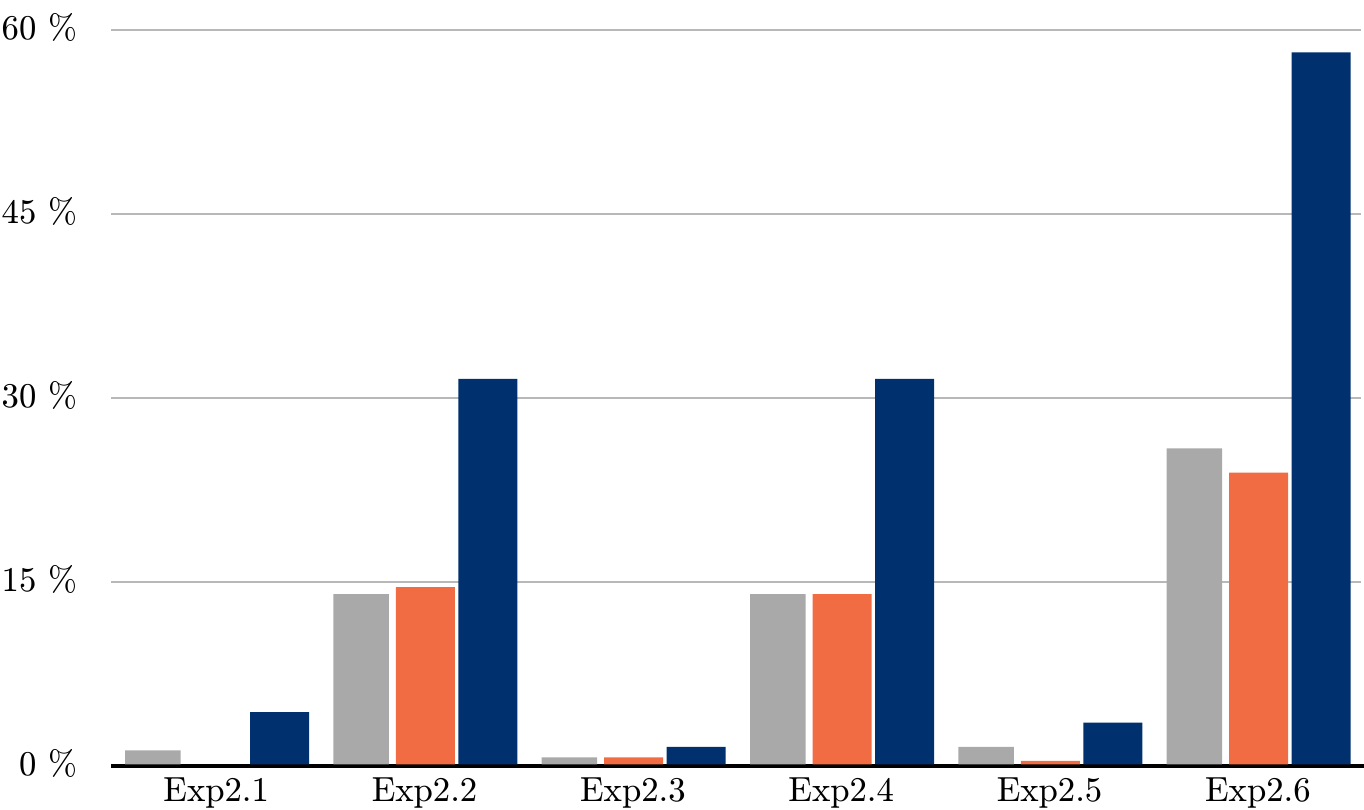}%
\label{knn_4d}}
\caption{Model-wise performance evaluation in four-dimensional experiments.}
\label{models_4d}
\end{figure}

In this section, the ML model's performance in the different experimental settings to address \textbf{RQ2} is scrutinized. The model-wise results of the two-dimensional experiments are organized in Fig. \ref{models_2d}, while Fig. \ref{models_4d} details the results of the four-dimensional experiments. The common ground between all ML models in the two-dimensional datasets is that they are mostly damaged in the experiments formulated by three types of transformation, i.e., in experiments \textbf{Exp1.11} and \textbf{Exp1.12}. Whereas in the four-dimensional datasets, the models are more affected by the complexity of the classification function, i.e., in the experiments where the true classification function is $F_2$, namely, the experiments with even numbers.


In the two-dimensional experiments, as shown in Fig. \ref{svm_2d}, the SVM algorithm was found to be more susceptible to scaling than to translating the data distribution, i.e., in experiments \textbf{Exp1.5-1.8}. This is interwoven with the decision boundary built by the SVM algorithm for the classification problem. Scaling the covariance matrix will lead to changes in the dispersion of the data points that could require a change in the shape of the decision boundary, whereas translating the mean would require shifting the decision boundary without adjusting its shape. In contrast, as shown in Fig. \ref{lr_2d}, the LR algorithm has shown poorer performance in the translated one-axis distribution and the scaled two-axis distribution. RF algorithm has exhibited a steady degradation rate that increases with the complexity of the drift simulated in the dataset, as shown in Fig. \ref{rf_2d}. Similarly to RF, GNB and KNN algorithms have shown relative degradation rates to the types of drift simulated in the data distribution, by scoring low degradation rates in experiments that include single drifts, and higher rates in experiments that include mixed types of drift, as shown in Fig. \ref{gnb_2d} and Fig. \ref{knn_2d}.

In four-dimensional experiments, it is clear that performance degradation is most dominant by the complexity of the class posterior probability function used across all ML models; see Fig. \ref{models_4d}. Specifically, the MCC metric was the most sensitive, showing the highest degradation rates in all experiments. However, the degradation rates of accuracy and F-score are quite close to each other.

\subsection{Region-Based Performance Evaluation}

After evaluating the overall performance of the ML models in the presence of distributional shifts, the performance of the ML models have been evaluated on subpopulations of the data to address \textbf{RQ3}. To ensure an impartial evaluation, we assessed the performance on 10,000 data points in each region. The findings of the region-based evaluation of the two-dimensional experiments are shown in Table \ref{tab:2dregresults} and in Table \ref{tab:4dregresults} for the four-dimensional experiments.

From the results, it can be seen that the performance of the different ML models is very high in the \textbf{R1} region, that is, in the region with a high training input density in most experiments. However, the models are more erroneous in the \textbf{R2} region, that is, in the region with a low training input density. This is an important observation that the models tend to be accurate in the high-training input density regions and work well in such regions, even though the distribution has changed. Furthermore, upon comparing Tables \ref{tab:2dresults} and \ref{tab:2dregresults}, and Tables \ref{tab:4dresults} and \ref{tab:4dregresults}, it can be observed that the results obtained from data sampled from the same training distribution are similar to the outcomes observed in region \textbf{R1}. This is because the majority of data points are concentrated in this region due to a higher training input density. Likewise, for the drifted distribution, the results are similar to those obtained from region \textbf{R2} since the test density is high in this region, resulting in a higher number of data points contributing to the experiments.

This result can be beneficial for interpreting where models tend to fail when making predictions. Additionally, it can be used to develop an adaptive solution for covariate shift situations, where a region-based importance weight can be introduced to correct the bias and ensure high-robustness in the affected regions. Moreover, introducing such region-based importance weights can reduce the cost of computing point-wise importance weights to region-wise importance weights in the cases of large datasets. This would also solve the drawback of the conventional importance weighting technique, where $w(x)$ can become unbounded and very large for a few points, leading to large variances of the estimated ratios \cite{Cortes2014Domain}.

In the two-dimensional experiments, it can be seen that the LR and GNB algorithms perform poorly in the \textbf{R1} region in the experiments with mixed drifts. However, they do not tend to be sensitive to this region in the four-dimensional experiments, as they exhibit similar performance across the evaluation metrics in all the experiments. It is also noteworthy to see that the RF algorithm has a high performance in \textbf{R1} in all two- and four-dimensional experiments in all evaluation metrics, and most of the misclassifications were in the \textbf{R2} region. Similarly to the overall robustness results presented in the previous section, the MCC appears to be the most susceptible metric compared to accuracy and F-score in the \textbf{R2} region, showing higher performance loss rates. Furthermore, the complexity of the decision function has been shown to have a higher impact on the \textbf{R2} region. 

The performance of ML models in the context of region-based analysis was further investigated by conducting an examination of the data points based on the density ratio values. Specifically, the region \textbf{R1} consists of data points with a density ratio of a positive number that is less than 1, while region \textbf{R2} includes data points with a density ratio higher than 1. To provide a comprehensive overview of the density ratio values in the experiments, quartiles were calculated, which can be found summarized in Table \ref{tab:2d_quartiles} for the two-dimensional experiments, and Table \ref{tab:4d_quartiles} in the appendix section \ref{sec:app_b}. The performance metric used in this analysis is the accuracy, as it was found to have high correlations with other performance metrics, see Tables \ref{fig:2dcorr} and \ref{fig:4dcorr}. Therefore, similar insights could be obtained by analyzing the different performance metrics.

\begin{figure*}[!ht]
\centering

\begin{subfigure}{\linewidth}
\centering
\includegraphics[scale=0.5]{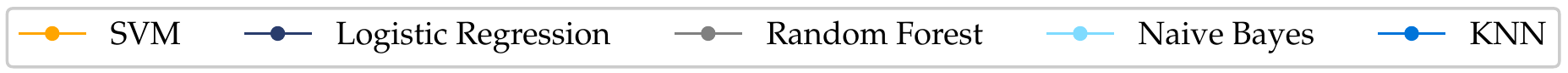}
\end{subfigure}

\begin{subfigure}{0.33\textwidth}
    \centering
    \includegraphics[width=\linewidth]{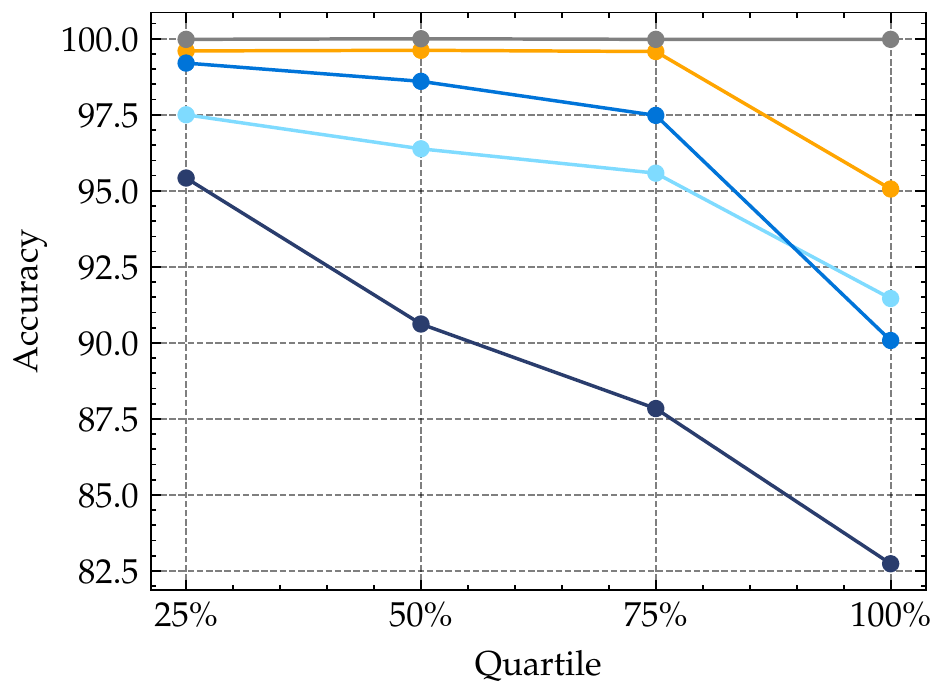}
    \caption{Exp1.1}
    \label{fig:DR_Exp1.1}
\end{subfigure}\hfil 
\begin{subfigure}{0.32\textwidth}
    \centering
    \includegraphics[width=\linewidth]{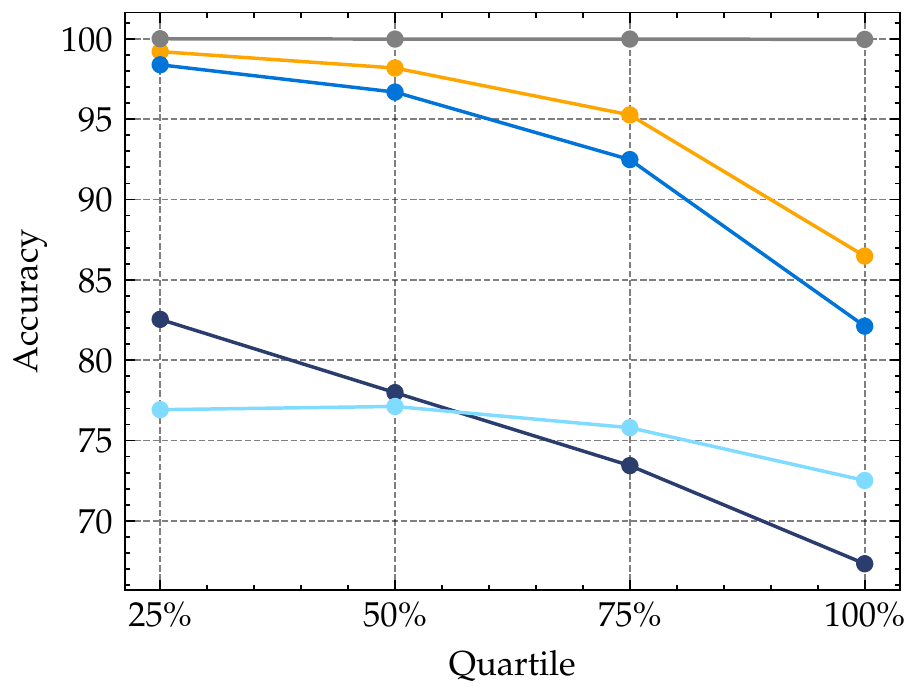}
    \caption{Exp1.2}
    \label{fig:DR_Exp1.2}
\end{subfigure}\hfil 
\begin{subfigure}{0.32\textwidth}
    \centering
    \includegraphics[width=\linewidth]{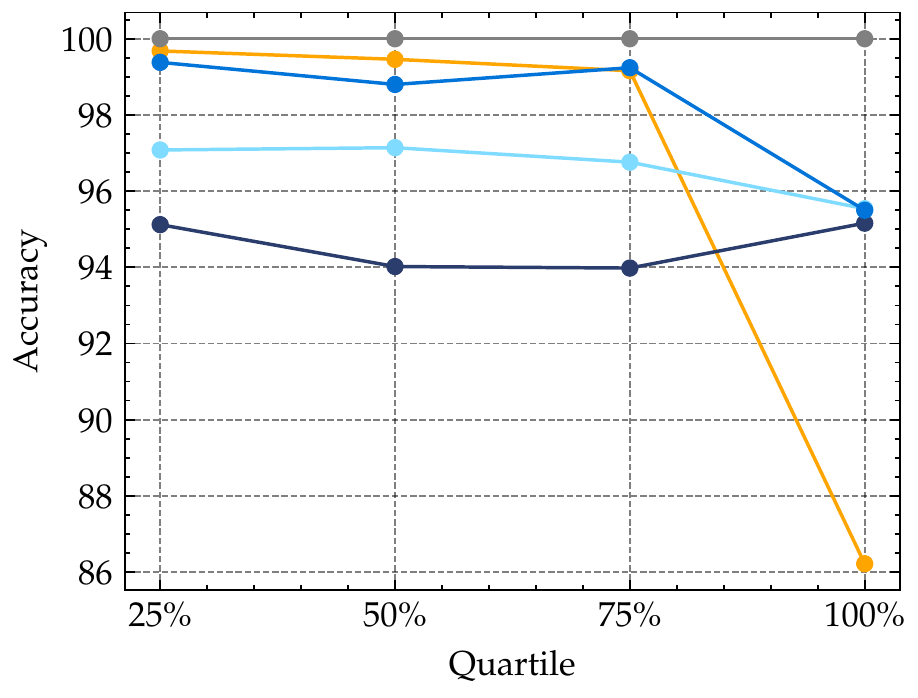}
    \caption{Exp1.3}
    \label{fig:DR_Exp1.3}
\end{subfigure}\hfil 

\begin{subfigure}{0.32\textwidth}
    \centering
    \includegraphics[width=\linewidth]{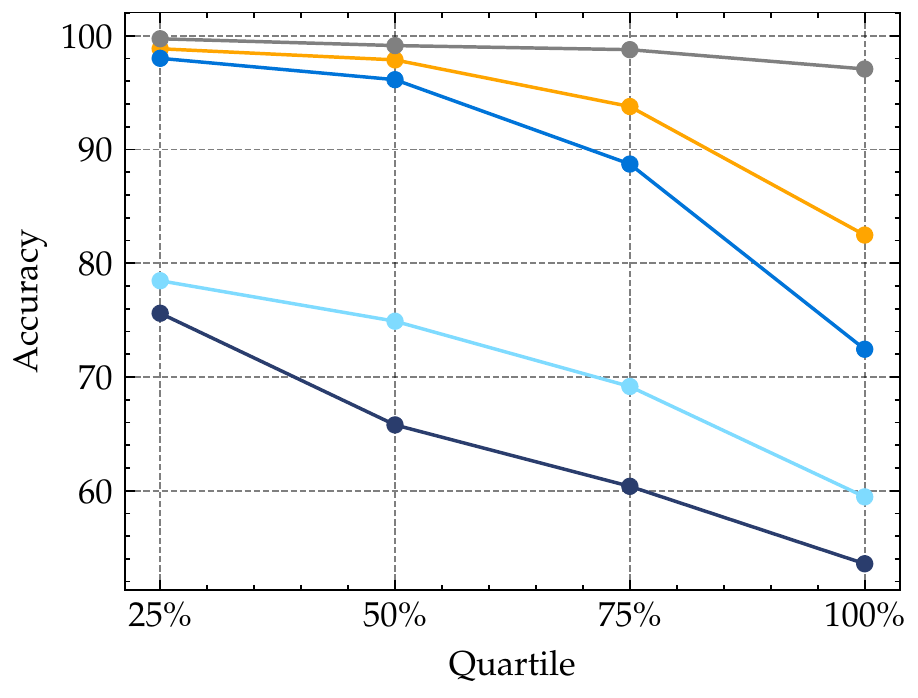}
    \caption{Exp1.4}
    \label{fig:DR_Exp1.4}
\end{subfigure}\hfil 
\begin{subfigure}{0.32\textwidth}
    \centering
    \includegraphics[width=\linewidth]{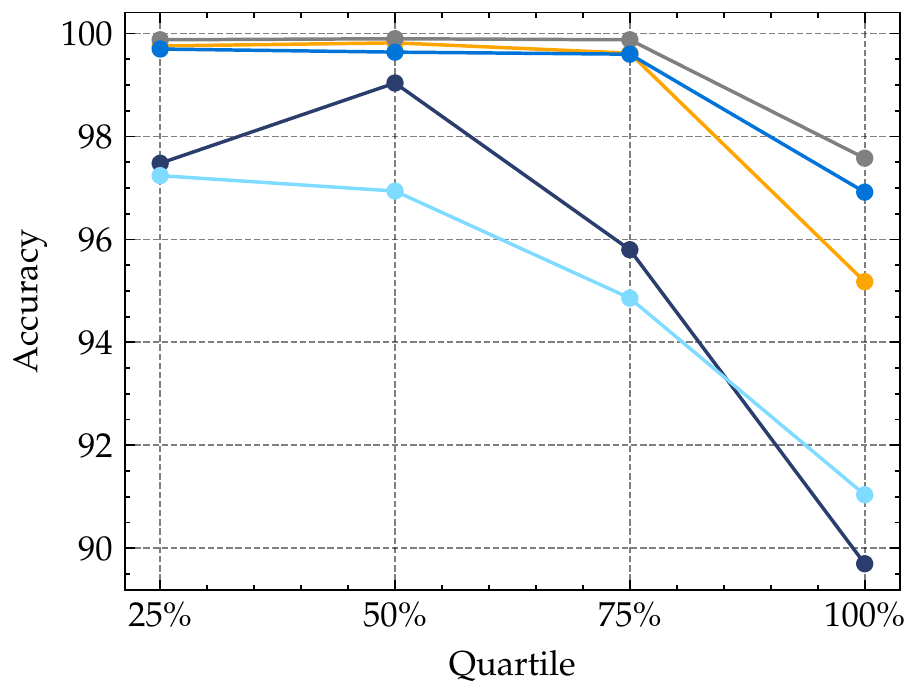}
    \caption{Exp1.5}
    \label{fig:DR_Exp1.5}
\end{subfigure}\hfil 
\begin{subfigure}{0.32\textwidth}
    \centering
    \includegraphics[width=\linewidth]{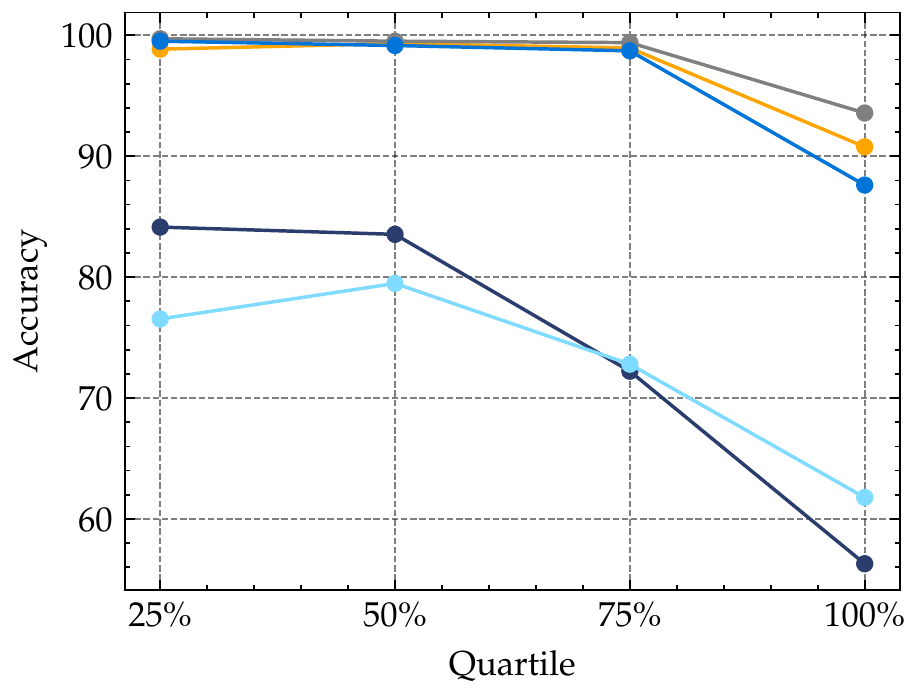}
    \caption{Exp1.6}
    \label{fig:DR_Exp1.6}
\end{subfigure}\hfil 

\begin{subfigure}{0.32\textwidth}
    \centering
    \includegraphics[width=\linewidth]{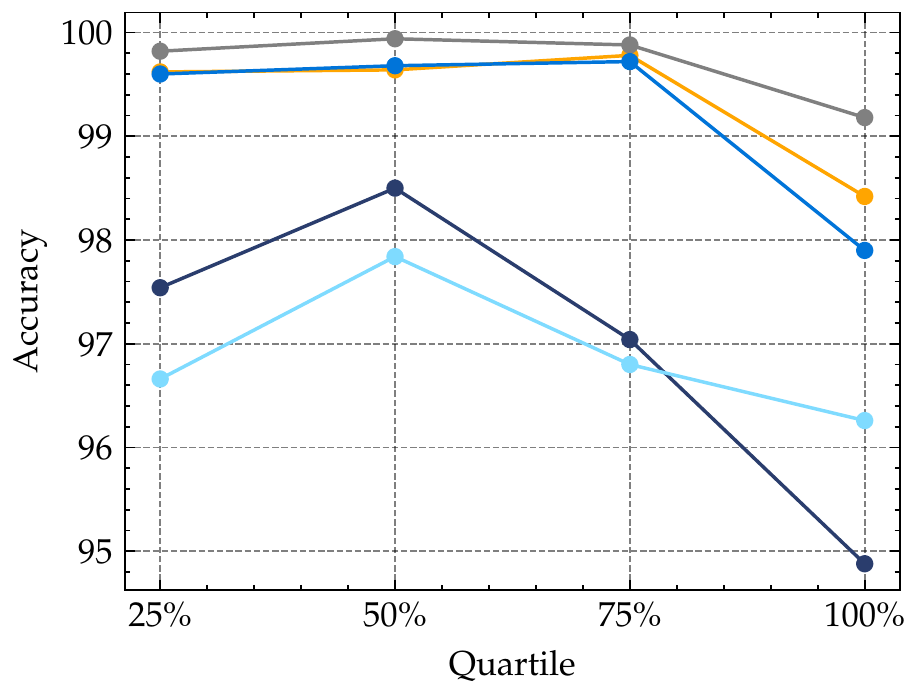}
    \caption{Exp1.7}
    \label{fig:DR_Exp1.7}
\end{subfigure}\hfil 
\begin{subfigure}{0.32\textwidth}
    \centering
    \includegraphics[width=\linewidth]{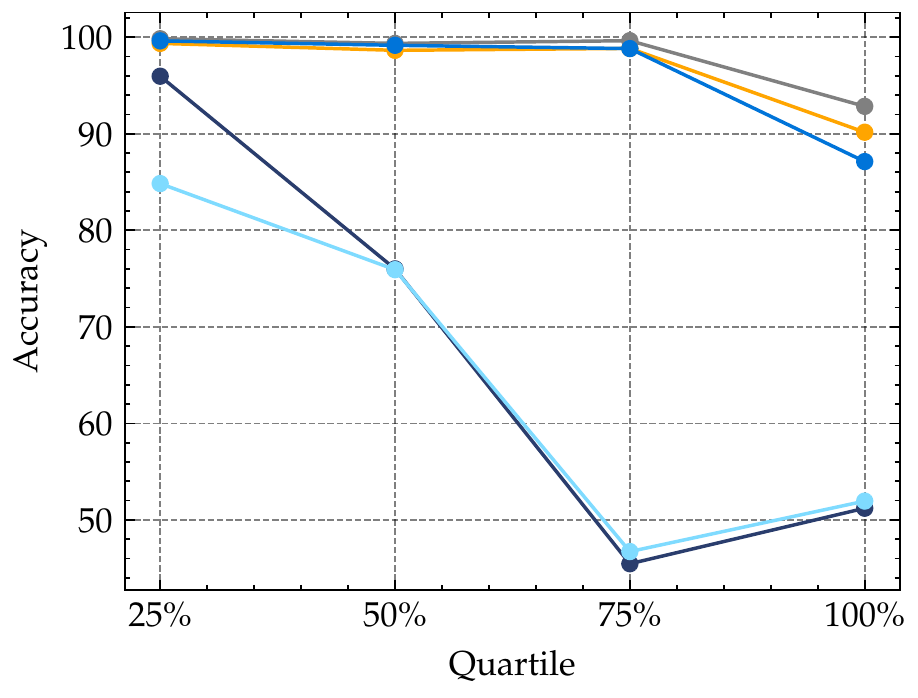}
    \caption{Exp1.8}
    \label{fig:DR_Exp1.8}
\end{subfigure}\hfil 
\begin{subfigure}{0.32\textwidth}
    \centering
    \includegraphics[width=\linewidth]{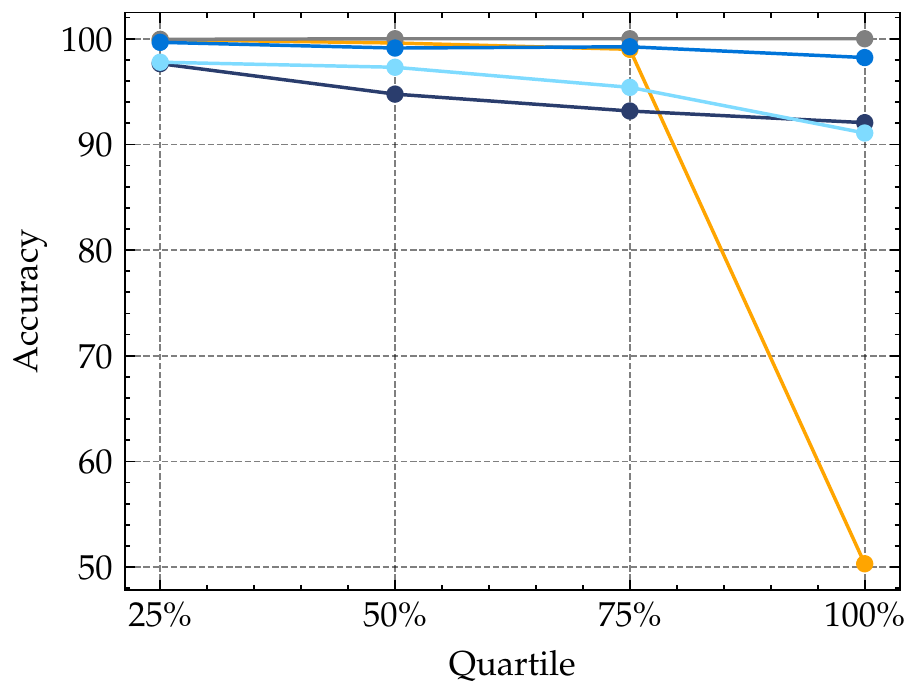}
    \caption{Exp1.9}
    \label{fig:DR_Exp1.9}
\end{subfigure}\hfil

\begin{subfigure}{0.32\textwidth}
    \centering
    \includegraphics[width=\linewidth]{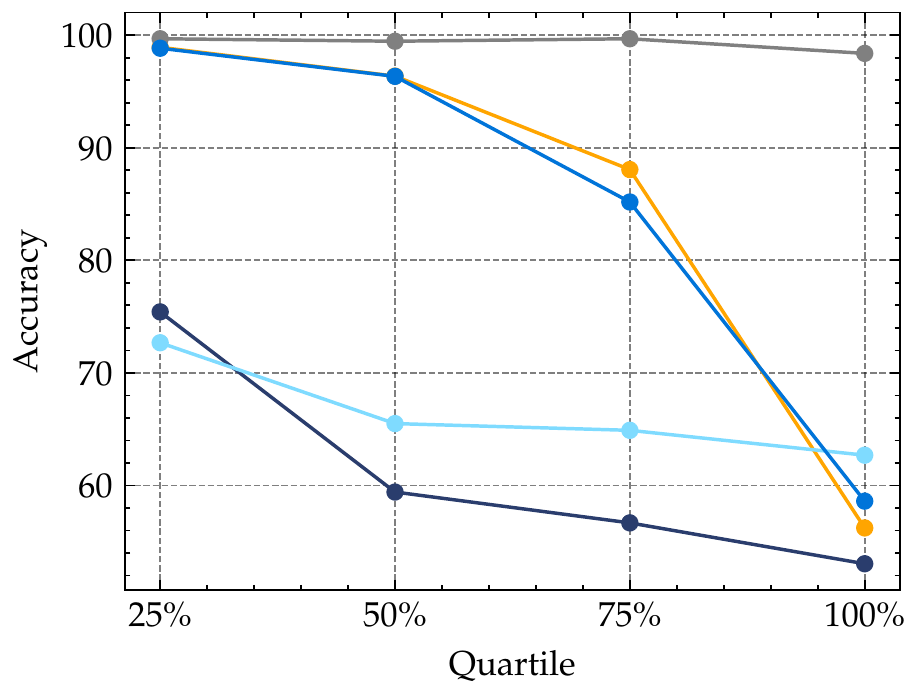}
    \caption{Exp1.10}
    \label{fig:DR_Exp1.10}
\end{subfigure}\hfil 
\begin{subfigure}{0.32\textwidth}
    \centering
    \includegraphics[width=\linewidth]{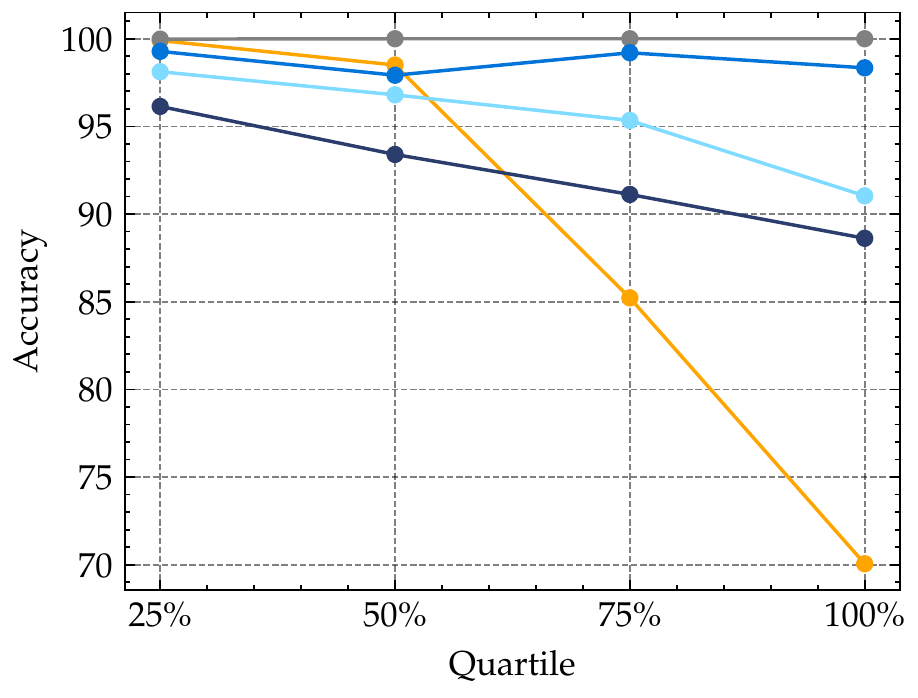}
    \caption{Exp1.11}
    \label{fig:DR_Exp1.11}
\end{subfigure}\hfil 
\begin{subfigure}{0.32\textwidth}
    \centering
    \includegraphics[width=\linewidth]{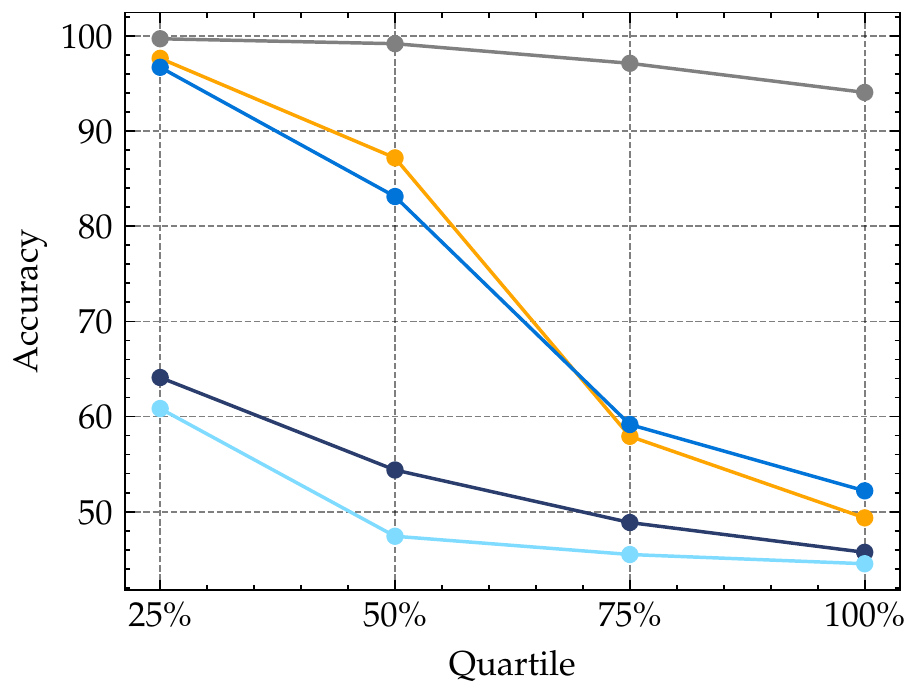}
    \caption{Exp1.12}
    \label{fig:DR_Exp1.12}
\end{subfigure}\hfil

\caption{Accuracy of machine learning models for each quartile of test-to-training density ratios in the two-dimensional experiments.}
\label{fig:2d_DR_figs}
\end{figure*}

\begin{figure*}
\centering

\begin{subfigure}{\linewidth}
\centering
\includegraphics[scale=0.5]{Legend_DR.pdf}
\end{subfigure}

\begin{subfigure}{0.32\textwidth}
\centering
\includegraphics[width=\linewidth]{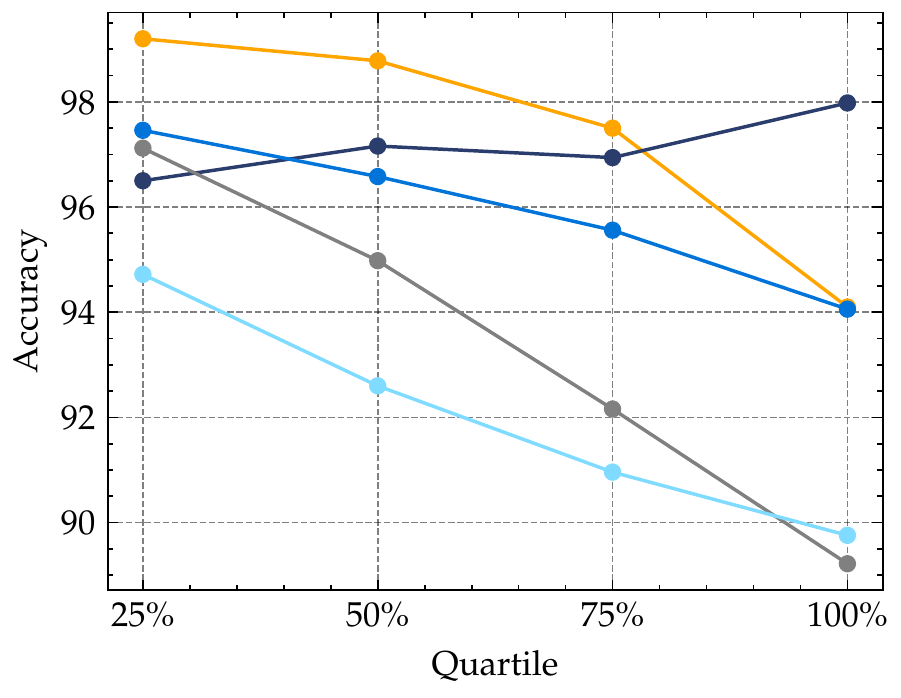}
\caption{Exp2.1}
\label{fig:DR_Exp2.1}
\end{subfigure}\hfil
\begin{subfigure}{0.32\textwidth}
\centering
\includegraphics[width=\linewidth]{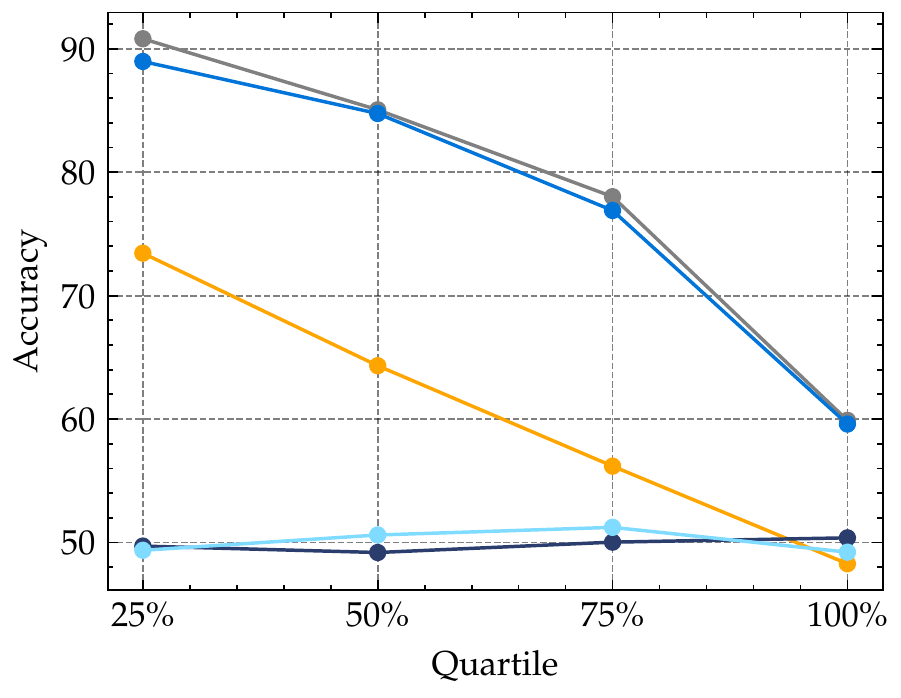}
\caption{Exp2.2}
\label{fig:DR_Exp2.2}
\end{subfigure}\hfil
\begin{subfigure}{0.32\textwidth}
\centering
\includegraphics[width=\linewidth]{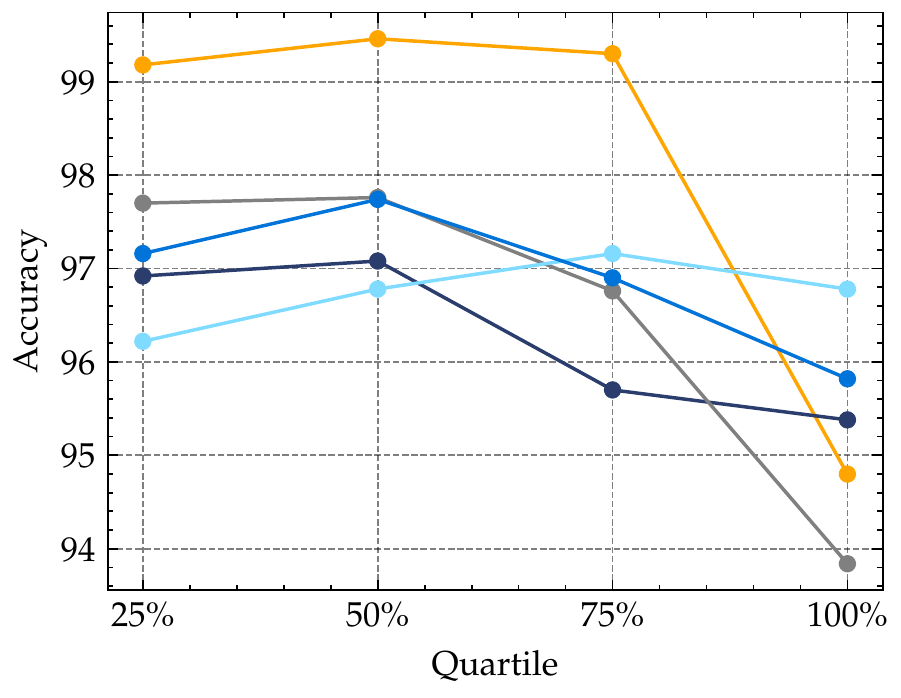}
\caption{Exp2.3}
\label{fig:DR_Exp2.3}
\end{subfigure}\hfil

\begin{subfigure}{0.32\textwidth}
\centering
\includegraphics[width=\linewidth]{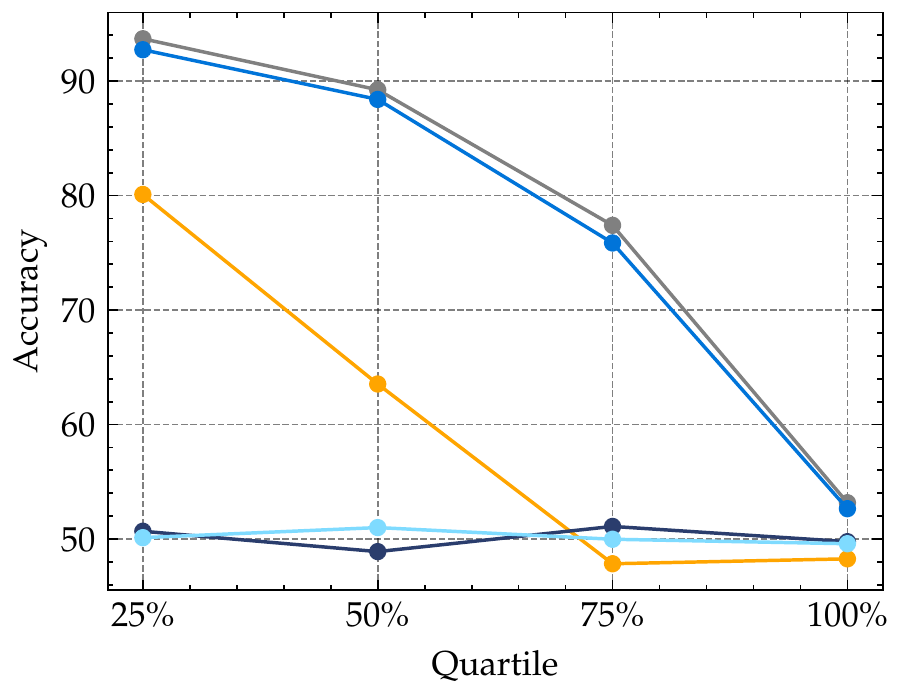}
\caption{Exp2.4}
\label{fig:DR_Exp2.4}
\end{subfigure}\hfil
\begin{subfigure}{0.33\textwidth}
\centering
\includegraphics[width=\linewidth]{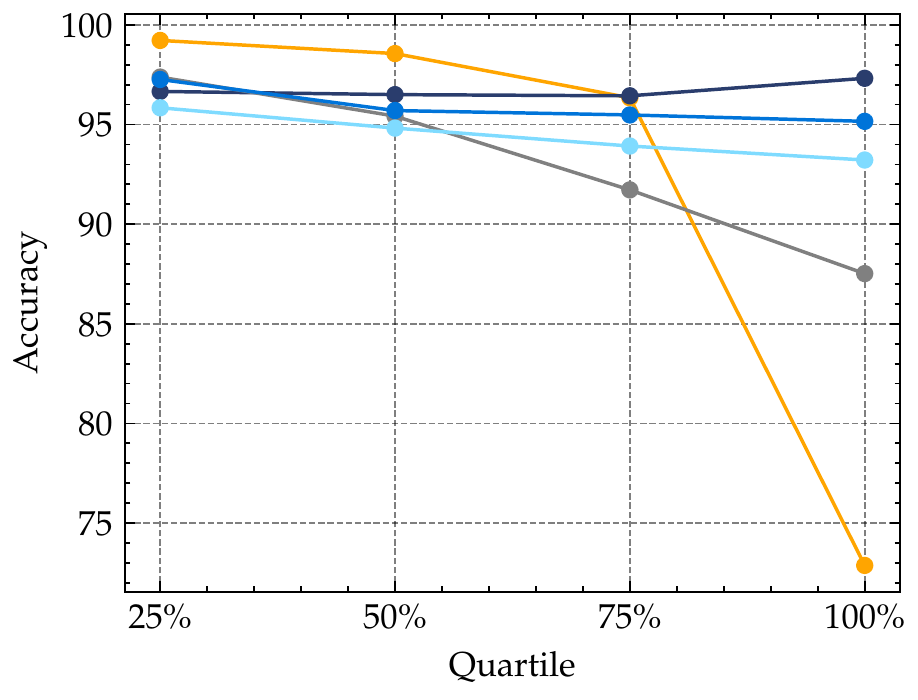}
\caption{Exp2.5}
\label{fig:DR_Exp2.5}
\end{subfigure}\hfil
\begin{subfigure}{0.32\textwidth}
\centering
\includegraphics[width=\linewidth]{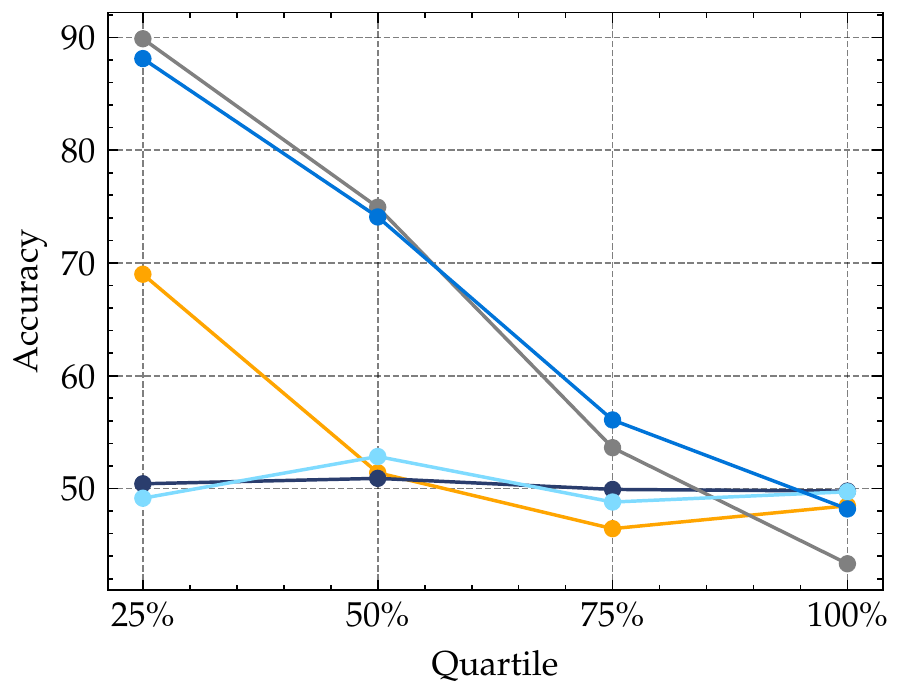}
\caption{Exp2.6}
\label{fig:DR_Exp2.6}
\end{subfigure}\hfil

\caption{Accuracy of machine learning models for each quartile of test-to-training density ratios in the four-dimensional experiments.}
\label{fig:4d_DR_figs}
\end{figure*}

The performance of the ML models in each quartile of the density ratio for the two-dimensional experiments is shown in Fig. \ref{fig:2d_DR_figs}. As can be observed, in most cases, the performance of the models degrades significantly as the density ratio increases, indicating that the models become less reliable as we move further from the training domain region. In particular, SVM exhibits the highest level of sensitivity as the density ratio increases, while RF demonstrates the highest level of robustness at high density ratio values. Similar trends are observed in the four-dimensional experiments, as shown in Fig. \ref{fig:4d_DR_figs}. However, LR and NB notably exhibit relatively stable performance across the density ratio quartiles in most experiments.

\begin{table}
\centering
\caption{Comparison of region-based performance in the two-dimensional experiments}
\label{tab:2dregresults}
\setlength{\tabcolsep}{1.5pt}
\tiny
\begin{tabular}{clllllll} 
\hline
\multicolumn{1}{l}{\multirow{2}{*}{\textbf{Exp \#}}} & \multirow{2}{*}{\begin{tabular}[c]{@{}l@{}}\textbf{ML }\\\textbf{model}\end{tabular}}& \multicolumn{3}{l}{\textbf{R1 region performance}}         & \multicolumn{3}{l}{\textbf{R2 region performance}}       \\
\multicolumn{1}{l}{}                                                                               & \multicolumn{1}{c}{}                                   & \textbf{Acc} & \textbf{F-score} & \textbf{MCC}    & \textbf{Acc} & \textbf{F-score} & \textbf{MCC}  \\ 
\toprule
\multirow{5}{*}{\textbf{Exp1.1}}                                                                   & \textbf{ SVM }                                         & \textbf{1.0*}       & \textbf{1.0*}       & \textbf{1.0*}    & 0.9835             & 0.9836             & 0.9673         \\
                                                                                                   & \textbf{ LR }                                          & \textbf{0.9846*}   & \textbf{0.9846*}    & \textbf{0.9692*}  & 0.8848             & 0.8959             & 0.7912         \\
                                                                                                   & \textbf{ RF }                                          & \textbf{1.0*}       & \textbf{1.0*}       & \textbf{1.0*}     & 0.9998             & 0.9998             & 0.9997         \\
                                                                                                   & \textbf{ GNB }                                         & \textbf{0.9721*}   & \textbf{0.9713*}   & \textbf{0.9457*} & 0.9509             & 0.9478             & 0.906          \\
                                                                                                   & \textbf{ KNN }                                         & \textbf{0.9949*}   & \textbf{0.9948*}   & \textbf{0.9897*} & 0.961              & 0.9615             & 0.923          \\ 
\midrule
\multirow{5}{*}{\begin{tabular}[c]{@{}c@{}}\\\textbf{Exp1.2}\end{tabular}}                         & \textbf{ SVM }                                         & \textbf{0.9965*}   & \textbf{0.9974*}   & \textbf{0.9923*} & 0.9591             & 0.9755             & 0.8616         \\
                                                                                                   & \textbf{ LR }                                          & \textbf{0.9784*}   & \textbf{0.9838*}   & \textbf{0.9518*} & 0.9437             & 0.9681             & 0.7571         \\
                                                                                                   & \textbf{ RF }                                          & \textbf{1.0*}      & \textbf{1.0*}      & \textbf{1.0*}    & 0.9981             & 0.998              & 0.997          \\
                                                                                                   & \textbf{ GNB }                                         & \textbf{0.9697*}   & 0.9765             & \textbf{0.9361*} & 0.9661             & \textbf{0.9798*}   & 0.8821         \\
                                                                                                   & \textbf{ KNN }                                         & \textbf{0.9974*}   & \textbf{0.998*}    & \textbf{0.9942*} & 0.9814             & 0.9892             & 0.9234         \\ 
\midrule
\multirow{5}{*}{\textbf{Exp1.3}}                                                                   & \textbf{ SVM }                                         & \textbf{0.9977*}   & \textbf{0.9975*}   & \textbf{0.9954*} & 0.9739             & 0.9653             & 0.9446         \\
                                                                                                   & \textbf{ LR }                                          & \textbf{0.9826*}   & \textbf{0.981*}    & \textbf{0.9651*} & 0.9268             & 0.9125             & 0.8601         \\
                                                                                                   & \textbf{ RF }                                          & \textbf{0.9989*}   & \textbf{0.9988*}   & \textbf{0.9978*} & 0.9871             & 0.9834             & 0.9733         \\
                                                                                                   & \textbf{ GNB }                                         & \textbf{0.9708*}   & \textbf{0.968*}    & \textbf{0.9412*} & 0.9291             & 0.91               & 0.8527         \\
                                                                                                   & \textbf{ KNN }                                         & \textbf{0.9967*}   & \textbf{0.9965*}   & \textbf{0.9934*} & 0.9824             & 0.9772             & 0.963          \\ 
\midrule
\multirow{5}{*}{\textbf{Exp1.4}}                                                                   & \textbf{ SVM }                                         & \textbf{0.9966*}   & \textbf{0.9963*}   & \textbf{0.9931*} & 0.9911             & 0.9899             & 0.982          \\
                                                                                                   & \textbf{ LR }                                          & \textbf{0.9807*}   & \textbf{0.9792*}   & \textbf{0.9612*} & 0.9605             & 0.9573             & 0.9236         \\
                                                                                                   & \textbf{ RF }                                          & \textbf{0.9988*}   & \textbf{0.9987*}   & \textbf{0.9976*} & 0.9955             & 0.9949             & 0.9909         \\
                                                                                                   & \textbf{ GNB }                                         & \textbf{0.9727*}   & \textbf{0.9704*}   & \textbf{0.9454*} & 0.9656             & 0.9615             & 0.9305         \\
                                                                                                   & \textbf{ KNN }                                         & \textbf{0.9968*}   & \textbf{0.9966*}   & \textbf{0.9935*} & 0.9883             & 0.9866             & 0.9764         \\ 
\midrule
\multirow{5}{*}{\textbf{Exp1.5}}                                                                   & \textbf{ SVM }                                         & \textbf{0.9993*}   & \textbf{0.9994*}   & \textbf{0.9986*} & 0.8499             & 0.8941             & 0.6906         \\
                                                                                                   & \textbf{ LR }                                          & \textbf{0.9838*}   & \textbf{0.9867*}   & \textbf{0.9661*} & 0.9371             & 0.9614             & 0.8104         \\
                                                                                                   & \textbf{ RF }                                          & \textbf{1.0*}      & \textbf{1.0*}      & \textbf{1.0*}    & 0.9989             & 0.9982             & 0.9979         \\
                                                                                                   & \textbf{ GNB }                                         & \textbf{0.9774*}   & \textbf{0.9811*}   & \textbf{0.9536*} & 0.9498             & 0.9669             & 0.8716         \\
                                                                                                   & \textbf{ KNN }                                         & \textbf{0.9973*}   & \textbf{0.9978*}   & \textbf{0.9944*} & 0.9894             & 0.9933             & 0.9688         \\ 
\midrule
\multirow{5}{*}{\textbf{Exp1.6}}                                                                   & \textbf{ SVM }                                         & \textbf{0.9981*}   & \textbf{0.9974*}   & \textbf{0.9958*} & 0.8746             & 0.7204             & 0.6879         \\
                                                                                                   & \textbf{ LR }                                          & \textbf{0.9819*}   & \textbf{0.9758*}   & \textbf{0.9617*} & 0.9183             & 0.8745             & 0.8296         \\
                                                                                                   & \textbf{ RF }                                          & \textbf{1.0*}      & \textbf{1.0*}      & \textbf{1.0*}    & 0.9981             & 0.998              & 0.996          \\
                                                                                                   & \textbf{ GNB }                                         & \textbf{0.9806*}   & \textbf{0.9729*}   & \textbf{0.9587*} & 0.951              & 0.9057             & 0.8801         \\
                                                                                                   & \textbf{ KNN }                                         & \textbf{0.9974*}   & \textbf{0.9965*}   & \textbf{0.9944*} & 0.986              & 0.9752             & 0.9654         \\ 
\midrule
\multirow{5}{*}{\textbf{Exp1.7}}                                                                   & \textbf{ SVM }                                         & \textbf{0.9918*}   & \textbf{0.9919*}   & \textbf{0.9836*} & 0.9446             & 0.9449             & 0.8893         \\
                                                                                                   & \textbf{ LR }                                          & \textbf{0.856*}    & \textbf{0.859*}    & \textbf{0.7118*} & 0.7459             & 0.7766             & 0.5118         \\
                                                                                                   & \textbf{ RF }                                          & \textbf{1.0*}      & \textbf{1.0*}      & \textbf{1.0*}    & 0.995              & 0.995              & 0.991          \\
                                                                                                   & \textbf{ GNB }                                         & \textbf{0.756*}    & \textbf{0.7291*}   & \textbf{0.5307*} & 0.7558             & 0.719              & 0.53           \\
                                                                                                   & \textbf{ KNN }                                         & \textbf{0.9881*}   & \textbf{0.9883*}   & \textbf{0.9761*} & 0.9196             & 0.9209             & 0.8396         \\ 
\midrule
\multirow{5}{*}{\textbf{Exp1.8}}                                                                   & \textbf{ SVM }                                         & \textbf{0.9895*}   & \textbf{0.9908*}   & \textbf{0.9786*} & 0.9291             & 0.9369             & 0.8567         \\
                                                                                                   & \textbf{ LR }                                          & \textbf{0.8273*}   & \textbf{0.8631*}   & \textbf{0.6566*} & 0.6269             & 0.7528             & 0.2628         \\
                                                                                                   & \textbf{ RF }                                          & \textbf{0.9982*}   & \textbf{0.9985*}   & \textbf{0.9964*} & 0.9862             & 0.9878             & 0.9722         \\
                                                                                                   & \textbf{ GNB }                                         & \textbf{0.7844*}   & \textbf{0.8105*}   & \textbf{0.5605*} & 0.7001             & 0.7874             & 0.4139         \\
                                                                                                   & \textbf{ KNN }                                         & \textbf{0.9895*}   & \textbf{0.9908*}   & \textbf{0.9786*} & 0.8821             & 0.9024             & 0.7609         \\ 
\midrule
\multirow{5}{*}{\textbf{Exp1.9}}                                                                   & \textbf{ SVM }                                         & \textbf{0.9909*}   & \textbf{0.9904*}   & \textbf{0.9817*} & 0.9482             & 0.9463             & 0.8963         \\
                                                                                                   & \textbf{ LR }                                          & \textbf{0.837*}    & \textbf{0.8191*}   & \textbf{0.6746*} & 0.642              & 0.6084             & 0.2826         \\
                                                                                                   & \textbf{ RF }                                          & \textbf{0.9962*}   & \textbf{0.996*}    & \textbf{0.9925*} & 0.9645             & 0.9629             & 0.929          \\
                                                                                                   & \textbf{ GNB }                                         & \textbf{0.7788*}   & \textbf{0.7309*}   & \textbf{0.5701*} & 0.673              & 0.5847             & 0.3631         \\
                                                                                                   & \textbf{ KNN }                                         & \textbf{0.9935*}   & \textbf{0.9931*}   & \textbf{0.9869*} & 0.9309             & 0.928              & 0.8618         \\ 
\midrule
\multirow{5}{*}{\textbf{Exp1.10}}                                                                  & \textbf{ SVM }                                         & \textbf{0.9903*}   & \textbf{0.9896*}   & \textbf{0.9805*} & 0.9482             & 0.9499             & 0.8965         \\
                                                                                                   & \textbf{ LR }                                          & \textbf{0.8889*}   & \textbf{0.8773*}   & \textbf{0.7779*} & 0.4882             & 0.4831             & -0.0206        \\
                                                                                                   & \textbf{ RF }                                          & \textbf{0.9967*}   & \textbf{0.9965*}   & \textbf{0.9934*} & 0.9644             & 0.9658             & 0.9288         \\
                                                                                                   & \textbf{ GNB }                                         & \textbf{0.8265*}   & \textbf{0.7857*}   & \textbf{0.6703*} & 0.4985             & 0.4886             & 0.0008         \\
                                                                                                   & \textbf{ KNN }                                         & \textbf{0.9941*}   & \textbf{0.9937*}   & \textbf{0.9882*} & 0.9345             & 0.9376             & 0.8688         \\ 
\midrule
\multirow{5}{*}{\textbf{Exp1.11}}                                                                  & \textbf{ SVM }                                         & \textbf{0.9909*}   & \textbf{0.9916*}   & \textbf{0.9816*} & 0.8244             & 0.8163             & 0.652          \\
                                                                                                   & \textbf{ LR }                                          & \textbf{0.814*}    & \textbf{0.8369*}   & \textbf{0.625*}  & 0.5763             & 0.6911             & 0.2213         \\
                                                                                                   & \textbf{ RF }                                          & \textbf{0.9956*}   & \textbf{0.996*}    & \textbf{0.9911*} & 0.9926             & 0.9926             & 0.9851         \\
                                                                                                   & \textbf{ GNB }                                         & \textbf{0.7588*}   & \textbf{0.7722*}   & \textbf{0.5174*} & 0.6481             & 0.7257             & 0.3546         \\
                                                                                                   & \textbf{ KNN }                                         & \textbf{0.9909*}   & \textbf{0.9916*}   & \textbf{0.9816*} & 0.8226             & 0.8276             & 0.6458         \\ 
\midrule
\multirow{5}{*}{\textbf{Exp1.12}}                                                                  & \textbf{ SVM }                                         & \textbf{0.9865*}   & \textbf{0.9858*}   & \textbf{0.973*}  & 0.7086             & 0.6234             & 0.4801         \\
                                                                                                   & \textbf{ LR }                                          & \textbf{0.7478*}   & \textbf{0.7056*}   & \textbf{0.4966*} & 0.5145             & 0.4746             & 0.0321         \\
                                                                                                   & \textbf{ RF }                                          & \textbf{0.9961*}   & \textbf{0.9959*}   & \textbf{0.9923*} & 0.9734             & 0.9731             & 0.9479         \\
                                                                                                   & \textbf{ GNB }                                         & \textbf{0.6996*}   & \textbf{0.61*}     & \textbf{0.412*}  & 0.4785             & 0.3576             & -0.0402        \\
                                                                                                   & \textbf{ KNN }                                         & \textbf{0.9878*}   & \textbf{0.9872*}   & \textbf{0.9757*} & 0.706              & 0.6961             & 0.4146         \\
\bottomrule
\end{tabular}
\end{table}

\begin{table}
\centering
\tiny
\caption{Comparison of region-based performance in the four-dimensional experiments}
\label{tab:4dregresults}
\setlength{\tabcolsep}{1.5pt}
\begin{tabular}{clllllll} 
\hline
\multicolumn{1}{l}{\multirow{2}{*}{\textbf{Exp \#}}} & \multirow{2}{*}{\begin{tabular}[c]{@{}l@{}}\textbf{ML }\\\textbf{model}\end{tabular}} & \multicolumn{3}{l}{\textbf{R1 region performance}}         & \multicolumn{3}{l}{\textbf{R2 region performance}}          \\
\multicolumn{1}{l}{}                                                                               & \multicolumn{1}{c}{}                                   & \textbf{Acc} & \textbf{F-score} & \textbf{MCC}    & \textbf{Acc} & \textbf{F-score} & \textbf{MCC}     \\ 
\toprule
\multirow{5}{*}{\textbf{Exp2.1}}                                                                   & \textbf{ SVM }                                         & \textbf{0.9928*}   & \textbf{0.9932*}   & \textbf{0.9856*} & 0.9715             & 0.9801             & 0.93              \\
                                                                                                   & \textbf{ LR }                                          & 0.9645             & \textbf{0.9662*}   & 0.9288           & \textbf{0.9723*}   & 0.9307             & \textbf{0.9319*}  \\
                                                                                                   & \textbf{ RF }                                          & \textbf{0.9784*}   & \textbf{0.9795*}   & \textbf{0.9567*} & 0.9281             & 0.9491             & 0.8281            \\
                                                                                                   & \textbf{ GNB }                                         & \textbf{0.9591*}   & \textbf{0.9597*}   & \textbf{0.9207*} & 0.9152             & 0.9372             & 0.8234            \\
                                                                                                   & \textbf{ KNN }                                         & \textbf{0.978*}    & \textbf{0.9791*}   & \textbf{0.9558*} & 0.9568             & 0.9698             & 0.894             \\ 
\midrule
\multirow{5}{*}{\textbf{Exp2.2}}                                                                   & \textbf{ SVM }                                         & \textbf{0.7681*}   & \textbf{0.7662*}   & \textbf{0.5367*} & 0.5862             & 0.5955             & 0.1727            \\
                                                                                                   & \textbf{ LR }                                          & \textbf{0.5019*}   & \textbf{0.5615*}   & \textbf{0.0011*} & 0.4977             & 0.4617             & -0.0058           \\
                                                                                                   & \textbf{ RF }                                          & \textbf{0.9214*}   & \textbf{0.922*}    & \textbf{0.8429*} & 0.7682             & 0.7639             & 0.5366            \\
                                                                                                   & \textbf{ GNB }                                         & 0.4995             & \textbf{0.5741*}   & -0.0047          & \textbf{0.5012*}   & 0.5074             & \textbf{0.0024*}  \\
                                                                                                   & \textbf{ KNN }                                         & \textbf{0.9008*}   & \textbf{0.9008*}   & \textbf{0.8017*} & 0.7607             & 0.754              & 0.5221            \\ 
\midrule
\multirow{5}{*}{\textbf{Exp2.3}}                                                                   & \textbf{ SVM }                                         & \textbf{0.9927*}   & \textbf{0.9919*}   & \textbf{0.9853*} & 0.9759             & 0.9722             & 0.9514            \\
                                                                                                   & \textbf{ LR }                                          & \textbf{0.9696*}   & \textbf{0.9664*}   & \textbf{0.9386*} & 0.959              & 0.9553             & 0.92              \\
                                                                                                   & \textbf{ RF }                                          & \textbf{0.9767*}   & \textbf{0.9743*}   & \textbf{0.953*}  & 0.9589             & 0.9538             & 0.917             \\
                                                                                                   & \textbf{ GNB }                                         & 0.9631             & \textbf{0.9584*}   & \textbf{0.9258*} & \textbf{0.9696*}   & 0.9649             & 0.9386            \\
                                                                                                   & \textbf{ KNN }                                         & \textbf{0.9728*}   & \textbf{0.97*}     & \textbf{0.9452*} & 0.967              & 0.9627             & 0.9331            \\ 
\midrule
\multirow{5}{*}{\textbf{Exp2.4}}                                                                   & \textbf{ SVM }                                         & \textbf{0.7649*}   & \textbf{0.7631*}   & \textbf{0.5298*} & 0.5086             & 0.4879             & 0.0168            \\
                                                                                                   & \textbf{ LR }                                          & 0.4958             & 0.4743             & -0.0088          & \textbf{0.504*}    & \textbf{0.4857*}   & \textbf{0.0075*}  \\
                                                                                                   & \textbf{ RF }                                          & \textbf{0.9319*}   & \textbf{0.9311*}   & \textbf{0.864*}  & 0.7026             & 0.701              & 0.4051            \\
                                                                                                   & \textbf{ GNB }                                         & \textbf{0.5039*}   & \textbf{0.5114*}   & \textbf{0.008*}  & 0.5005             & 0.3178             & -0.0032           \\
                                                                                                   & \textbf{ KNN }                                         & \textbf{0.9208*}   & \textbf{0.9204*}   & \textbf{0.8416*} & 0.6937             & 0.6924             & 0.3873            \\ 
\midrule
\multirow{5}{*}{\textbf{Exp2.5}}                                                                   & \textbf{ SVM }                                         & \textbf{0.9933*}   & \textbf{0.9928*}   & \textbf{0.9865*} & 0.9026             & 0.9187             & 0.8121            \\
                                                                                                   & \textbf{ LR }                                          & 0.9648             & 0.9626             & 0.9295           & \textbf{0.9678*}   & \textbf{0.9753*}   & \textbf{0.93*}    \\
                                                                                                   & \textbf{ RF }                                          & \textbf{0.9758*}   & \textbf{0.9742*}   & \textbf{0.9514*} & 0.9212             & 0.9365             & 0.8365            \\
                                                                                                   & \textbf{ GNB }                                         & \textbf{0.9608*}   & \textbf{0.9565*}   & \textbf{0.9229*} & 0.9413             & 0.9521             & 0.8833            \\
                                                                                                   & \textbf{ KNN }                                         & \textbf{0.9743*}   & \textbf{0.9726*}   & \textbf{0.9484*} & 0.956              & 0.966              & 0.9039            \\ 
\midrule
\multirow{5}{*}{\textbf{Exp2.6}}                                                                   & \textbf{ SVM }                                         & \textbf{0.7269*}   & \textbf{0.7179*}   & \textbf{0.4532*} & 0.502              & 0.5272             & 0.0037            \\
                                                                                                   & \textbf{ LR }                                          & \textbf{0.5045*}            & \textbf{0.4402*}             & 0.0032           & 0.5021             & 0.1798             & \textbf{0.0108*}  \\
                                                                                                   & \textbf{ RF }                                          & \textbf{0.9189*}   & \textbf{0.9158*}   & \textbf{0.8376*} & 0.6035             & 0.6222             & 0.2077            \\
                                                                                                   & \textbf{ GNB }                                         & 0.4896             & 0.4341             & -0.026           & \textbf{0.5035*}   & \textbf{0.5227*}   & \textbf{0.0068*}  \\
                                                                                                   & \textbf{ KNN }                                         & \textbf{0.9059*}   & \textbf{0.9033*}   & \textbf{0.8116*} & 0.6201             & 0.6452             & 0.242             \\
\bottomrule
\end{tabular}
\end{table}

\section{Threats to Validity}
\label{sec:threats}
Like any other empirical study, the main threat to external validity is related to the generalization of the findings. Although the scope of this paper is to compare the performance of different ML algorithms under distributional shifts, the default parameters provided by the scikit-learn library have been used. Using a fixed set of parameters would narrow the focus to monitor the performance of the different algorithms rather than to monitor the same algorithm under different parameter settings. Furthermore, taking the impact of the models' parameters into account as a factor, in addition to the drift parameters, would extend the parameter space of the experiments expansively. However, different parameter settings would yield different results clearly, but we argue that the concluding remarks would be the same. 

On the other hand, the experiments have been conducted using synthetic data, which might not always be similar and mimic the real-world context. As a result, this may affect the generalization of our findings to real-world datasets. Despite this, in the literature, synthetic datasets are widely used in research on changing environments \cite{Tsuboi2009Direct, Shimodaira2000Covariate}, as they include information about drift, such as drift time and type \cite{Micevska2021SDDM, Lima2021Comparative}, and they provide valuable insights and a basis for further research. Moreover, real-world datasets lack annotations of ground-truth changes, and their underlying PDFs are often unknown \cite{cano2020kappa}. Additionally, these generated datasets allow us to work on controlled classification settings and simulate various drift situations, and draw concrete conclusions about the performance of the ML models.

\section{Conclusion}
\label{sec:conclusion}

In this paper, the robustness of the performance of popular machine learning algorithms was investigated in covariate shift situations through an exhaustive comparative study. Several problems using a tractable classification framework have been generated. Each problem is parameterized by specific conditions that simulate a particular type of change. Our results show that the Random Forests algorithm is the most robust algorithm in the two-dimensional experiments, showing the lowest degradation rates in the evaluation metrics. The complexity of the classification rule has a major impact on the performance in higher-dimensional experiments.

The decomposition of the input space domain into regions based on the test-to-training density ratio have allowed to diagnose the models' performance on subpopulations of the data points. The experimental results show the high bias of the algorithms in the regions where the training input density is high, despite inducing a change in the distribution of the test data points. Our future work will consist of using this observation to develop a covariate shift solution based on region-based importance weights. 

Additionally, and to address the threats presented in Section \ref{sec:threats}, we would like to investigate the effects of the models' hyperparameters in coping with distributional changes. Also, the impact of using real-world datasets in our evaluation environment. Another potential continuation for future research could be the evaluation of deep learning-based models to further validate the conclusions drawn from conventional ML models.

\section*{Acknowledgment}

This work has been funded by the Knowledge Foundation of Sweden (KKS) through the Synergy Project AIDA - A Holistic AI-driven Networking and Processing Framework for Industrial IoT (Rek:20200067).
%

\section*{Conflict of Interest}
The authors have no conflicts of interest to declare that are relevant to the content of this article.

\section*{Data Availability}
The datasets generated and/or analysed during the current study are not publicly available currently. However, the authors are willing to provide the data per request. 

\begin{appendices}
\section{Summary of Experimental Settings}\label{sec:AppA}
This Appendix provides a summary of the parameter settings and specifications used in the two-dimensional and four-dimensional experiments conducted in this paper. Tables \ref{tab:2dsummary} and \ref{tab:4dsummary} list the key parameters and their respective values for the two-dimensional and four-dimensional experiments, respectively. 

\begin{table*}
\centering
\footnotesize	
\caption{Summary of the two-dimensional experimental settings.\\
$F_{1}: \frac{1}{2}\left(1+\tanh \left(\min \left(0, x_{1}\right)+4 x_{2}\right)\right)$, $\quad$ 
$F_{2}: \frac{1}{2}\left(1+\sin \left(\min \left(0, x_{1}\right)+2 x_{2}\right)\right)$
}
\label{tab:2dsummary}
\setlength\tabcolsep{5pt} 
\begin{tabular}{cccc} 
\toprule
\textbf{\Longstack{Experiment \\ }} & \textbf{\Longstack{Transformation \\ type}} &  \textbf{\Longstack{Test data \\ distribution}}& \textbf{\Longstack{Classification \\ function}} \\  
\toprule
\textbf{Exp1.1}  & One-axis translation              &              $N\left(\boldsymbol{x} ;\left[\begin{array}{l}3 \\ 0\end{array}\right],\left[\begin{array}{ll}1 & 0 \\ 0 & 1\end{array}\right]\right)$                   &         $F_1$                          \\ 
\midrule
\textbf{Exp1.2} & One-axis translation              &                $N\left(\boldsymbol{x} ;\left[\begin{array}{l}3 \\ 0\end{array}\right],\left[\begin{array}{ll}1 & 0 \\ 0 & 1\end{array}\right]\right)$                &        $F_2$                             \\ 
\midrule
\textbf{Exp1.3} & Two-axis translation              &               $N\left(\boldsymbol{x} ;\left[\begin{array}{l}3 \\ 1\end{array}\right],\left[\begin{array}{ll}1 & 0 \\ 0 & 1\end{array}\right]\right)$                  &     $F_1$                                \\ 
\midrule
\textbf{Exp1.4} & Two-axis translation              &               $N\left(\boldsymbol{x} ;\left[\begin{array}{l}3 \\ 1\end{array}\right],\left[\begin{array}{ll}1 & 0 \\ 0 & 1\end{array}\right]\right)$                 &      $F_2$                             \\ 
\midrule
\textbf{Exp1.5} & One-axis scaling                  &               $N\left(\boldsymbol{x} ;\left[\begin{array}{l}0 \\ 0\end{array}\right],\left[\begin{array}{ll}4 & 0 \\ 0 & 1\end{array}\right]\right)$                  &     $F_1$                                \\ 
\midrule
\textbf{Exp1.6} & One-axis scaling                  &               $N\left(\boldsymbol{x} ;\left[\begin{array}{l}0 \\ 0\end{array}\right],\left[\begin{array}{ll}4 & 0 \\ 0 & 1\end{array}\right]\right)$                 &    $F_2$                               \\ 
\midrule
\textbf{Exp1.7} & Two-axis scaling                  &               $N\left(\boldsymbol{x} ;\left[\begin{array}{l}0 \\ 0\end{array}\right],\left[\begin{array}{ll}3 & 0 \\ 0 & 2\end{array}\right]\right)$                  &    $F_1$                                 \\ 
\midrule
\textbf{Exp1.8} & Two-axis scaling                  &               $N\left(\boldsymbol{x} ;\left[\begin{array}{l}0 \\ 0\end{array}\right],\left[\begin{array}{ll}3 & 0 \\ 0 & 2\end{array}\right]\right)$                  &   $F_2$                                  \\ 
\midrule
\textbf{Exp1.9} & Translation and scaling           &              $N\left(\boldsymbol{x} ;\left[\begin{array}{l}3 \\ 1\end{array}\right],\left[\begin{array}{ll}3 & 0 \\ 0 & 2\end{array}\right]\right)$                   &      $F_1$                             \\ 
\midrule
\textbf{Exp1.10} & Translation and scaling           &              $N\left(\boldsymbol{x} ;\left[\begin{array}{l}3 \\ 1\end{array}\right],\left[\begin{array}{ll}3  & 0 \\ 0 & 2\end{array}\right]\right)$                   &   $F_2$                                  \\ 
\midrule
\textbf{Exp1.11} & Translation, scaling, rotation &              $N\left(\boldsymbol{x} ;\left[\begin{array}{l}4 \\ \matminus1\end{array}\right],\left[\begin{array}{ll}3.5 & 0.5 \\ 3.5 & 0.5\end{array}\right]\right)$                   &  $F_1$                                 \\ 
\midrule
\textbf{Exp1.12} & Translation, scaling, rotation &               $N\left(\boldsymbol{x} ;\left[\begin{array}{l}4 \\ \matminus1\end{array}\right],\left[\begin{array}{ll}3.5 & 0.5 \\ 0.5 & 3.5\end{array}\right]\right)$                  &   $F_2$                                  \\
\bottomrule
\end{tabular}
\end{table*}

\begin{table*}
\centering
\footnotesize	
\caption{Summary of the four-dimensional experimental settings.\\
$F_{3}: \frac{1}{2}\left(1+\tanh \left(\min \left(0, x_{1}\right)- x_{2}+2x_3+2x_4\right)\right)$,\\
$F_{4}: \frac{1}{2}\left(1+\sin \left(\min \left(0, x_{1}\right)+4 x_{2}-3x_3+2x_4\right)\right)$}
\label{tab:4dsummary}
\setlength\tabcolsep{2pt} 
\begin{tabular}{cccc} 
\toprule
\textbf{\Longstack{Experiment \\ }} & \textbf{\Longstack{Transformation \\ type}} &  \textbf{\Longstack{Test data \\ distribution}}& \textbf{\Longstack{Classification \\ function}}  \\ 
\toprule
\textbf{Exp2.1}  & Two-axis translation              &              $N\left(\boldsymbol{x} ;\left[\begin{array}{l}0 \\ \matminus2\\\matminus1\\1\end{array}\right],\left[\begin{array}{llll}
1 & 0 & 0 & 0 \\
0 & 1 & 0 & 0 \\
0 & 0 & 1 & 0 \\
0 & 0 & 0 & 1
\end{array}\right]\right)$                   &         $F_3$                          \\ 
\midrule
\textbf{Exp2.2} & Two-axis translation              &                $N\left(\boldsymbol{x} ;\left[\begin{array}{l}0 \\ \matminus2\\\matminus1\\1\end{array}\right],\left[\begin{array}{llll}
1 & 0 & 0 & 0 \\
0 & 1 & 0 & 0 \\
0 & 0 & 1 & 0 \\
0 & 0 & 0 & 1
\end{array}\right]\right)$                &        $F_4$                             \\ 
\midrule
\textbf{Exp2.3} & Two-axis scaling              &               $N\left(\boldsymbol{x} ;\left[\begin{array}{l}0 \\ 0\\0\\0\end{array}\right],\left[\begin{array}{llll}
3 & 0 & 0 & 0 \\
0 & 2 & 0 & 0 \\
0 & 0 & 2 & 0 \\
0 & 0 & 0 & 3
\end{array}\right]\right)$                  &     $F_3$                                \\ 
\midrule
\textbf{Exp2.4} & Two-axis scaling              &              $N\left(\boldsymbol{x} ;\left[\begin{array}{l}0 \\ 0\\0\\0\end{array}\right],\left[\begin{array}{llll}
3 & 0 & 0 & 0 \\
0 & 2 & 0 & 0 \\
0 & 0 & 2 & 0 \\
0 & 0 & 0 & 3
\end{array}\right]\right)$                    &      $F_4$                             \\ 
\midrule
\textbf{Exp2.5} & Translation, scaling, rotation                  &               $N\left(\boldsymbol{x} ;\left[\begin{array}{l}0 \\ \matminus2\\\matminus1\\1\end{array}\right],\left[\begin{array}{llll}
3 & 0 & 0 & 0 \\
0 & 2 & 0 & 0 \\
0 & 0 & 2 & 0 \\
0 & 0 & 0 & 3
\end{array}\right]\right)$                    &     $F_3$                                \\ 
\midrule
\textbf{Exp2.6} & Translation, scaling, rotation                  &              $N\left(\boldsymbol{x} ;\left[\begin{array}{l}0 \\ \matminus2\\\matminus1\\1\end{array}\right],\left[\begin{array}{llll}
2.5 & 0.5 & 0 & 0 \\
0.5 & 2.5 & 0 & 0 \\
0 & 0 & 2 & 0 \\
0 & 0 & 0 & 3
\end{array}\right]\right)$                   &    $F_4$                               \\ 

\bottomrule
\end{tabular}
\end{table*}

\section{Summary of Quartile Values for Test-to-Training Density Ratios}
\label{sec:app_b}

This appendix section includes summary tables of quartile values for test-to-training density ratios in the two-dimensional and four-dimensional experiments. Table \ref{tab:2d_quartiles} shows the quartile values for the two-dimensional experiments, while Table \ref{tab:4d_quartiles} shows the quartile values for the four-dimensional experiments. The reported values include the minimum value (0\%), the 25th percentile (25\%), the median value (50\%), the 75th percentile (75\%) and the maximum value (100\%). These quartile values are useful in assessing the spread of the test-to-training density ratios.

\begin{table*}
\centering
\caption{Summary of quartile values for test-to-training density ratios in the two-dimensional experiments.}
\label{tab:2d_quartiles}
\begin{tabular}{@{}cccccc@{}}
\toprule
\textbf{Experiment} & \textbf{0\%} & \textbf{25\%} & \textbf{50\%} & \textbf{75\%} & \textbf{100\%} \\ 
\midrule
\textbf{Exp1.1} & 7.893e-04 & 3.550e-01 & 1.002e+00 & 1.126e+02 & 4.159e+04\\\midrule
\textbf{Exp1.2} & 7.798e-04 & 3.685e-01 & 1.003e+00 & 1.187e+02 & 4.272e+04 \\\midrule
\textbf{Exp1.3} & 8.932e-04 & 3.779e-01 & 1.002e+00 & 1.971e+02 & 1.046e+05 \\\midrule
\textbf{Exp1.4} & 6.042e-04 & 3.587e-01 & 1.001e+00 & 1.876e+02 & 9.802e+04 \\\midrule
\textbf{Exp1.5} & 5.000e-01 & 5.852e-01 & 1.001e+00 & 3.749e+00 & 1.971e+03 \\\midrule
\textbf{Exp1.6} & 5.000e-01 & 5.811e-01 & 1.002e+00 & 3.698e+00 & 1.670e+03 \\\midrule
\textbf{Exp1.7} & 4.083e-01 & 5.922e-01 & 1.003e+00 & 2.781e+00 & 1.910e+02 \\\midrule
\textbf{Exp1.8}& 4.083e-01 & 5.961e-01 & 1.004e+00 & 2.821e+00 & 2.043e+02 \\\midrule
\textbf{Exp1.9}& 2.611e-02 & 2.942e-01 & 1.002e+00 & 2.958e+02 & 4.480e+08 \\\midrule
\textbf{Exp1.10} & 2.618e-02 & 3.024e-01 & 1.001e+00 & 2.838e+02 & 4.531e+08 \\\midrule
\textbf{Exp1.11} & 6.460e-03 & 2.264e-01 & 1.005e+00 & 1.150e+04 & 6.803e+12 \\\midrule
\textbf{Exp1.12} & 6.009e-03 & 2.405e-01 & 1.005e+00 & 1.204e+04 & 5.318e+12 \\
\bottomrule
\end{tabular}
\end{table*}

\begin{table*}
\centering
\caption{Summary of quartile values for test-to-training density ratios in the four-dimensional experiments.}
\label{tab:4d_quartiles}
\begin{tabular}{@{}cccccc@{}}
\toprule
\textbf{Experiment} & \textbf{0\%} & \textbf{25\%} & \textbf{50\%} & \textbf{75\%} & \textbf{100\%} \\
\midrule
\textbf{Exp2.1} & 1.789e-03 & 4.076e-01 & 1.001e+00 & 2.818e+01 & 3.185e+03 \\ \midrule
\textbf{Exp2.2} & 1.333e-03 & 4.001e-01 & 1.002e+00 & 2.863e+01 & 3.180e+03 \\ \midrule
\textbf{Exp2.3} & 1.683e-01 & 4.952e-01 & 1.002e+00 & 5.169e+00 & 1.832e+03 \\ \midrule
\textbf{Exp2.4} & 1.682e-01 & 4.875e-01 & 1.002e+00 & 5.284e+00 & 1.408e+03 \\ \midrule
\textbf{Exp2.5} & 2.087e-02 & 3.180e-01 & 1.001e+00 & 7.914e+01 & 4.845e+06 \\ \midrule
\textbf{Exp2.6} & 1.875e-02 & 3.291e-01 & 1.001e+00 & 8.119e+01 & 5.672e+06 \\ 
\bottomrule
\end{tabular}
\end{table*}

\end{appendices}

\clearpage

\balance
\bibliography{sn-bibliography}


\begin{thebibliography}{63}
\ifx \bisbn   \undefined \def \bisbn  #1{ISBN #1}\fi
\ifx \binits  \undefined \def \binits#1{#1}\fi
\ifx \bauthor  \undefined \def \bauthor#1{#1}\fi
\ifx \batitle  \undefined \def \batitle#1{#1}\fi
\ifx \bjtitle  \undefined \def \bjtitle#1{#1}\fi
\ifx \bvolume  \undefined \def \bvolume#1{\textbf{#1}}\fi
\ifx \byear  \undefined \def \byear#1{#1}\fi
\ifx \bissue  \undefined \def \bissue#1{#1}\fi
\ifx \bfpage  \undefined \def \bfpage#1{#1}\fi
\ifx \blpage  \undefined \def \blpage #1{#1}\fi
\ifx \burl  \undefined \def \burl#1{\textsf{#1}}\fi
\ifx \doiurl  \undefined \def \doiurl#1{\url{https://doi.org/#1}}\fi
\ifx \betal  \undefined \def \betal{\textit{et al.}}\fi
\ifx \binstitute  \undefined \def \binstitute#1{#1}\fi
\ifx \binstitutionaled  \undefined \def \binstitutionaled#1{#1}\fi
\ifx \bctitle  \undefined \def \bctitle#1{#1}\fi
\ifx \beditor  \undefined \def \beditor#1{#1}\fi
\ifx \bpublisher  \undefined \def \bpublisher#1{#1}\fi
\ifx \bbtitle  \undefined \def \bbtitle#1{#1}\fi
\ifx \bedition  \undefined \def \bedition#1{#1}\fi
\ifx \bseriesno  \undefined \def \bseriesno#1{#1}\fi
\ifx \blocation  \undefined \def \blocation#1{#1}\fi
\ifx \bsertitle  \undefined \def \bsertitle#1{#1}\fi
\ifx \bsnm \undefined \def \bsnm#1{#1}\fi
\ifx \bsuffix \undefined \def \bsuffix#1{#1}\fi
\ifx \bparticle \undefined \def \bparticle#1{#1}\fi
\ifx \barticle \undefined \def \barticle#1{#1}\fi
\bibcommenthead
\ifx \bconfdate \undefined \def \bconfdate #1{#1}\fi
\ifx \botherref \undefined \def \botherref #1{#1}\fi
\ifx \url \undefined \def \url#1{\textsf{#1}}\fi
\ifx \bchapter \undefined \def \bchapter#1{#1}\fi
\ifx \bbook \undefined \def \bbook#1{#1}\fi
\ifx \bcomment \undefined \def \bcomment#1{#1}\fi
\ifx \oauthor \undefined \def \oauthor#1{#1}\fi
\ifx \citeauthoryear \undefined \def \citeauthoryear#1{#1}\fi
\ifx \endbibitem  \undefined \def \endbibitem {}\fi
\ifx \bconflocation  \undefined \def \bconflocation#1{#1}\fi
\ifx \arxivurl  \undefined \def \arxivurl#1{\textsf{#1}}\fi
\csname PreBibitemsHook\endcsname

\bibitem{HEUREUX12017Big}
\begin{barticle}
\bauthor{\bsnm{L’heureux}, \binits{A.}},
\bauthor{\bsnm{Grolinger}, \binits{K.}},
\bauthor{\bsnm{Elyamany}, \binits{H.F.}},
\bauthor{\bsnm{Capretz}, \binits{M.A.}}:
\batitle{Machine learning with big data: Challenges and approaches}.
\bjtitle{IEEE Access}
\bvolume{5},
\bfpage{7776}--\blpage{7797}
(\byear{2017})
\end{barticle}
\endbibitem

\bibitem{WITTEN2017Probabilistic}
\begin{bchapter}
\bauthor{\bsnm{Witten}, \binits{I.H.}},
\bauthor{\bsnm{Frank}, \binits{E.}},
\bauthor{\bsnm{Hall}, \binits{M.A.}},
\bauthor{\bsnm{Pal}, \binits{C.J.}}:
\bctitle{Chapter 9 - probabilistic methods}.
In: \beditor{\bsnm{Witten}, \binits{I.H.}},
\beditor{\bsnm{Frank}, \binits{E.}},
\beditor{\bsnm{Hall}, \binits{M.A.}},
\beditor{\bsnm{Pal}, \binits{C.J.}} (eds.)
\bbtitle{Data Mining (Fourth Edition)},
\bedition{Fourth edition} edn.,
pp. \bfpage{335}--\blpage{416}.
\bpublisher{Morgan Kaufmann},
\blocation{San Francisco, CA, USA}
(\byear{2017})
\end{bchapter}
\endbibitem

\bibitem{Brownlee2019Probability}
\begin{botherref}
\oauthor{\bsnm{Brownlee}, \binits{J.}}:
Probability for machine learning: Discover how to harness uncertainty with
  {P}ython
(2019)
\end{botherref}
\endbibitem

\bibitem{Gama2014Adaptation}
\begin{botherref}
\oauthor{\bsnm{Gama}, \binits{J.a.}},
\oauthor{\bsnm{\v{Z}liobaitundefined}, \binits{I.}},
\oauthor{\bsnm{Bifet}, \binits{A.}},
\oauthor{\bsnm{Pechenizkiy}, \binits{M.}},
\oauthor{\bsnm{Bouchachia}, \binits{A.}}:
A survey on concept drift adaptation.
ACM Comput. Surv.
\textbf{46}(4)
(2014)
\end{botherref}
\endbibitem

\bibitem{Lu2016LearningUnder}
\begin{barticle}
\bauthor{\bsnm{Lu}, \binits{J.}},
\bauthor{\bsnm{Liu}, \binits{A.}},
\bauthor{\bsnm{Dong}, \binits{F.}},
\bauthor{\bsnm{Gu}, \binits{F.}},
\bauthor{\bsnm{Gama}, \binits{J.}},
\bauthor{\bsnm{Zhang}, \binits{G.}}:
\batitle{Learning under concept drift: A review}.
\bjtitle{IEEE Transactions on Knowledge and Data Engineering}
\bvolume{31}(\bissue{12}),
\bfpage{2346}--\blpage{2363}
(\byear{2019})
\end{barticle}
\endbibitem

\bibitem{Moreno2012DShift}
\begin{barticle}
\bauthor{\bsnm{Moreno-Torres}, \binits{J.G.}},
\bauthor{\bsnm{Raeder}, \binits{T.}},
\bauthor{\bsnm{Alaiz-Rodríguez}, \binits{R.}},
\bauthor{\bsnm{Chawla}, \binits{N.V.}},
\bauthor{\bsnm{Herrera}, \binits{F.}}:
\batitle{A unifying view on dataset shift in classification}.
\bjtitle{Pattern Recognition}
\bvolume{45}(\bissue{1}),
\bfpage{521}--\blpage{530}
(\byear{2012})
\end{barticle}
\endbibitem

\bibitem{Raza2015EWMA}
\begin{barticle}
\bauthor{\bsnm{Raza}, \binits{H.}},
\bauthor{\bsnm{Prasad}, \binits{G.}},
\bauthor{\bsnm{Li}, \binits{Y.}}:
\batitle{Ewma model based shift-detection methods for detecting covariate
  shifts in non-stationary environments}.
\bjtitle{Pattern Recognition}
\bvolume{48}(\bissue{3}),
\bfpage{659}--\blpage{669}
(\byear{2015})
\end{barticle}
\endbibitem

\bibitem{Bayram2022Degradation}
\begin{barticle}
\bauthor{\bsnm{Bayram}, \binits{F.}},
\bauthor{\bsnm{Ahmed}, \binits{B.S.}},
\bauthor{\bsnm{Kassler}, \binits{A.}}:
\batitle{From concept drift to model degradation: An overview on
  performance-aware drift detectors}.
\bjtitle{Knowledge-Based Systems}
\bvolume{245},
\bfpage{108632}
(\byear{2022})
\end{barticle}
\endbibitem

\bibitem{Shimodaira2000Covariate}
\begin{barticle}
\bauthor{\bsnm{Shimodaira}, \binits{H.}}:
\batitle{Improving predictive inference under covariate shift by weighting the
  log-likelihood function}.
\bjtitle{Journal of Statistical Planning and Inference}
\bvolume{90}(\bissue{2}),
\bfpage{227}--\blpage{244}
(\byear{2000})
\end{barticle}
\endbibitem

\bibitem{Sugiyama2012IntroductiontoCS}
\begin{bbook}
\bauthor{\bsnm{Sugiyama}, \binits{M.}},
\bauthor{\bsnm{Kawanabe}, \binits{M.}}:
\bbtitle{Machine Learning in Non-Stationary Environments: Introduction to
  Covariate Shift Adaptation}.
\bpublisher{The MIT Press},
\blocation{USA}
(\byear{2012})
\end{bbook}
\endbibitem

\bibitem{Tsymbal2004DefandRel}
\begin{botherref}
\oauthor{\bsnm{Tsymbal}, \binits{A.}}:
The problem of concept drift: definitions and related work.
Computer Science Department, Trinity College Dublin
(2004)
\end{botherref}
\endbibitem

\bibitem{Tasche2017Fisher}
\begin{barticle}
\bauthor{\bsnm{Tasche}, \binits{D.}}:
\batitle{Fisher consistency for prior probability shift}.
\bjtitle{The Journal of Machine Learning Research}
\bvolume{18}(\bissue{1}),
\bfpage{3338}--\blpage{3369}
(\byear{2017})
\end{barticle}
\endbibitem

\bibitem{Li2020Robust}
\begin{bchapter}
\bauthor{\bsnm{Li}, \binits{F.}},
\bauthor{\bsnm{Lam}, \binits{H.}},
\bauthor{\bsnm{Prusty}, \binits{S.}}:
\bctitle{Robust importance weighting for covariate shift}.
In: \bbtitle{International Conference on Artificial Intelligence and
  Statistics},
pp. \bfpage{352}--\blpage{362}
(\byear{2020}).
\bcomment{PMLR}
\end{bchapter}
\endbibitem

\bibitem{Subbaswamy2020Development}
\begin{barticle}
\bauthor{\bsnm{Subbaswamy}, \binits{A.}},
\bauthor{\bsnm{Saria}, \binits{S.}}:
\batitle{From development to deployment: dataset shift, causality, and
  shift-stable models in health ai}.
\bjtitle{Biostatistics}
\bvolume{21}(\bissue{2}),
\bfpage{345}--\blpage{352}
(\byear{2020})
\end{barticle}
\endbibitem

\bibitem{Schneider2020Compvision}
\begin{barticle}
\bauthor{\bsnm{Schneider}, \binits{S.}},
\bauthor{\bsnm{Rusak}, \binits{E.}},
\bauthor{\bsnm{Eck}, \binits{L.}},
\bauthor{\bsnm{Bringmann}, \binits{O.}},
\bauthor{\bsnm{Brendel}, \binits{W.}},
\bauthor{\bsnm{Bethge}, \binits{M.}}:
\batitle{Improving robustness against common corruptions by covariate shift
  adaptation}.
\bjtitle{Advances in Neural Information Processing Systems}
\bvolume{33},
\bfpage{11539}--\blpage{11551}
(\byear{2020})
\end{barticle}
\endbibitem

\bibitem{Duchi2019NLP}
\begin{botherref}
\oauthor{\bsnm{Duchi}, \binits{J.C.}},
\oauthor{\bsnm{Hashimoto}, \binits{T.}},
\oauthor{\bsnm{Namkoong}, \binits{H.}}:
Distributionally robust losses against mixture covariate shifts.
Under review
\textbf{2}
(2019)
\end{botherref}
\endbibitem

\bibitem{Fei2015Social}
\begin{bchapter}
\bauthor{\bsnm{Fei}, \binits{G.}},
\bauthor{\bsnm{Liu}, \binits{B.}}:
\bctitle{Social media text classification under negative covariate shift}.
In: \bbtitle{Proceedings of the 2015 Conference on Empirical Methods in Natural
  Language Processing},
pp. \bfpage{2347}--\blpage{2356}
(\byear{2015})
\end{bchapter}
\endbibitem

\bibitem{He2019Unlearn}
\begin{botherref}
\oauthor{\bsnm{He}, \binits{H.}},
\oauthor{\bsnm{Zha}, \binits{S.}},
\oauthor{\bsnm{Wang}, \binits{H.}}:
Unlearn dataset bias in natural language inference by fitting the residual.
arXiv preprint arXiv:1908.10763
(2019)
\end{botherref}
\endbibitem

\bibitem{Wiemann2022Bias}
\begin{barticle}
\bauthor{\bsnm{Wiemann}, \binits{P.F.}},
\bauthor{\bsnm{Klein}, \binits{N.}},
\bauthor{\bsnm{Kneib}, \binits{T.}}:
\batitle{Correcting for sample selection bias in bayesian distributional
  regression models}.
\bjtitle{Computational Statistics \& Data Analysis}
\bvolume{168},
\bfpage{107382}
(\byear{2022})
\end{barticle}
\endbibitem

\bibitem{Tsuboi2009Direct}
\begin{barticle}
\bauthor{\bsnm{Tsuboi}, \binits{Y.}},
\bauthor{\bsnm{Kashima}, \binits{H.}},
\bauthor{\bsnm{Hido}, \binits{S.}},
\bauthor{\bsnm{Bickel}, \binits{S.}},
\bauthor{\bsnm{Sugiyama}, \binits{M.}}:
\batitle{Direct density ratio estimation for large-scale covariate shift
  adaptation}.
\bjtitle{Journal of Information Processing}
\bvolume{17},
\bfpage{138}--\blpage{155}
(\byear{2009})
\end{barticle}
\endbibitem

\bibitem{Sugiyama2007ICV}
\begin{botherref}
\oauthor{\bsnm{Sugiyama}, \binits{M.}},
\oauthor{\bsnm{Krauledat}, \binits{M.}},
\oauthor{\bsnm{M{\"u}ller}, \binits{K.-R.}}:
Covariate shift adaptation by importance weighted cross validation.
Journal of Machine Learning Research
\textbf{8}(5)
(2007)
\end{botherref}
\endbibitem

\bibitem{Subbaswamy2021Evaluating}
\begin{bchapter}
\bauthor{\bsnm{Subbaswamy}, \binits{A.}},
\bauthor{\bsnm{Adams}, \binits{R.}},
\bauthor{\bsnm{Saria}, \binits{S.}}:
\bctitle{Evaluating model robustness and stability to dataset shift}.
In: \bbtitle{International Conference on Artificial Intelligence and
  Statistics},
pp. \bfpage{2611}--\blpage{2619}
(\byear{2021}).
\bcomment{PMLR}
\end{bchapter}
\endbibitem

\bibitem{Liu2020Well}
\begin{barticle}
\bauthor{\bsnm{Liu}, \binits{H.}},
\bauthor{\bsnm{Wu}, \binits{Y.}},
\bauthor{\bsnm{Cao}, \binits{Y.}},
\bauthor{\bsnm{Lv}, \binits{W.}},
\bauthor{\bsnm{Han}, \binits{H.}},
\bauthor{\bsnm{Li}, \binits{Z.}},
\bauthor{\bsnm{Chang}, \binits{J.}}:
\batitle{Well logging based lithology identification model establishment under
  data drift: A transfer learning method}.
\bjtitle{Sensors}
\bvolume{20}(\bissue{13}),
\bfpage{3643}
(\byear{2020})
\end{barticle}
\endbibitem

\bibitem{Sakai2019PU}
\begin{bchapter}
\bauthor{\bsnm{Sakai}, \binits{T.}},
\bauthor{\bsnm{Shimizu}, \binits{N.}}:
\bctitle{Covariate shift adaptation on learning from positive and unlabeled
  data}.
In: \bbtitle{Proceedings of the AAAI Conference on Artificial Intelligence},
vol. \bseriesno{33},
pp. \bfpage{4838}--\blpage{4845}
(\byear{2019})
\end{bchapter}
\endbibitem

\bibitem{Candela2009DatasetS}
\begin{bbook}
\bauthor{\bsnm{Quionero-Candela}, \binits{J.}},
\bauthor{\bsnm{Sugiyama}, \binits{M.}},
\bauthor{\bsnm{Schwaighofer}, \binits{A.}},
\bauthor{\bsnm{Lawrence}, \binits{N.D.}}:
\bbtitle{Dataset Shift in Machine Learning}.
\bpublisher{The MIT Press},
\blocation{USA}
(\byear{2009})
\end{bbook}
\endbibitem

\bibitem{Huang2006KMM}
\begin{botherref}
\oauthor{\bsnm{Huang}, \binits{J.}},
\oauthor{\bsnm{Gretton}, \binits{A.}},
\oauthor{\bsnm{Borgwardt}, \binits{K.}},
\oauthor{\bsnm{Sch{\"o}lkopf}, \binits{B.}},
\oauthor{\bsnm{Smola}, \binits{A.}}:
Correcting sample selection bias by unlabeled data.
Advances in neural information processing systems
\textbf{19}
(2006)
\end{botherref}
\endbibitem

\bibitem{Sugiyama2007KLIEP}
\begin{botherref}
\oauthor{\bsnm{Sugiyama}, \binits{M.}},
\oauthor{\bsnm{Nakajima}, \binits{S.}},
\oauthor{\bsnm{Kashima}, \binits{H.}},
\oauthor{\bsnm{Buenau}, \binits{P.}},
\oauthor{\bsnm{Kawanabe}, \binits{M.}}:
Direct importance estimation with model selection and its application to
  covariate shift adaptation.
Advances in neural information processing systems
\textbf{20}
(2007)
\end{botherref}
\endbibitem

\bibitem{Kanamori2009LSIF}
\begin{barticle}
\bauthor{\bsnm{Kanamori}, \binits{T.}},
\bauthor{\bsnm{Hido}, \binits{S.}},
\bauthor{\bsnm{Sugiyama}, \binits{M.}}:
\batitle{A least-squares approach to direct importance estimation}.
\bjtitle{The Journal of Machine Learning Research}
\bvolume{10},
\bfpage{1391}--\blpage{1445}
(\byear{2009})
\end{barticle}
\endbibitem

\bibitem{Sugiyama2008Direct}
\begin{barticle}
\bauthor{\bsnm{Sugiyama}, \binits{M.}},
\bauthor{\bsnm{Suzuki}, \binits{T.}},
\bauthor{\bsnm{Nakajima}, \binits{S.}},
\bauthor{\bsnm{Kashima}, \binits{H.}},
\bauthor{\bparticle{von} \bsnm{B{\"u}nau}, \binits{P.}},
\bauthor{\bsnm{Kawanabe}, \binits{M.}}:
\batitle{Direct importance estimation for covariate shift adaptation}.
\bjtitle{Annals of the Institute of Statistical Mathematics}
\bvolume{60}(\bissue{4}),
\bfpage{699}--\blpage{746}
(\byear{2008})
\end{barticle}
\endbibitem

\bibitem{Chapaneri2019Covariate}
\begin{barticle}
\bauthor{\bsnm{Chapaneri}, \binits{S.V.}},
\bauthor{\bsnm{Jayaswal}, \binits{D.J.}}:
\batitle{Covariate shift adaptation for structured regression with frank--wolfe
  algorithms}.
\bjtitle{IEEE Access}
\bvolume{7},
\bfpage{73804}--\blpage{73818}
(\byear{2019})
\end{barticle}
\endbibitem

\bibitem{Alaiz2008Assessing}
\begin{bchapter}
\bauthor{\bsnm{Alaiz-Rodr{\'\i}guez}, \binits{R.}},
\bauthor{\bsnm{Japkowicz}, \binits{N.}}:
\bctitle{Assessing the impact of changing environments on classifier
  performance}.
In: \bbtitle{Conference of the Canadian Society for Computational Studies of
  Intelligence},
pp. \bfpage{13}--\blpage{24}
(\byear{2008}).
\bcomment{Springer}
\end{bchapter}
\endbibitem

\bibitem{Abbasian2010Robustness}
\begin{bchapter}
\bauthor{\bsnm{Abbasian}, \binits{H.}},
\bauthor{\bsnm{Drummond}, \binits{C.}},
\bauthor{\bsnm{Japkowicz}, \binits{N.}},
\bauthor{\bsnm{Matwin}, \binits{S.}}:
\bctitle{Robustness of classifiers to changing environments}.
In: \bbtitle{Canadian Conference on Artificial Intelligence},
pp. \bfpage{232}--\blpage{243}
(\byear{2010}).
\bcomment{Springer}
\end{bchapter}
\endbibitem

\bibitem{Chen2021Mandoline}
\begin{bchapter}
\bauthor{\bsnm{Chen}, \binits{M.}},
\bauthor{\bsnm{Goel}, \binits{K.}},
\bauthor{\bsnm{Sohoni}, \binits{N.S.}},
\bauthor{\bsnm{Poms}, \binits{F.}},
\bauthor{\bsnm{Fatahalian}, \binits{K.}},
\bauthor{\bsnm{R{\'e}}, \binits{C.}}:
\bctitle{Mandoline: Model evaluation under distribution shift}.
In: \bbtitle{International Conference on Machine Learning},
pp. \bfpage{1617}--\blpage{1629}
(\byear{2021}).
\bcomment{PMLR}
\end{bchapter}
\endbibitem

\bibitem{Sagawa2019Distributionally}
\begin{botherref}
\oauthor{\bsnm{Sagawa}, \binits{S.}},
\oauthor{\bsnm{Koh}, \binits{P.W.}},
\oauthor{\bsnm{Hashimoto}, \binits{T.B.}},
\oauthor{\bsnm{Liang}, \binits{P.}}:
Distributionally robust neural networks for group shifts: On the importance of
  regularization for worst-case generalization.
arXiv preprint arXiv:1911.08731
(2019)
\end{botherref}
\endbibitem

\bibitem{Garg2022Leveraging}
\begin{bchapter}
\bauthor{\bsnm{Garg}, \binits{S.}},
\bauthor{\bsnm{Balakrishnan}, \binits{S.}},
\bauthor{\bsnm{Lipton}, \binits{Z.C.}},
\bauthor{\bsnm{Neyshabur}, \binits{B.}},
\bauthor{\bsnm{Sedghi}, \binits{H.}}:
\bctitle{Leveraging unlabeled data to predict out-of-distribution performance}.
In: \bbtitle{ICLR}
(\byear{2022})
\end{bchapter}
\endbibitem

\bibitem{Guillory2021Predicting}
\begin{bchapter}
\bauthor{\bsnm{Guillory}, \binits{D.}},
\bauthor{\bsnm{Shankar}, \binits{V.}},
\bauthor{\bsnm{Ebrahimi}, \binits{S.}},
\bauthor{\bsnm{Darrell}, \binits{T.}},
\bauthor{\bsnm{Schmidt}, \binits{L.}}:
\bctitle{Predicting with confidence on unseen distributions}.
In: \bbtitle{Proceedings of the IEEE/CVF International Conference on Computer
  Vision},
pp. \bfpage{1134}--\blpage{1144}
(\byear{2021})
\end{bchapter}
\endbibitem

\bibitem{Deng2021Labels}
\begin{bchapter}
\bauthor{\bsnm{Deng}, \binits{W.}},
\bauthor{\bsnm{Zheng}, \binits{L.}}:
\bctitle{Are labels always necessary for classifier accuracy evaluation?}
In: \bbtitle{Proceedings of the IEEE/CVF Conference on Computer Vision and
  Pattern Recognition},
pp. \bfpage{15069}--\blpage{15078}
(\byear{2021})
\end{bchapter}
\endbibitem

\bibitem{Recht2019Imagenet}
\begin{bchapter}
\bauthor{\bsnm{Recht}, \binits{B.}},
\bauthor{\bsnm{Roelofs}, \binits{R.}},
\bauthor{\bsnm{Schmidt}, \binits{L.}},
\bauthor{\bsnm{Shankar}, \binits{V.}}:
\bctitle{Do imagenet classifiers generalize to imagenet?}
In: \bbtitle{International Conference on Machine Learning},
pp. \bfpage{5389}--\blpage{5400}
(\byear{2019}).
\bcomment{PMLR}
\end{bchapter}
\endbibitem

\bibitem{Taori2020Measuring}
\begin{barticle}
\bauthor{\bsnm{Taori}, \binits{R.}},
\bauthor{\bsnm{Dave}, \binits{A.}},
\bauthor{\bsnm{Shankar}, \binits{V.}},
\bauthor{\bsnm{Carlini}, \binits{N.}},
\bauthor{\bsnm{Recht}, \binits{B.}},
\bauthor{\bsnm{Schmidt}, \binits{L.}}:
\batitle{Measuring robustness to natural distribution shifts in image
  classification}.
\bjtitle{Advances in Neural Information Processing Systems}
\bvolume{33},
\bfpage{18583}--\blpage{18599}
(\byear{2020})
\end{barticle}
\endbibitem

\bibitem{Miller2021Accuracy}
\begin{bchapter}
\bauthor{\bsnm{Miller}, \binits{J.P.}},
\bauthor{\bsnm{Taori}, \binits{R.}},
\bauthor{\bsnm{Raghunathan}, \binits{A.}},
\bauthor{\bsnm{Sagawa}, \binits{S.}},
\bauthor{\bsnm{Koh}, \binits{P.W.}},
\bauthor{\bsnm{Shankar}, \binits{V.}},
\bauthor{\bsnm{Liang}, \binits{P.}},
\bauthor{\bsnm{Carmon}, \binits{Y.}},
\bauthor{\bsnm{Schmidt}, \binits{L.}}:
\bctitle{Accuracy on the line: on the strong correlation between
  out-of-distribution and in-distribution generalization}.
In: \bbtitle{International Conference on Machine Learning},
pp. \bfpage{7721}--\blpage{7735}
(\byear{2021}).
\bcomment{PMLR}
\end{bchapter}
\endbibitem

\bibitem{Yadav2019Cold}
\begin{botherref}
\oauthor{\bsnm{Yadav}, \binits{C.}},
\oauthor{\bsnm{Bottou}, \binits{L.}}:
Cold case: The lost mnist digits.
Advances in neural information processing systems
\textbf{32}
(2019)
\end{botherref}
\endbibitem

\bibitem{Miller2020Effect}
\begin{bchapter}
\bauthor{\bsnm{Miller}, \binits{J.}},
\bauthor{\bsnm{Krauth}, \binits{K.}},
\bauthor{\bsnm{Recht}, \binits{B.}},
\bauthor{\bsnm{Schmidt}, \binits{L.}}:
\bctitle{The effect of natural distribution shift on question answering
  models}.
In: \bbtitle{International Conference on Machine Learning},
pp. \bfpage{6905}--\blpage{6916}
(\byear{2020}).
\bcomment{PMLR}
\end{bchapter}
\endbibitem

\bibitem{Rajpurkar2016squad}
\begin{botherref}
\oauthor{\bsnm{Rajpurkar}, \binits{P.}},
\oauthor{\bsnm{Zhang}, \binits{J.}},
\oauthor{\bsnm{Lopyrev}, \binits{K.}},
\oauthor{\bsnm{Liang}, \binits{P.}}:
Squad: 100,000+ questions for machine comprehension of text.
arXiv preprint arXiv:1606.05250
(2016)
\end{botherref}
\endbibitem

\bibitem{sugiyama2007covariate}
\begin{botherref}
\oauthor{\bsnm{Sugiyama}, \binits{M.}},
\oauthor{\bsnm{Krauledat}, \binits{M.}},
\oauthor{\bsnm{M{\"u}ller}, \binits{K.-R.}}:
Covariate shift adaptation by importance weighted cross validation.
Journal of Machine Learning Research
\textbf{8}(5)
(2007)
\end{botherref}
\endbibitem

\bibitem{hachiya2012importance}
\begin{barticle}
\bauthor{\bsnm{Hachiya}, \binits{H.}},
\bauthor{\bsnm{Sugiyama}, \binits{M.}},
\bauthor{\bsnm{Ueda}, \binits{N.}}:
\batitle{Importance-weighted least-squares probabilistic classifier for
  covariate shift adaptation with application to human activity recognition}.
\bjtitle{Neurocomputing}
\bvolume{80},
\bfpage{93}--\blpage{101}
(\byear{2012})
\end{barticle}
\endbibitem

\bibitem{Almeida2018Adapting}
\begin{barticle}
\bauthor{\bsnm{Almeida}, \binits{P.R.}},
\bauthor{\bsnm{Oliveira}, \binits{L.S.}},
\bauthor{\bsnm{Britto~Jr}, \binits{A.S.}},
\bauthor{\bsnm{Sabourin}, \binits{R.}}:
\batitle{Adapting dynamic classifier selection for concept drift}.
\bjtitle{Expert Systems with Applications}
\bvolume{104},
\bfpage{67}--\blpage{85}
(\byear{2018})
\end{barticle}
\endbibitem

\bibitem{Khamassi2018DiscussionAR}
\begin{barticle}
\bauthor{\bsnm{Khamassi}, \binits{I.}},
\bauthor{\bsnm{Mouchaweh}, \binits{M.S.}},
\bauthor{\bsnm{Hammami}, \binits{M.}},
\bauthor{\bsnm{Gh{\'e}dira}, \binits{K.}}:
\batitle{Discussion and review on evolving data streams and concept drift
  adapting}.
\bjtitle{Evolving Systems}
\bvolume{9},
\bfpage{1}--\blpage{23}
(\byear{2018})
\end{barticle}
\endbibitem

\bibitem{Cortes1995SVM}
\begin{barticle}
\bauthor{\bsnm{Cortes}, \binits{C.}},
\bauthor{\bsnm{Vapnik}, \binits{V.}}:
\batitle{Support-vector networks}.
\bjtitle{Machine learning}
\bvolume{20}(\bissue{3}),
\bfpage{273}--\blpage{297}
(\byear{1995})
\end{barticle}
\endbibitem

\bibitem{Hastie2009Elements}
\begin{bbook}
\bauthor{\bsnm{Hastie}, \binits{T.}},
\bauthor{\bsnm{Tibshirani}, \binits{R.}},
\bauthor{\bsnm{Friedman}, \binits{J.H.}},
\bauthor{\bsnm{Friedman}, \binits{J.H.}}:
\bbtitle{The Elements of Statistical Learning: Data Mining, Inference, and
  Prediction}
vol. \bseriesno{2}.
\bpublisher{Springer},
\blocation{New York}
(\byear{2009})
\end{bbook}
\endbibitem

\bibitem{Breiman2001RF}
\begin{barticle}
\bauthor{\bsnm{Breiman}, \binits{L.}}:
\batitle{Random forests}.
\bjtitle{Machine learning}
\bvolume{45}(\bissue{1}),
\bfpage{5}--\blpage{32}
(\byear{2001})
\end{barticle}
\endbibitem

\bibitem{Bishop2006Pattern}
\begin{bbook}
\bauthor{\bsnm{Bishop}, \binits{C.M.}},
\bauthor{\bsnm{Nasrabadi}, \binits{N.M.}}:
\bbtitle{Pattern Recognition and Machine Learning}
vol. \bseriesno{4}.
\bpublisher{Springer},
\blocation{New York}
(\byear{2006})
\end{bbook}
\endbibitem

\bibitem{Hand2007Principles}
\begin{barticle}
\bauthor{\bsnm{Hand}, \binits{D.J.}}:
\batitle{Principles of data mining}.
\bjtitle{Drug safety}
\bvolume{30}(\bissue{7}),
\bfpage{621}--\blpage{622}
(\byear{2007})
\end{barticle}
\endbibitem

\bibitem{Ovadia2019Uncertainty}
\begin{botherref}
\oauthor{\bsnm{Ovadia}, \binits{Y.}},
\oauthor{\bsnm{Fertig}, \binits{E.}},
\oauthor{\bsnm{Ren}, \binits{J.}},
\oauthor{\bsnm{Nado}, \binits{Z.}},
\oauthor{\bsnm{Sculley}, \binits{D.}},
\oauthor{\bsnm{Nowozin}, \binits{S.}},
\oauthor{\bsnm{Dillon}, \binits{J.}},
\oauthor{\bsnm{Lakshminarayanan}, \binits{B.}},
\oauthor{\bsnm{Snoek}, \binits{J.}}:
Can you trust your model's uncertainty? evaluating predictive uncertainty under
  dataset shift.
Advances in neural information processing systems
\textbf{32}
(2019)
\end{botherref}
\endbibitem

\bibitem{Amodei2016Concrete}
\begin{botherref}
\oauthor{\bsnm{Amodei}, \binits{D.}},
\oauthor{\bsnm{Olah}, \binits{C.}},
\oauthor{\bsnm{Steinhardt}, \binits{J.}},
\oauthor{\bsnm{Christiano}, \binits{P.}},
\oauthor{\bsnm{Schulman}, \binits{J.}},
\oauthor{\bsnm{Man{\'e}}, \binits{D.}}:
Concrete problems in ai safety.
arXiv preprint arXiv:1606.06565
(2016)
\end{botherref}
\endbibitem

\bibitem{Lesch2005metric}
\begin{bchapter}
\bauthor{\bsnm{Lesch}, \binits{S.}},
\bauthor{\bsnm{Kleinbauer}, \binits{T.}},
\bauthor{\bsnm{Alexandersson}, \binits{J.}}:
\bctitle{A new metric for the evaluation of dialog act classification}.
In: \bbtitle{Proceedings of the 9th Workshop on the Semantics and Pragmatics of
  Dialogue (SEMDIAL: DIALOR), Nancy, France},
pp. \bfpage{143}--\blpage{6}
(\byear{2005}).
\bcomment{Citeseer}
\end{bchapter}
\endbibitem

\bibitem{Rabanser2019Failing}
\begin{botherref}
\oauthor{\bsnm{Rabanser}, \binits{S.}},
\oauthor{\bsnm{G{\"u}nnemann}, \binits{S.}},
\oauthor{\bsnm{Lipton}, \binits{Z.}}:
Failing loudly: An empirical study of methods for detecting dataset shift.
Advances in Neural Information Processing Systems
\textbf{32}
(2019)
\end{botherref}
\endbibitem

\bibitem{Clark2019Ensemble}
\begin{botherref}
\oauthor{\bsnm{Clark}, \binits{C.}},
\oauthor{\bsnm{Yatskar}, \binits{M.}},
\oauthor{\bsnm{Zettlemoyer}, \binits{L.}}:
Don't take the easy way out: ensemble based methods for avoiding known dataset
  biases.
arXiv preprint arXiv:1909.03683
(2019)
\end{botherref}
\endbibitem

\bibitem{Shafieezadeh2015RobustLR}
\begin{botherref}
\oauthor{\bsnm{Shafieezadeh~Abadeh}, \binits{S.}},
\oauthor{\bsnm{Mohajerin~Esfahani}, \binits{P.M.}},
\oauthor{\bsnm{Kuhn}, \binits{D.}}:
Distributionally robust logistic regression.
Advances in Neural Information Processing Systems
\textbf{28}
(2015)
\end{botherref}
\endbibitem

\bibitem{Atashpaz2013complexity}
\begin{barticle}
\bauthor{\bsnm{Atashpaz-Gargari}, \binits{E.}},
\bauthor{\bsnm{Sima}, \binits{C.}},
\bauthor{\bsnm{Braga-Neto}, \binits{U.M.}},
\bauthor{\bsnm{Dougherty}, \binits{E.R.}}:
\batitle{Relationship between the accuracy of classifier error estimation and
  complexity of decision boundary}.
\bjtitle{Pattern recognition}
\bvolume{46}(\bissue{5}),
\bfpage{1315}--\blpage{1322}
(\byear{2013})
\end{barticle}
\endbibitem

\bibitem{Cortes2014Domain}
\begin{barticle}
\bauthor{\bsnm{Cortes}, \binits{C.}},
\bauthor{\bsnm{Mohri}, \binits{M.}}:
\batitle{Domain adaptation and sample bias correction theory and algorithm for
  regression}.
\bjtitle{Theoretical Computer Science}
\bvolume{519},
\bfpage{103}--\blpage{126}
(\byear{2014})
\end{barticle}
\endbibitem

\bibitem{Micevska2021SDDM}
\begin{barticle}
\bauthor{\bsnm{Micevska}, \binits{S.}},
\bauthor{\bsnm{Awad}, \binits{A.}},
\bauthor{\bsnm{Sakr}, \binits{S.}}:
\batitle{Sddm: an interpretable statistical concept drift detection method for
  data streams}.
\bjtitle{Journal of Intelligent Information Systems}
\bvolume{56}(\bissue{3}),
\bfpage{459}--\blpage{484}
(\byear{2021})
\end{barticle}
\endbibitem

\bibitem{Lima2021Comparative}
\begin{bchapter}
\bauthor{\bsnm{Lima}, \binits{M.}},
\bauthor{\bsnm{A~Fagundes}, \binits{R.A.d.}}, \betal:
\bctitle{A comparative study on concept drift detectors for regression}.
In: \bbtitle{Brazilian Conference on Intelligent Systems},
pp. \bfpage{390}--\blpage{405}
(\byear{2021}).
\bcomment{Springer}
\end{bchapter}
\endbibitem

\bibitem{cano2020kappa}
\begin{barticle}
\bauthor{\bsnm{Cano}, \binits{A.}},
\bauthor{\bsnm{Krawczyk}, \binits{B.}}:
\batitle{Kappa updated ensemble for drifting data stream mining}.
\bjtitle{Machine Learning}
\bvolume{109},
\bfpage{175}--\blpage{218}
(\byear{2020})
\end{barticle}
\endbibitem

\end{thebibliography}


\end{document}